\documentclass[11pt]{article}
\usepackage{hyperref}
\usepackage{jair}
\usepackage{times}
\usepackage{helvet}
\usepackage{courier}
\usepackage{theapa}
\usepackage{tabularx}
\usepackage[svgnames]{xcolor}
\usepackage{colortbl}
\usepackage{xparse}
\usepackage{amsmath}
\usepackage{amsthm}
\usepackage{amssymb}
\usepackage{xspace}
\usepackage{relsize}
\usepackage{graphicx}
\usepackage{multirow}
\usepackage{comment}
\usepackage{math_commands}
\usepackage{adjustbox}

\usepackage{algorithm}
\usepackage[noend]{algorithmic}

\usepackage{booktabs}   

\usepackage[frozencache,cachedir=.]{minted}

\usepackage{stackengine}
\stackMath
\newcommand\tsup[2][2]{%
 \def\useanchorwidth{T}%
  \ifnum#1>1%
    \stackon[-.5pt]{\tsup[\numexpr#1-1\relax]{#2}}{\scriptscriptstyle\sim}%
  \else%
    \stackon[.5pt]{#2}{\scriptscriptstyle\sim}%
  \fi%
}

\renewcommand{\tilde}[1]{\tsup[1]{#1}}

\newcommand{\pddl}[1]{\textsf{\small #1}}

\newcommand{\function}[1]{\textsc{#1}}

\usepackage{tcolorbox}
\usepackage{manfnt}
\tcbuselibrary{breakable}
\newcommand{\dnote}[1]{%
\vspace{1em}%
\begin{tcolorbox}[breakable]%
  \setlength{\parindent}{2em}%
  #1
\end{tcolorbox}%
}

\newcommand{\mycolor}[2]{\textcolor{#1}{#2}}

\newcommand{\red}[1]{\mycolor{red}{#1}}

\newcommand{\blue}[1]{\mycolor{blue}{#1}}
\newcommand{\cyan}[1]{\mycolor{cyan}{#1}}

\newcommand{\black}[1]{\mycolor{black}{#1}}
\newcommand{\gray}[1]{\mycolor{gray}{#1}}

\newcommand{\orange}[1]{\mycolor{orange}{#1}}

\newcommand{\teal}[1]{\mycolor{teal}{#1}}
\newcommand{\violet}[1]{\mycolor{violet}{#1}}

\def\_{\\[-0.3em]}

\newtheorem{defi}{Definition}
\newtheorem{ex}{Example}
\newtheorem{theo}{Theorem}

\makeatletter
\let\@myref\ref

\newcommand{\refsec}[1]{Section \@myref{#1}}
\newcommand{\refseq}[1]{Section \@myref{#1}}
\newcommand{\refsecs}[2]{Sections \@myref{#1}-\@myref{#2}}
\newcommand{\refig}[1]{Figure  \@myref{#1}}
\newcommand{\refigs}[2]{Figures \@myref{#1}-\@myref{#2}}
\newcommand{\reftbl}[1]{Table \@myref{#1}}
\newcommand{\refstep}[1]{Step \@myref{#1}}
\newcommand{\refalgo}[1]{Algorithm \@myref{#1}}
\newcommand{\refchap}[1]{Chapter \@myref{#1}}
\newcommand{\reflst}[1]{List \@myref{#1}}
\newcommand{\refeq}[1]{Equation \@myref{#1}}
\newcommand{\refeqs}[2]{Equations \@myref{#1}-\@myref{#2}}

\newcommand{\refthm}[1]{Theorem \@myref{#1}}
\newcommand{\refdef}[1]{Definition \@myref{#1}}
\newcommand{\refex}[1]{Example \@myref{#1}}
\newcommand{\refline}[1]{line\,\@myref{#1}}
\newcommand{\reflines}[2]{lines\,\@myref{#1}-\@myref{#2}}
\newcommand{\reftheo}[1]{Theorem \@myref{#1}}

\makeatother

\newcounter{list}[section]

\let\cite\shortcite

\NewDocumentCommand{\citet}{o m}{%
  \IfNoValueTF{#1}%
    {\citeauthor{#2}~\citeyear{#2}}
    {\citeauthor{#2}~\citeyear[#1]{#2}}%
}
\NewDocumentCommand{\citep}{o m}{%
  \IfNoValueTF{#1}%
    {\cite{#2}}
    {\cite[#1]{#2}}%
}

\newcommand{\ama}[1]{AMA$_{#1}^+$}

\newcommand{\BN}{\function{BN}}
\newcommand{\GS}{\function{GS}_{\tau}}
\newcommand{\BC}{\function{BC}_{\tau}}
\newcommand{\STGS}{\function{ST-GS}}
\newcommand{\loss}{\mathcal{L}}

\newcommand{\posp}{\function{pos}}
\newcommand{\negp}{\function{neg}}
\newcommand{\adde}{\function{add}}
\newcommand{\dele}{\function{del}}
\newcommand{\effect}{\function{effect}}

\newcommand{\sota}{State-of-the-Art\xspace}
\newcommand{\lsota}{state-of-the-art\xspace}  
\newcommand{\domind}{domain-independent\xspace}
\newcommand{\astar}{\xspace {$A^*$}\xspace}

\makeatletter

\newcommand{\newheuristic}[2]{%
 \def#1{%
  \ifmmode%
  h^\text{#2}\xspace%
  \else%
  \text{#2}\xspace%
  \fi%
 }%
}

\newheuristic{\blind}{blind}
\newheuristic{\lmcut}{LMcut}
\newheuristic{\mands}{M\&S}
\newheuristic{\pdb}{PDB}
\newheuristic{\ff}{FF}
\newheuristic{\ce}{CEA}
\newheuristic{\cg}{CG}
\newheuristic{\ad}{add}
\newheuristic{\lc}{LC}

\newcommand{\newUnitCostHeuristic}[2]{%
 \def#1{%
  \ifmmode%
  \hat{h}^\text{#2}\xspace%
  \else%
  \text{#2}\xspace%
  \fi%
 }%
}

\newUnitCostHeuristic{\lmcuto}{LMcut}
\newUnitCostHeuristic{\mandso}{M\&S}
\newUnitCostHeuristic{\ffo}{FF}
\newUnitCostHeuristic{\ceo}{CEA}
\newUnitCostHeuristic{\cgo}{CG}
\newUnitCostHeuristic{\ado}{add}
\newUnitCostHeuristic{\gco}{GoalCount}
\newUnitCostHeuristic{\lco}{LC}

\makeatother

\def\latentplanner{Latplan\xspace}

\newcommand{\init}{{\vx^{I}}}
\newcommand{\goal}{{\vx^{G}}}
\newcommand{\zinit}{{\vz^{I}}}
\newcommand{\zgoal}{{\vz^{G}}}
\newcommand{\encode}{\function{encode}}
\newcommand{\decode}{\function{decode}}

\newcommand{\defaultindex}{i}
\newcommand{\defaultcomma}{,}
\newcommand{\Xtr}{\mathcal{X}}
\newcommand{\Ztr}{\mathcal{Z}}
\newcommand{\xtr}[1][\defaultindex]{\vx^{#1}}
\newcommand{\ztr}[1][\defaultindex]{\vz^{#1}}

\newcommand{\rxbefore}[1][0]{\rvx^{#1}}
\newcommand{\rxafter}[1][1]{\rvx^{#1}}
\newcommand{\rzbefore}[1][0]{\rvz^{#1}}
\newcommand{\rzafter}[1][1]{\rvz^{#1}}

\newcommand{\xbefore}[1][\defaultindex\defaultcomma{}0]{\vx^{#1}}
\newcommand{\xafter}[1][\defaultindex\defaultcomma{}1]{\vx^{#1}}

\newcommand{\xbeforerec}[1][\defaultindex\defaultcomma{}0]{{\tilde{\vx}}^{#1}}
\newcommand{\xafterrec}[1][\defaultindex\defaultcomma{}1]{{\tilde{\vx}}^{#1}}
\newcommand{\xbeforealtrec}[1][\defaultindex\defaultcomma{}3]{{\tilde{\vx}}^{#1}}
\newcommand{\xafteraltrec}[1][\defaultindex\defaultcomma{}2]{{\tilde{\vx}}^{#1}}

\newcommand{\zbefore}[1][\defaultindex\defaultcomma{}0]{\vz^{#1}}
\newcommand{\zafter}[1][\defaultindex\defaultcomma{}1]{\vz^{#1}}
\newcommand{\qbefore}[1][\defaultindex\defaultcomma{}0]{\vq^{#1}}
\newcommand{\qafter}[1][\defaultindex\defaultcomma{}1]{\vq^{#1}}
\newcommand{\lbefore}[1][\defaultindex\defaultcomma{}0]{\vl^{#1}}
\newcommand{\lafter}[1][\defaultindex\defaultcomma{}1]{\vl^{#1}}

\newcommand{\zafterrec}[1][\defaultindex\defaultcomma{}1]{\tilde{\vz}^{#1}}

\newcommand{\zbeforealt}[1][\defaultindex\defaultcomma{}3]{\vz^{#1}}
\newcommand{\zafteralt}[1][\defaultindex\defaultcomma{}2]{\vz^{#1}}

\newcommand{\qafteralt}[1][\defaultindex\defaultcomma{}2]{\vq^{#1}}
\newcommand{\lbeforealt}[1][\defaultindex\defaultcomma{}3]{\vl^{#1}}
\newcommand{\lafteralt}[1][\defaultindex\defaultcomma{}2]{\vl^{#1}}

\newcommand{\action}[1][\defaultindex]{\va^{#1}}
\newcommand{\raction}[1][]{\rva^{#1}}
\newcommand{\qaction}[1][\defaultindex\defaultcomma{}a]{\vq^{#1}}
\newcommand{\laction}[1][\defaultindex\defaultcomma{}a]{\vl^{#1}}

\def\rxboth{\rxbefore,\rxafter}
\def\rzboth{\rzbefore,\rzafter}
\def\rza{\rzbefore,\raction}

\def\xboth{\xbefore,\xafter}

\def\xbothrec{\xbeforerec,\xafterrec}
\def\zboth{\zbefore,\zafter}

\def\za{\zbefore,\action}

\newcommand{\nn}{\nonumber\\}

\newcommand{\aaee}{\function{action}}
\newcommand{\aaed}{\function{apply}}
\newcommand{\aaer}{\function{regress}}

\newcommand{\aaep}{\function{applicable}}
\newcommand{\aaeq}{\function{regressable}}

\def\ref{\todo{Do not use ``ref'' directly!}}

\usepackage{fancyhdr}
\fancypagestyle{plain}{%
\fancyhf{} 
\fancyfoot[C]{\iffloatpage{}{\thepage}} 

}
\usepackage{longtable}
\makeatletter
  
  \@addtoreset{figure}{section}
  
 \@addtoreset{table}{section}
\makeatother

\jairheading{1}{2020}{1-15}{1/1}{1/1}
\ShortHeadings{Classical Planning in Deep Latent Space}{Asai, Kajino, Fukunaga \& Muise}
\firstpageno{1}

\author{%
\name Masataro Asai
\email masataro.asai\textcircled{$\alpha$}ibm.com \\
\addr MIT-IBM Watson AI Lab, IBM Research, Cambridge USA
\AND
\name Hiroshi Kajino
\email kajino\textcircled{$\alpha$}jp.ibm.com \\
\addr IBM Research - Tokyo, Tokyo Japan
\AND
\name Alex Fukunaga
\email fukunaga\textcircled{$\alpha$}idea.c.u-tokyo.ac.jp \\
\addr Graduate School of Arts and Sciences, University of Tokyo, Tokyo Japan
\AND
\name Christian Muise
\email christian.muise\textcircled{$\alpha$}queensu.ca \\
\addr School of Computing, Queen's University, Kingston Canada
}
\title{Classical Planning in Deep Latent Space}

\begin{document}

\maketitle
\begin{abstract}
Current domain-independent, classical planners require symbolic models of the problem domain and instance as input, resulting in a knowledge acquisition bottleneck.
Meanwhile, although deep learning has achieved significant success in many fields, the knowledge is encoded in a subsymbolic representation which is incompatible with symbolic systems such as planners.
We propose \latentplanner, an unsupervised architecture combining deep learning and classical planning.
Given only an unlabeled set of image pairs showing a subset of transitions allowed in the environment (training inputs),
\latentplanner learns a complete propositional PDDL action model of the environment.
Later, when a pair of images representing the initial and the goal states (planning inputs) is given,
\latentplanner finds a plan to the goal state in a symbolic latent space and returns a visualized plan execution.
We evaluate \latentplanner using image-based versions of 6 planning domains: 8-puzzle, 15-Puzzle, Blocksworld, Sokoban and Two variations of LightsOut.
\end{abstract}

\section{Introduction}

\label{sec:introduction}

Recent advances in domain-independent planning have greatly enhanced their capabilities.
However, planning problems need to be provided to the planner in a structured, symbolic representation such as Planning Domain Definition Language (PDDL) \cite{McDermott00}, and in general, such symbolic models need to be provided by a human, either directly in a modeling language such as PDDL, or via a compiler which transforms some other symbolic problem representation into PDDL.
This results in the {\it knowledge-acquisition bottleneck}, where the modeling step is sometimes the bottleneck in the problem-solving cycle.
The requirement for symbolic input poses a significant obstacle to applying planning in {\it new, unforeseen} situations where no human is available to create such a model or a generator, e.g., autonomous spacecraft exploration.
In particular, this first requires generating symbols from raw sensor input, i.e., the {\it symbol grounding problem} \cite{Steels2008}.

Recently,  significant advances have been made in neural network (NN) deep learning approaches for perceptually-based cognitive tasks including image classification \cite{deng2009imagenet}, object recognition \cite{ren2015faster}, speech recognition \cite{deng2013new}, machine translation
as well as  NN-based problem-solving systems \cite{dqn,neuraltm}.
However, the current state-of-the-art, pure NN-based systems do not yet provide guarantees provided by symbolic planning systems, such as deterministic completeness and solution optimality.

Using a NN-based perceptual system to
{\it automatically} provide input models for domain-independent planners could greatly expand the applicability of planning technology and offer the benefits of both paradigms.
\emph{We consider the problem of robustly,  automatically bridging the gap between such subsymbolic representations and the symbolic representations required by domain-independent planners}.

\begin{figure}[tbp]
 \centering
 \includegraphics[width=\linewidth]{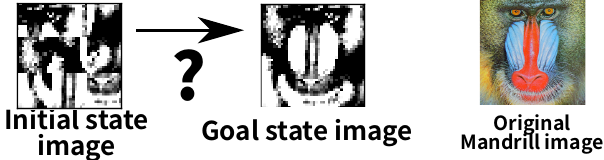}
 \caption{An image-based 8-puzzle.}
 \label{fig:mandrill-intro}
\end{figure}

\begin{figure*}[tb]
 \center
 \includegraphics[width=0.9\linewidth]{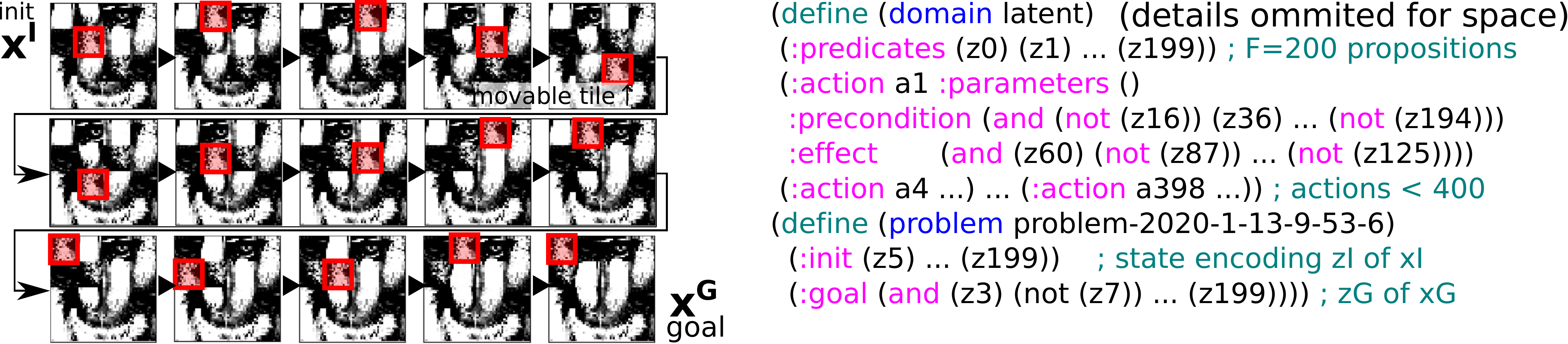}
 \caption{
(left) A 14-step optimal plan for a 15-puzzle instance generated by our system using off-the-shelf Fast Downward with $\lmcut$
using the PDDL generated by our system.
(right) The intermediate PDDL output from our NN-based system.
}
\label{fig:15puzzle}
\end{figure*}

\refig{fig:mandrill-intro} (left) shows a scrambled, 3x3 tiled version of the photograph on the right, i.e., an image-based instance of the 8-puzzle.
Even for humans, this photograph-based task is arguably more difficult to solve than the standard 8-puzzle because of the distracting visual aspects.
We seek a domain-independent system which, given only a set of unlabeled images showing the valid moves for this image-based puzzle, finds an optimal solution to the puzzle (\refig{fig:15puzzle}).
Although the 8-puzzle is trivial for symbolic planners, solving this image-based problem with
a domain-independent system which (1)  \emph{has no prior assumptions/knowledge}
 (e.g., ``sliding objects'', ``tile arrangement''), and (2) \emph{must acquire all knowledge from the images}, is nontrivial.
Such a system should not make assumptions about the image (e.g., ``a grid-like structure'').
The only assumption allowed about the nature of the task is that it can be modeled as a classical planning problem (deterministic and fully observable).

We propose Latent-space Planner (\latentplanner), an architecture that automatically generates a
symbolic problem representation from the subsymbolic input, that can be used as the input for a classical planner.
The current implementation of \latentplanner contains four neural components in addition to the classical planner:

\begin{itemize}
 \item A discrete autoencoder with multinomial binary latent variables, named \emph{State Autoencoder} (SAE), which learns a bidirectional mapping between the raw observations of the environment and its propositional representation.
 \item A discrete autoencoder with a single categorical latent variable and a skip connection, named \emph{Action Autoencoder} (AAE), which performs an unsupervised clustering over the latent transitions and generates action symbols.
 \item A specific decoder implementation of the AAE, named Back-To-Logit (BTL), which
       models the state progression / forward dynamics that directly compiles into STRIPS action effects.
 \item An identical BTL structure to model the state \emph{regression} / time-inverse dynamics which directly compiles into STRIPS action preconditions.
\end{itemize}

Given only a set of {\it unlabeled images} of the environment, and in an unsupervised manner,
we train Latplan to generate a symbolic representation.
Then, given a planning problem instance as a pair of initial and goal images such as \refig{fig:mandrill-intro}, \latentplanner
uses the SAE to map the problem to a symbolic planning instance, invokes a planner, then visualizes the plan execution by a sequence of images.

A system that generates symbols from the scratch has an advantage of being able to work on multiple domains more easily.
In planning, symbolic manipulation
enables the encoding of powerful domain-independent knowledge
that can be easily applied to multiple tasks without training data. 
For example, given a problem instance from a previously unseen STRIPS planning domain $D$,
a planning algorithm can often solve an instance of $D$ much faster than blind search by using domain-independent heuristic functions
that exploit $D$ based purely on the symbolic structure of the action model of $D$ \cite{hoffmann01,Helmert2009}.
The advantage of exploiting symbolic structures is a predicament
of the Physical Symbol Systems Hypothesis \cite{newell1976computer,newell1980physical,nilsson2007physical},
which states that ``[a] physical symbol system has the necessary and sufficient means for general intelligent action.''
In contrast,
while current learning-based approaches to planning such as AlphaZero \cite{alphazero} or MuZero \cite{muzero}
achieve impressive performance,
they require using or generating massive amounts of data in order to learn task-dependent evaluation functions and policies that result in high performance on a given task.
Transferring domain-independent strategies across tasks remains a challenge for learning-based approaches,
as they currently lack a convenient representation for expressing and exchanging task-independent knowledge between different systems.
It is trivial in classical planning, where domain-independent heuristics are available.

The paper is organized as follows.
We first begin with a review of preliminaries and background (\refsec{sec:background}).
We next provide our high-level problem statement (\refsec{sec:processes4classical-palnning}).
We next give an overview of the Latplan architecture (\refsec{sec:overview}).
In \refsec{sec:state-autoencoder},
we describe the SAE implemented as a Binary-Concrete Variational Auto-Encoder,
which generates propositional symbols from images.
We identify and define the {\it Symbol Stability Problem} which arises when grounding propositional symbols,
and propose countermeasures to address it.

Next, in \refsecs{sec:ama1-overview}{sec:ama4}, we explain our approach to action model learning.
Since the action model learning is a complex problem,
we introduce four increasingly sophisticated versions (AMA$_1$-\ama4),
where each version inherits the entire model of its previous version as a component.
We chose this presentation to help illustrate which aspect of the learning problem is addressed by each component.
AMA$_1$-\ama4 can be summarized as follows:
(\refsec{sec:ama1-overview}) AMA$_1$, a direct translation of image transitions to grounded actions,
(\refsec{sec:ama2})          AMA$_2$, which uses the AAE as a  general, black-box successor function ,
(\refsec{sec:ama3})         \ama3, an approach which trains a \emph{Cube-Space AE} network which jointly trains an SAE and a Back-to-Logit AAE for STRIPS domains and extracts a PDDL model compatible with off-the-shelf \sota planners, and
(\refsec{sec:ama4})         \ama4, an approach which trains a \emph{Bidirectional Cube-Space AE} network which improves upon \ama3 by using \emph{complete state regression} semantics to learn accurate action preconditions.

We then evaluate \latentplanner using image-based versions of the 8-puzzle, 15-puzzle, LightsOut (two versions), Blocksworld, and Sokoban domains.
\refsec{sec:training-evaluation} presents empirical evaluations of the accuracy and stability of the SAE, as well as the action model accuracy of \ama3 and \ama4.
\refsec{sec:planning-evaluation} presents empirical evaluation of end-to-end planning with Latplan, including the effectiveness of standard planning heuristics.
\refsec{sec:related} surveys related work, and we conclude with a discussion of our contributions and directions for future work (\refsec{sec:discussion}).
Some additional technical details, background, and data are presented in the Appendix.

Latplan is a first step in
bridging the gap between symbolic and subsymbolic reasoning,
therefore it currently has various limitations.
For example,
Latplan is evaluated in a fully-observable environment (although it is noisy).
Also, Latplan's state representation is entirely propositional and lack first-order logic generalization,
thus requires a retraining when new objects are added.
Latplan is limited to tasks where a single goal state is specified.
Finally, Latplan requires uniform sampling from the environment, which is nontrivial in many scenarios.
We discuss these limitations in detail in the discussion (\refsec{sec:discussion}).

This paper summarizes and extends the work that has appeared in \cite{Asai2018,Asai2019a,Asai2020}.
The major new technical contributions in this journal version are:
(1) the improved precondition learning enabled by the regressive action modeling (\refsec{sec:ama4}),
(2) theoretical justifications of the training objectives throughout the paper,
and (3) thorough empirical evaluations (\refsecs{sec:training-evaluation}{sec:planning-evaluation}).

\section{Preliminaries and Important Background Concepts}
\label{sec:background}

\subsection{Notations}

We denote a multi-dimensional array in bold and
its elements with a subscript (e.g., $\vx\in \R^{N\times M}$, $\vx_2 \in \R^M$).
An integer range $n\leq i<m$ is represented by $n..m$.
By analogy, we use dotted subscripts to denote a subarray, e.g. $\vx_{2..5}=(\vx_2,\vx_3,\vx_4)$.
$\1^D$ and $\0^D$ denote constant matrices of shape $D$ with all elements being 1/0, respectively.
$\va;\vb$ denotes a concatenation of tensors $\va$ and $\vb$ in the first axis
where the rest of the dimensions are same between $\va$ and $\vb$.
The $i$-th data point of a dataset is denoted by a superscript $^{i}$ which we may omit for clarity.
These superscripts are also sometimes abbreviated by $..$ to improve the readability,
e.g., $\vx^{0..2}=(\vx^0,\vx^1,\vx^2)$.
Functions (e.g., $\log,\exp$) are applied to arrays element-wise.
Finally, we denote $\B = [0,1]$.

\subsection{Propositional Classical Planning}
\label{sec:strips}
We define a grounded (propositional) STRIPS Planning problem with negative preconditions and unit costs
as a 4-tuple $\brackets{P,A,I,G}$
where
 $P$ is a set of propositions,
 $A$ is a set of actions,
 $I\subseteq P$ is the initial state, and
 $G\subseteq P$ is a goal condition.
Each action $a\in A$ is a 4-tuple
 $a=\brackets{\posp(a), \allowbreak \negp(a), \allowbreak \adde(a), \allowbreak \dele(a)}$ where
 $\posp(a)$, $\negp(a)$ are the positive and negative preconditions,
 $\adde(a)$, $\dele(a)$ are the add-effects and delete-effects,
 $\posp(a),\negp(a),\adde(a),\dele(a) \subseteq P$,
 $\posp(a) \cap \negp(a) = \emptyset$,
 and $\adde(a) \cap \dele(a) = \emptyset$.
A complete state (or just state) $s\subseteq P$ is a set of true propositions where those in $P \setminus s$ are presumed to be false.
A partial state is similarly represented (i.e. a subset of $P$), but those propositions not mentioned may be either true or false.
The initial state ($I$) is a complete state while the goal condition ($G$) and effect sets ($\posp(a)$, $\negp(a)$, $\adde(a)$, and $\dele(a)$) are partial states.
We say that a state $s$ entails a partial state $ps$ when $ps \subseteq s$ -- intuitively, every proposition that must be true in $ps$ is true in the state $s$.
An action $a$ is \emph{applicable} when $s\supseteq \posp(a)$ and $s\cap\negp(a)=\emptyset$,
and applying an action $a$ to $s$ yields a new successor state
$a(s) = (s \setminus \dele(a)) \cup \adde(a)$.
The task of classical planning is to find a \emph{plan} $(a_1,\cdots,a_n)$
which satisfies $G$ by the repeated application of applicable actions, i.e.,
$G \subseteq a_n \circ \cdots \circ a_1(I)$.
A plan $\pi$ is optimal when there are no other plans whose length is shorter than $\pi$.

\subsection{Autoencoders and Variational Autoencoders}
\label{sec:aevae}

\dnote{We review relevant probability theory in the appendix (\refsec{sec:info}).}

An Autoencoder (AE) is a type of neural network that learns
an identity function whose output matches the input \cite{hinton2006reducing}.
Autoencoders consist of an input $\vx$, a \emph{latent vector} $\vz$, an output $\hat{\vx}$,
an encoder $f$, and a decoder $g$, where $\vz=f(\vx)$ and $\hat{\vx}=g(\vz)=g(f(\vx))$.
The output is also called a \emph{reconstruction}.
The dimension of $\vz$ is typically smaller than the input; thus $\vz$ is considered a compressed representation of $\vx$.

The networks $f,g$ are optimized by minimizing a \emph{reconstruction loss} $\|\vx-\hat{\vx}\|$ under some norm,
which is typically a Square Error / L2-norm.
Which norm to use is semi-arbitrarily decided by a model designer ---
More explanation on the choice of the loss is given in the appendix (\refsec{sec:l2-norm}).
Assuming a 1-dimensional case,
let $x$ be a data point in the dataset, $z$ be a certain latent value,
and a probability distribution $p(x|z)$ be
what the neural network (and the model designer) believes is the distribution of $x$ given $z$.
Typical AEs for images assume that
$x$ follows a Gaussian distribution centered around the predicted value $\hat{x}=g(z)$, i.e.,
$p(x|z)=\mathcal{N}(x|\hat{x},\sigma) = \frac{1}{\sqrt{2\pi\sigma^2}} e^{-\frac{(x-\hat{x})^2}{2\sigma^2}}$ for an arbitrary fixed constant $\sigma$.
This leads to a \emph{negative log likelihood} (NLL)
\footnote{Likelihood is a synonym of probability.
``Log likelihood of $\vx$'' means a logarithm of a probability of observing the data $\vx$.
Negative log likelihood is its negation.}
$-\log p(x|z)=\frac{(x-\hat{x})^2}{2\sigma^2}+\log \sqrt{2\pi\sigma^2}$
which is a scaled/shifted square error / L2-norm reconstruction loss.
In images, they are summed across dimensions (pixels).

A Variational Autoencoder (VAE) is an autoencoder that additionally assumes that
the latent variable $\vz$ by default follows a certain distribution $p(\vz)$, typically called a \emph{prior distribution}.
A prior distribution is chosen arbitrarily by a model designer, such as a unit Normal distribution $\mathcal{N}(\0,\1)$.
More precisely, $\vz$ follows $p(\vz)$ unless the observation $\vx$ forces it otherwise ---
$\vz$ is meant to diverge from it based on the data through the training.

A VAE is trained by minimizing a sum of
a reconstruction loss $-\log p(\vx|\vz)$ and a KL divergence $\KL(q(\vz|\vx)\Mid p(\vz))$
between $q(\vz|\vx)$ and $p(\vz)$,
where
$q(\vz|\vx)$ is a distribution of the latent vector $\vz$ obtained by the encoder at hand.
$q(\vz|\vx)$ must be in the \emph{same family} of distributions as $p(\vz)$,
e.g., if $p(\vz)$ is Gaussian, $q(\vz|\vx)$ should also be a Gaussian,
in order to obtain an analytical form of the KL divergence that can be processed by automatic differentiation frameworks.
When $p(\vz)=\mathcal{N}(\0,\1)$,
a model engineer can then design an encoder that returns two vectors $\vmu,\vsigma=f(\vx)$
and sample $\vz$ as $\vz\sim \mathcal{N}(\vmu,\vsigma) = q(\vz|\vx)$,
using $\vz=\vmu + \vsigma \vepsilon$ where $\vepsilon$ is a noise that follows $\mathcal{N}(\0,\1)$ (reparameterization trick).
Then the analytical form of the KL divergence is obtained from the closed form of Gaussian distribution as follows:
\begin{align}
 \KL(\mathcal{N}(\vmu,\vsigma)||\mathcal{N}(\0,\1))
 &= \sum_{i} \frac{1}{2} \parens{\vsigma_i + \vmu_i^2 - 1 - \log \vsigma_i^2}.
 \label{eq:gaussian-kl}
\end{align}

A negative sum of the reconstruction loss and the KL divergence is called an \emph{Evidence Lower Bound} (ELBO),
a lower bound of the log likelihood $\log p(\vx)$ of observing $\vx$ \citep{kingma2013auto}.
Under \emph{maximum likelihood estimation} framework,
this lower bound characteristics make VAEs theoretically more appealing than AEs
because maximizing $\log p(\vx)$ is the true goal of unsupervised learning:
To maximize $\log p(\vx)$, we maximize its lower bound $\text{ELBO}(\vx)$,
and thus we minimize the loss function $-\text{ELBO}(\vx)$.
To derive a lower bound, ELBO uses the \emph{variational method}
which sees $q(\vz|\vx)$ as an approximation of $p(\vz|\vx)$.
\footnote{
We say $q(\vz|\vx)$ is a \emph{variational distribution}.
In addition,
both $q(\vz|\vx)$ and $p(\vz|\vx)$ are called \emph{posterior distribution}
because they are distributions of latent variables given observed variables.
When combined, we say $q(\vz|\vx)$ is a \emph{variational posterior distribution},
which is a \emph{variational approximation} of the true posterior distribution $p(\vz|\vx)$.
}
The lower bound matches $\log p(\vx)$ when $q=p$.
The proof is shown by using Jensen's inequality about exchanging an expectation and a convex function (in this case, $\log$)
as follows:
\begin{align}
\log p(\vx)
&=\log \sum_{\vz} p(\vx,\vz)
 =\log \sum_{\vz} p(\vx|\vz) p(\vz)
 =\log \sum_{\vz} p(\vx|\vz) \frac{p(\vz)}{q(\vz|\vx)} q(\vz|\vx) \nn
&=\log \parens{\E_{q(\vz|\vx)} \brackets{p(\vx|\vz) \frac{p(\vz)}{q(\vz|\vx)}}}\quad\text{(definition of an expectation.)}\nn
&\geq \E_{q(\vz|\vx)} \brackets{\log \parens{p(\vx|\vz) \frac{p(\vz)}{q(\vz|\vx)}}} \quad\text{(Jensen's inequality.)} \nn
& =    \E_{q(\vz\mid\vx)} \brackets{\log p(\vx\mid\vz)} -\E_{q(\vz|\vx)}\brackets{\log\frac{q(\vz\mid\vx)}{p(\vz)}} \nn
&=\E_{q(\vz\mid\vx)}\brackets{\log p(\vx\mid\vz)}
 -    \KL (q(\vz\mid\vx)\Mid p(\vz)) \quad\text{(definition of KL.)}\nn
 &=\text{ELBO}(\vx).\label{eq:elbo}
\end{align}

In contrast to VAEs,
the loss function of an AE (= reconstruction loss $\log p(\vx|\vz)$)
does not have this lower bound property.
Maximizing $\log p(\vx|\vz)$ does not guarantee that it maximizes $\log p(\vx)$.

Finally, among popular extensions of VAEs,
$\beta$-VAE \cite{higgins2017beta} uses a loss function that scales the KL divergence term with a hyperparameter $\beta\geq 1$.
\[
 \text{ELBO}_\beta(\vx)=\E_{q(\vz|\vx)}\brackets{\log p(\vx\mid\vz)} - \beta\KL (q(\vz\mid\vx)||p(\vz)).
\]
$\text{ELBO}(\vx) \geq \text{ELBO}_\beta(\vx)$ because $\KL$ is always positive,
thus $\text{ELBO}_\beta(\vx)$ is also a lower bound of $\log p(\vx)$.
However, some literature uses $\beta<1$
which is tuned manually by a visual inspection of the reconstructed image, which violates ELBO.
In this paper, we always use $\beta\geq 1$ so that it does not violate ELBO.
We will discuss the effect of different $\beta$ during the empirical evaluation.

\subsection{Discrete Variational Autoencoders}
\label{sec:discrete-vae}

While typically $p(z)$ is a Normal distribution $\gN(0,1)$ for continuous latent variables,
there are multiple VAE methods for discrete latent distributions.
A notable example is a method we use in this paper:
Gumbel-Softmax (GS) VAE \citep{jang2017categorical}.
Gumbel-Softmax VAE is independently discovered by \citet{MaddisonMT17} as Concrete VAE;
therefore it may be called as such in some literature.
The binary special case of Gumbel Softmax VAE is called Binary-Concrete (BC) VAE \citep{MaddisonMT17}.
These VAEs use a discrete, uniform categorical/Bernoulli(0.5) distribution as the prior $p(z)$,
and further approximate it with a continuous relaxation
by introducing a temperature parameter $\tau$ that is annealed down to 0.
In this section, we briefly summarize the information necessary for implementing GS and BC.

For Gumbel Softmax and Binary Concrete,
we denote corresponding stochastic activation functions in the latent space as $\GS(\vl)$ and $\BC(l)$,
where $\vl\in\R^C$, $l\in\R$, and $C$ denotes the number of classes represented by Gumbel Softmax.
$\vl$ and $l$ can be an arbitrary vector/scalar produced by the encoder neural network $f$.
Both functions use a temperature parameter $\tau$ which is annealed during the training
in a fixed schedule such as an exponential schedule $\tau(t)=Ar^{-(t-t_0)}$, where $t$ is a training epoch.
Each output, $\vz=\GS(\vl)$ or $z=\BC(l)$,
follows Gumbel-Softmax distribution $\gsdist(\vl,\tau)$ and Binary-Concrete distribution $\bcdist(l,\tau)$.
In other words,
the stochastic activation is equivalent to a sampling process from these distribution,
and the output is a sample from these distributions.
It can also be seen as a reparameterization trick for this distribution.

GS is defined as
\begin{align}
 \vz=\GS(\vl)=\softmax\parens{\frac{\vl+\function{Gumbel}^C(0,1)}{\tau}} \label{eq:gs}
\end{align}
where $\function{Gumbel}^C(0,1)=-\log(-\log \vu)$ and $\vu\sim\function{Uniform}^C(0,1)\in \B^C$.

BC is a binary special case (i.e. $C=2$) of GS.
It is defined as
\begin{align}
 z=\BC(l)=\function{Sigmoid}\parens{\frac{l+\function{Logistic}(0,1)}{\tau}} \label{eq:bc}
\end{align}
where $\function{Logistic}(0,1)=\log u-\log(1-u)$ and
$u\sim\function{Uniform}(0,1)\in \B$.

Both functions converge to discrete functions at the limit $\tau\rightarrow 0$:
$\GS(\vl)\to\argmax(\vl)$
and $\BC(l)\to\function{step}(l)=(l<0)\,?\,0:1$.
Note that we assume $\argmax$ returns a one-hot representation rather than the index of the maximum value
in order to maintain dimensional compatibility with $\softmax$ function.

In practice, GS-VAEs and BC-VAEs contain multiple latent vectors / latent scalars to model complex inputs,
which we call $n$-way Gumbel-Softmax / Binary-Concrete.
The latent space is denoted as $\vz\in \B^{n\times C}$ for Gumbel-Softmax, and $\vz\in \B^{n}$ for Binary Concrete.

\dnote{
For the derivation of the KL divergence of Gumbel-Softmax, refer to \refsec{sec:gsvae-elbo} in the Appendix.
To see how Gumbel-Softmax and Binary-Concrete relate to other discrete methods, refer to \refsec{sec:other-discrete} in the Appendix.
}

\subsection{Notational Convention for Neural Networks}

In this paper, we define each sub-network of the entire neural network
so that it does not contain the last activation.
This is because each network does not produce a sample of the probability distribution such as $\vz$;
the output of each network is instead a parameter of the probability distribution,
such as $\vl$ of a Gumbel-Softmax distribution $\gsdist(\vl,\tau)$
or $\vmu,\vsigma$ of a Gaussian distribution $\mathcal{N}(\vmu,\vsigma)$.
The sampling process for $\vz\sim\gsdist(\vl,\tau)$
is placed outside each subnetwork, although it is still part of the entire network.

\section{The Problem: Unsupervised Acquisition of a Classical Planning Model From Low-Level Inputs}
\label{sec:processes4classical-palnning}

Our goal is to build a system that acquires a classical planning model from low-level inputs (e.g., images) in a real-world (e.g., physical) environment without human involvement.
In order to fully automatically acquire symbolic models for Classical Planning,
we need
\emph{Symbol Grounding} \cite{harnad1990symbol,Steels2008,taddeo2005solving}
and \emph{Action Model Acquisition} \cite[AMA]{zhuo2014action,zhuo2015crowdsourced},
which are one of the key challenges in achieving autonomous symbolic systems for real-world environments and in Artificial Intelligence.

Symbol Grounding is an unsupervised process of establishing a mapping
between
symbols and
noisy, continuous and/or unstructured inputs.
Action Model Acquisition (also known as Planning Operator Acquisition \cite{wang1995learning}, or Action Model Learning \cite{YangWJ07}) is indeed a grounding process for actions,
typically with a focus on descriptive action schema (preconditions and effects in the STRIPS subset of PDDL) as the referents.
To clarify these high-level goals, we formalize symbol grounding and discuss its relation to PDDL construction.
Following a traditional 
definition seen in the LISP family of programming languages, formally:
\begin{defi}
 A symbol is a tuple of a \emph{name} and a \emph{referent},
 which can be uniquely looked up in a \emph{knowledge base} using the name as a key.
 \emph{Symbol grounding} is a process of assigning referents to symbols.
\end{defi}
For example, a \emph{name} can be an ASCII string or an integer.
A \emph{referent} is an arbitrary representation of a meaning of the symbol, e.g.,
a referent of a \emph{predicate symbol} is a boolean function, while a referent of an \emph{action symbol} is an action schema.
We do not distinguish whether referents are hand-coded or learned;
thus a predicate can be a binary classifier learned by a machine learning system.
A symbol \emph{designates} a referent \cite{newell1976computer}.
A \emph{knowledge base} is a namespace (e.g., a hash table) that defines a type of a symbol.

Symbols with the same name may exist in different namespaces and designate different referents.
For example,
many English words, such as ``move'', ``cut'', or ``train'', can be seen as
an action symbol (a verb), a predicate symbol (an adjective), or an object symbol (a noun) simultaneously.
Our view is in line with historic literature,
e.g., \citet{russell1995artificial} mentions propositional/constant(=object)/predicate/function symbols,
and \citet{vera1993situated} wrote ``The `wash' ... is a symbol denoting the action required'' (an action symbol).

PDDL has \emph{at least} six kinds (i.e., knowledge bases) of symbols:
Propositions, actions, objects, predicates, problems, and domains (\reftbl{tab:type-of-symbols}).
Each type of symbol requires its own mechanism for grounding.
For example, the large body of work on recognizing
objects (e.g., individual faces) and their attributes (male, female) in images, or scenes in videos (e.g., cooking),
can be viewed as grounding object, predicate and action symbols, respectively.
Extensions of PDDL may require more types of symbols, such as numeric fluents in numeric planning \cite{fox2003pddl2}.
Symbol grounding does not target purely syntactic constructs such as \texttt{and}, \texttt{or} or \texttt{:requirement}
that do not need referents.
Of six types, we aim to develop mechanisms for propositions and actions
because our goal is to generate a propositional classical planning model in a single environment.

Finally, in order to fully avoid the knowledge acquisition bottleneck,
a system must \emph{generate} symbols without human input.
The generation of symbols has long been recognized as an imporant aspect of symbol grounding \cite{harnad1990symbol}. 
In contrast to systems which can autonomously generate symbols,
systems which does not generate symbols but only grounds symbols given by humans are characterized as \emph{parasitic} by   
\citet{taddeo2005solving}, as 
such human-provided symbols already impose discrete categorization of the observations.

\begin{table}[tbp]
\centering
\begin{tabular}{ll}
Types of symbols & \\
\hline
Object symbols    & \textbf{panel7, x\(_{\text{0}}\), y\(_{\text{0}}\)} \ldots{}               \\
Predicate symbols & (\textbf{empty} ?x ?y) (\textbf{up} ?y\(_{\text{0}}\) ?y\(_{\text{1}}\))   \\
Propositions      & \textbf{empty\(_{\text{5}}\)} = (empty x\(_{\text{2}}\) y\(_{\text{1}}\)) \\
Action symbols    & (\textbf{slide-up} panel\(_{\text{7}}\) x\(_{\text{0}}\) y\(_{\text{1}}\)) \\
Problem symbols   & \textbf{eight-puzzle-instance1504}, etc.                                   \\
Domain  symbols   & \textbf{eight-puzzle}, \textbf{hanoi}                                      \\
\hline
\end{tabular}
\caption{Six types of symbols in a PDDL definition.}
\label{tab:type-of-symbols}
\end{table}

\section{\latentplanner: Overview of the System Architecture}
\label{sec:overview}

\latentplanner is a framework for
\emph{domain-independent classical planning in environments where symbolic input (PDDL models) are unavailable and only subsymbolic sensor input (e.g., images) are accessible.}
\latentplanner operates in two phases: The training phase and the planning phase.
Its abstract pipeline is described in \refalgo{alg:latplan}.
The rest of the section describe the role of each abstract procedure,
although the details may vary depending on the implementation.

\begin{algorithm}[htb]
 Training Phase:
 \begin{algorithmic}[1]
  \REQUIRE Dataset $\Xtr$, untrained machine learning model $M$
  \STATE Trained model $M' \gets \function{train}(M,\Xtr)$ \label{line:train}
  \STATE $M'$ provides functions $\encode$ and $\decode$. \label{line:provide-encode-decode}
  \STATE PDDL domain file $D \gets \function{generateDomain}(M')$ \label{line:generate-domain}
  \RETURN $M', D$
 \end{algorithmic}
 Planning Phase:
\begin{algorithmic}[1]
  \REQUIRE $M', D$, initial state observation $\init$, goal state observation $\goal$
  \STATE Encode $\init, \goal$ into propositional states $\zinit, \zgoal$ \label{line:encode-init-goal}
  \STATE PDDL problem file $P \gets \function{generateProblem}(\zinit, \zgoal)$ \label{line:generate-problem}
  \STATE Plan $\pi=(a_0, a_1, \ldots, ) \gets \function{solve}(P, D)$ using a planner (e.g., Fast Downward) \label{line:plan}
  \STATE State trace $(\zinit=\vz^0,\  \vz^1=a_0(\vz^0),\  \vz^2=a_1(\vz^1),\  \ldots, \zgoal) \gets \function{simulate}(\pi, \zinit, D)$
 using a plan validator for PDDL, e.g., VAL \cite{howey2003val}.  \label{line:simulate}
  \RETURN Decode an observation trace $(\init=\vx^0, \vx^1, \vx^2, \ldots, \goal)$. \label{line:decode-plan}
 \end{algorithmic}
 \caption{An abstract pipeline of Latplan framework.}
 \label{alg:latplan}
\end{algorithm}

\subsection{Training Phase}

In the training phase,
it first trains a model $M$ which learns state representations and transition rules of the environment
entirely from subsymbolic state transition data $\Xtr$ (e.g., image-based observations) using deep neural networks (\refline{line:train}).
This is done by an abstract procedure $\function{train}$ which obtains a trained model $M'$.
$M'$ provides a function $\encode$ that maps subsymbolic observations into symbolic states,
and a function $\decode$ that maps symbolic states back to subsymbolic visualization (\refline{line:provide-encode-decode}).
Then we produce a PDDL domain file $D$ using an abstract procedure $\function{generateDomain}$ (\refline{line:generate-domain}).

$\Xtr$ is a set of pairs of observations (e.g., raw image data) sampled from the environment.
The $\defaultindex$-th pair in the dataset $\xtr=(\xbefore, \xafter) \in \Xtr$ is
a transition from an observation $\xbefore$ to another observation $\xafter$ caused by an unknown high-level action.
As discussed in Section~\ref{sec:processes4classical-palnning}, there are mainly three key challenges in training \latentplanner:
\textbf{(1)} generating and grounding \emph{propositional symbols},
\textbf{(2)} generating and grounding \emph{action symbols}, and
\textbf{(3)} acquiring a \emph{descriptive action model}.
The first challenge arises because we have no access to the state representation.
The second challenge arises because we have no access to the action labels that caused the transitions.
The third challenge arises because no symbolic representation of the action is available.

Challenge \textbf{(1)} is addressed by a \emph{State AutoEncoder} (SAE) (\refig{fig:sae}) neural network that learns a bidirectional mapping between subsymbolic raw data $\vx$ (e.g., images)
 and propositional states $\vz\in\braces{0,1}^F$, i.e., $F$-dimensional bit vectors.
This generates and grounds propositional symbols to feature extractors encoding complex patterns found in the image.
The network consists of two functions $\encode$ and $\decode$, where
$\encode$ encodes an image $\vx$ to a bit vector $\vz$,
and $\decode$ decodes $\vz$ back to an image $\tsup[1]{\vx}$.
These are trained so that $\tsup[1]{\vx}\approx\vx$ holds.

Challenge \textbf{(2)} and \textbf{(3)} are more intricate, and we propose four variants of action model acquisition methods.
The mapping $\encode$ from $\braces{\ldots\xbefore, \xafter\ldots}$ to $\braces{\ldots\zbefore, \zafter\ldots}$
provides a set of propositional transitions $\ztr=(\zbefore,\zafter)\in\Ztr$.
Using $\Ztr$, action model acquisition methods have to learn not only action symbols but also their transition rules.
The four variants, in increasing order of sophistication, are as follows.
\begin{itemize}
 \item
   $\function{generateDomain}$ of AMA$_1$ applies the SAE to pairs of image transitions to obtain propositional transition $\ztr$ and directly converts each $\ztr$ to a ground STRIPS action.
For example, in a state space represented by 2 latent space propositions $\vz=(\vz_1,\vz_2)$,
a transition from $\zbefore=(0,1)$ to $\zafter=(1,0)$ is translated into an action
with $\posp(a)= \braces{\vz_2}, \negp(a)= \braces{\vz_1}, \adde(a) = \braces{\vz_1}, \dele(a) = \braces{\vz_2}$.
AMA$_1$ is developed to demonstrate the feasibility of using SAE-generated propositional symbols directly with existing planners.
While it successfully demonstrates the feasibility, AMA$_1$ is impractical in that it does not generalize $\Xtr$:
AMA$_1$ only encodes the transitions (ground actions) in the training data $\Xtr$. 
Thus, generating an AMA$_1$ model which can be used to find a path between an arbitrary start and goal state (assuming such a path exists in the domain) requires that $\Xtr$ contains the transitions necessary to form such a path -- in the worst case, this may require the entire state space (i.e., all possible valid image transitions) as input. Furthermore, the size of the PDDL model is proportional to the number of transitions in the state space,
slowing down the planning preprocessing and heuristic calculation at each search node.
      In other words, AMA$_1$ only addresses the challenge \textbf{(1)} (using SAE).
 \item
   AMA$_2$ is a neural architecture that jointly learns action symbols and action models from a small subset of transitions in an unsupervised manner.
   It learns a black-box successor generation function, which, given a latent space symbolic state, returns its successors.
   Unlike existing methods, AMA$_2$ does not require high-level action symbols as part of the input.
   While AMA$_2$ is a general approach to learning a successor function which does not make strong assumptions about the domain (e.g., STRIPS assumptions),
this limits its applicability to classical planning because it does not produce PDDL-compatible action descriptions
and it cannot implement $\function{generateDomain}$;
it is incompatible with PDDL-based solvers and cannot leverage existing implementations of domain-independent heuristics.
It requires a search algorithm implementation which does not require descriptive action models, such as novelty-based planners \cite{frances2017purely},
or a best-first search algorithm with trivial heuristics which work with AMA$_2$'s black-box limitation, such as Goal Count heuristics \cite{fikes1972strips}.
      In other words, AMA$_2$ addresses the challenge \textbf{(1)} and \textbf{(2)}.
 \item
AMA$_3^+$ is a neural architecture that improves AMA$_2$
so that the resulting neural network can be directly compiled into a descriptive action model (PDDL).
AMA$_3^+$ builds on the previously published AMA$_3$ to provide an architecture and an optimization objective that are theoretically motivated by the Evidence Lower BOund (ELBO).
AMA$_3^+$ has a neural network component that exclusively models the STRIPS/PDDL-compatible action effects
and can implement $\function{generateDomain}$,
but it requires a separate, ad-hoc process for generating preconditions.
      In other words, it addresses the challenge \textbf{(3)}, but only partially.
 \item
AMA$_4^+$, which we propose in this paper,
is an enhancement of AMA$_3^+$ with a new ability to learn and emit STRIPS/PDDL-compatible preconditions.
Unlike AMA$_3^+$, it does not require a separate, ad-hoc process which results in less accurate preconditions.
AMA$_4^+$ models the state space with \emph{complete state regression} semantics,
which is inspired by SAS+ formalism \cite{backstrom1995complexity}.
      This finally addresses all of challenges \textbf{(1)}, \textbf{(2)}, and \textbf{(3)}.
\end{itemize}

We reemphasize that each new model inherits the entire components of its predecessor:
SAE in AMA$_1$ is included in AMA$_2$ and later,
AAE in AMA$_2$ is included in \ama3 and later as a subnetwork,
and finally, \ama4 contains \ama3 as a subnetwork.
Therefore, understanding each earlier variant is important for
understanding the full potential and mechanics of our most sophisticated model \ama4.

\subsection{Planning Phase}

In the planning phase,
\latentplanner takes as input a \emph{planning input} $(\init, \goal)$, a pair of subsymbolic observations (raw data, images)
corresponding to an initial and goal state of the environment.
We first apply $\encode$ to each observation and obtain their symbolic representations (\refline{line:encode-init-goal}).
An abstract procedure $\function{generateProblem}$ then generates a PDDL problem file containing these symbolic initial and goal states (\refline{line:generate-problem}).
An abstract procedure $\function{solve}$ then solves the problem in the learned symbolic state space and returns a symbolic plan,
using an off-the-shelf planner such as Fast Downward \cite{Helmert2006} (\refline{line:plan}).
It then obtains a propositional intermediate state trace $(\zinit,\ \ldots,\ \zgoal)$ of the plan
by applying a plan simulator ($\function{simulate}$) to the symbolic initial state and the symbolic plan (\refline{line:simulate}).

While we have a symbolic state trace at this point, they are not interpretable for a human observer, which is an issue.
This is primarily because
the states in the state trace are generated by SAE neural network which was trained unsupervised ---
each bit of the state has a ``meaning'' in the latent space determined by the neural network, which does not necessarily directly correspond to human-interpretable notions.
Furthermore, the ``actions'' in the plan trace are transitions according to the AMA, but are not necessarily directly interpretable by a human.
In order to make the plan generated by \latentplanner understandable to a human (or other agents outside \latentplanner),
\latentplanner outputs a step-by-step visualization of the plan execution, $(\init,\ \ldots,\ \goal)$ (e.g. \refig{fig:15puzzle}),
by decoding the latent bit vectors for each intermediate state in $(\zinit,\ \ldots,\ \zgoal)$ (\refline{line:decode-plan}).

While we present an image-based implementation (``data'' = raw images),
the architecture itself does not make such assumptions
and could be applied to other types of data such as audio/text.

\section{State AutoEncoder~(SAE) for Propositional Symbol Grounding}
\label{sec:state-autoencoder}
This section presents a method to a propositional representation of a real-world environment on which a planning solver can perform logical reasoning.
We believe such a representation must satisfy the following three properties to be practical:
\begin{itemize}
 \item The ability to describe unseen world states using the same symbols,
 \item Similar images for ``the same world state'' should map to the same representation,
 \item The ability to map symbolic states back to images.
\end{itemize}
For example, one may come up with a trivial method to simply discretize the pixel values of an image array or to compute an image hash function.
Such a trivial representation lacks robustness and the ability to  generalize. 

In the following, \refsec{naive-sae} presents a State AutoEncoder~(SAE) as a framework to address propositional symbol grounding by a discrete VAE.
As elaborated in \refsec{sec:unstable}, we find that its naive implementation is unfavorable to a planning system because of unstable symbols caused by two sources of uncertainty.
\refsecs{sec:argmax}{sec:bcvae-prior} present two methods to address them.

\subsection{Naive State AutoEncoder}\label{naive-sae}

\begin{figure}[tbp]
 \centering
 \includegraphics[width=0.6\linewidth]{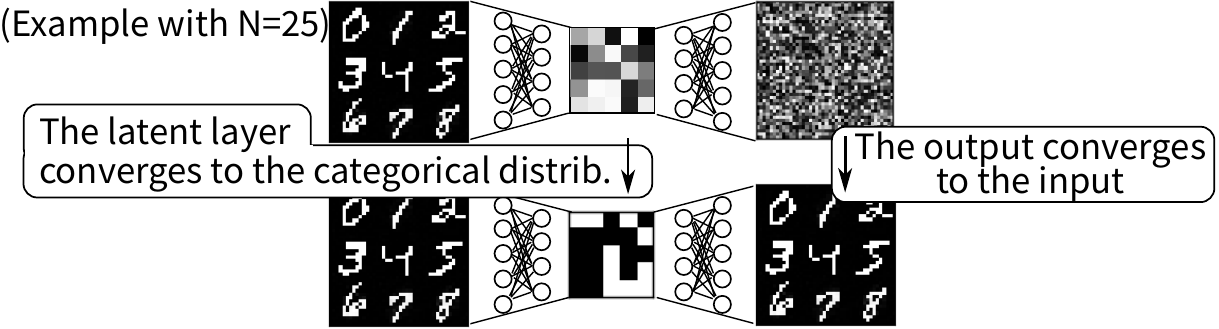}
 \caption{
We train the State Autoencoder by
 minimizing the sum of the reconstruction loss and the variational loss of Binary-Concrete.
As the training continues, the output of the network converges to the input images.
Also, as the temperature $\tau$ decreases during training,
the latent values approach either 0 or 1.}
\label{fig:sae}
\end{figure}

Our first technical contribution is a State AutoEncoder~(SAE) implemented by a discrete VAE (\refig{fig:sae}), which allows us to obtain a propositional representation satisfying the three requirements described above.
\emph{Our key observation is that the categorical variables modeled by discrete VAEs are compatible with symbolic reasoning systems}.
For example, binary latent variables of Binary-Concrete VAE
can be used as propositional symbols in STRIPS language,
providing a solution to the propositional symbol grounding, i.e., generation and grounding of propositional symbols.

The trained SAE provides an approximation of a bidirectional mapping $f, f^{-1}$ between the raw inputs (subsymbolic representation) and their symbolic representations:
\begin{itemize} 
\setlength{\itemsep}{-0.3em}
\item $\vz=f(\vx)$ maps an image $\vx\in \R^{H \times W \times C}$ to a boolean vector $\vz\in \B^F$.
\item $\tilde{\vx}\approx f^{-1}(\vz)$ maps a boolean vector $\vz$ to an image $\tilde{\vx}$.
\end{itemize}
Here, $H,W,C$ represents the height, the width, and the number of color channels of the image.
$f(\vx)$ maps a raw input $\vx$ to a symbolic representation by
feeding the raw input to the encoder network which ends with a Binary Concrete activation, resulting in a binary vector of length $F$.
$f^{-1}(\vz)$ maps a binary vector $\vz$ back to an image.
These are lossy compression/decompression functions, so in general,
$\tilde{\vx}$ may have an acceptable amount of errors
from $\vx$ for visualization.

In order to map an image to a latent state with sufficient accuracy,
the latent layer requires a certain capacity.
The lower bound of the number of propositional variables $F$ is
the encoding length of the states, i.e., $F \geq \log_2 \bars{S}$ for a state space $S$.
In practice, however, obtaining the most compact representation is not only difficult but also unnecessary,
and we use a hyperparameter $F$ which could be significantly larger than $\log_2 \bars{S}$.

It is {\it not} sufficient to simply use traditional activation functions such as sigmoid or softmax and round the continuous activation values in the latent layer to obtain discrete 0/1 values.
In order to map the propositional states back to images,
we need a decoding network trained for 0/1 values.
A rounding-based scheme would be unable to restore the images because the decoder is not trained with inputs near 0/1 values.
Also, simply using the rounding operation as a layer of the network is infeasible
because rounding is non-differentiable, precluding backpropagation-based training of the network.
Furthermore, as we discuss in Appendix \refsec{sec:other-discrete}, AEs with Straight-Through step function
is known to be outperformed by VAEs in terms of accuracy,
and its optimization objective lacks theoretical justification as a lower bound of the likelihood.

The SAE implementation can easily and significantly benefit from progress made by the image processing community.
We augmented the VAE implementation with a GaussianNoise layer to add noise robustness \cite{vincent2008extracting},
as well as Batch Normalization \cite{ioffe2015batch} and Dropout \cite{hinton2012improving},
which helps faster training under high learning rates (see experimental details in \refsec{sec:network-specifications}).

\subsection{The Symbol Stability Problem: Issues Caused by Unstable Propositional Symbols}
\label{sec:unstable}

\begin{figure}[htb]
 \centering
 \includegraphics{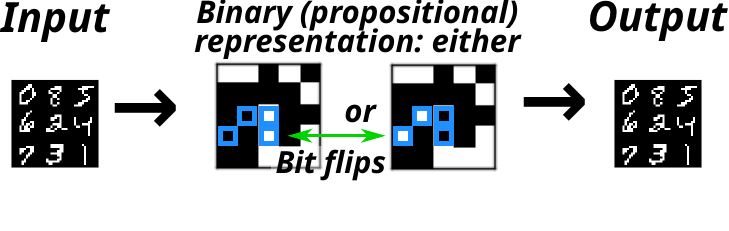}
 \caption{Propositions found by SAEs may contain uninformative random bits that do not affect the output.}
 \label{fig:unstable}
\end{figure}

Propositional representations learned by Binary Concrete VAEs
in the original form proposed and evaluated by \citep{jang2017categorical,MaddisonMT17}
have a problematic behavior that makes them less suitable for propositional reasoning.
While the SAE can reconstruct the input with high accuracy,
the learned latent representations are not ``stable,'' i.e.,
some propositions may flip the value (true/false) randomly
given identical or nearly identical image inputs (\refig{fig:unstable}).

The core issue with the vanilla Binary Concrete VAEs
is that the class probability for the class ``true'' and the class ``false''
could be neutral (e.g., around 0.5) at some neuron,
causing the value of the neuron to change frequently due to the stochasticity of the system.
The source of stochasticity is twofold:

\begin{itemize}
 \item \emph{Systemic/epistemic uncertainty}:
The first source is the random sampling in the network,
which introduces stochasticity and causes the propositions to change values even for the exact same inputs.
 \item \emph{Statistical/aleatoric uncertainty}:
The second source is the stochastic observation of the environment, which corrupts the input image.
When the class probabilities are almost neutral,
a tiny variation in the input image may cause the activation to cross the decision boundary for each neuron,
causing bit flips.
In contrast, humans still regard the corrupted image as the ``same'' image.
\end{itemize}

These unstable symbols are harmful to symbolic reasoning because
they break the identity assumption built into the recipient symbolic reasoning algorithms such as classical planners.
It causes several issues:
Firstly, the performance of
search algorithms (e.g., \astar) that run on the state space generated by the propositional vectors
are degraded by multiple propositional states which correspond to the same real-world state.
As standard symbolic duplicate detection techniques would treat such spurious representations of the same state as different states,
the search algorithm can unnecessarily re-expand the ``same'' real-world state several times, slowing down the search.

\begin{figure}[tb]
 \centering
 \includegraphics{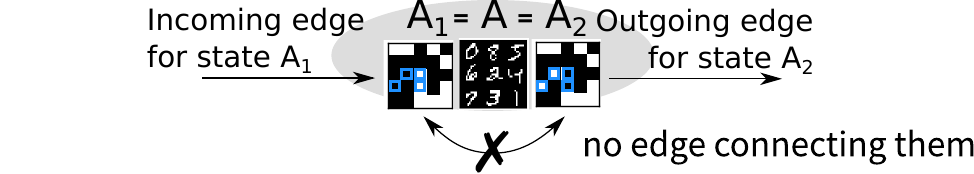}
 \caption{Random variations of the propositional encoding could disconnect the search space.}
 \label{fig:disconnected}
\end{figure}

Secondly, the state space could be disconnected due to such random variations (\refig{fig:disconnected}).
Some states may be reachable only via a single variation of the real-world state and are not connected to
another propositional variation of the same real-world state.
One way to circumvent this phenomenon is \emph{state augmentation},
which samples the propositional states at every node
by decoding and encoding the current state several times.
This is inefficient, and is also infeasible in the standard PDDL-based classical planners
that operate completely in the propositional space.

Thirdly, in order to reduce the stochasticity of the propositions, we encounter a hyperparameter tuning problem
which is costly when we train a large NN.
The neurons that behave randomly for the same or perturbed input do not affect the output,
i.e., they are unused and uninformative.
Unused neurons appear because the network has an excessive capacity to
model the entire state space, i.e., they are surplus neurons.
Therefore, a straightforward solution to address this issue is to reduce the size of the latent space $F$.
On the other hand, if $F$ is too small, it lacks the capacity to represent the state space,
and the SAE no longer learns to reconstruct the real-world image.
Thus we need to find an \emph{ideal} value of $F$, a costly and difficult hyperparameter tuning problem.

These unstable propositional symbols indicate that
the Binary Concrete VAE learns the mapping between images and binary representations as a many-to-many relationship.
While this property has not been considered as a major issue in the machine learning community
where accuracy is the primary consideration,
its unstable latent representation poses a significant impediment to the reasoning ability of the planners.
Unlike machine learning tasks,
symbolic planning requires a mapping
that abstracts many images into a single symbolic state, i.e., many-to-one mapping.
To this end, it is necessary to develop an autoencoder that learns a
many-to-one relationship between images and binary representations.

Fundamentally,
the first two harmful effects are caused by the fact that
the representation learned by a standard Binary Concrete VAE lacks a critical feature of symbols, \emph{designation} \cite{newell1976computer},
that each symbol uniquely designates a referent,
e.g., the referents of the symbols grounded by SAEs are the truth assignments based on visual features.
If the grounding (meaning) of a symbol changes frequently and unexpectedly, the entire symbolic manipulation is fruitless
because the underlying symbols are not tied to any particular concept and do not represent the real-world.

Thus, for a symbol grounding procedure to produce a set of symbols for symbolic reasoning,
it is not sufficient to find a set of symbols that 
can be mapped to real states in the environment;
It should find a \emph{stable} symbolic representation that \emph{uniquely} represents the environment.

\begin{defi}
A symbol is \emph{stable} when its referents are identical
for the same observation, under some equivalence relation (e.g., invariance or noise threshold on the observation).
\end{defi}

\begin{ex}
A propositional symbol points to a boolean value.
The value should not change under the same real-world state and its noisy observations.
\label{example:stability-state}
\end{ex}

\begin{ex}
A symbolic state ID points to a set of propositions whose values are true.
The content of the set should be the same under the same real-world state and its observations.
\end{ex}

\begin{ex}
A symbolic action label points to a tuple containing an action definition (e.g., preconditions).
It should point to the same action definition each time it observes the same state transition.
\end{ex}

\begin{ex}
Say that a real-world state $s_1$ transitions to another state $s_2$ by performing a certain symbolic action $a$.
The representation obtained by applying the symbolic action definition of $a$ to the symbolic representation of $s_1$
should be equivalent to the symbolic representation directly observed from $s_2$.
\label{example:stability-effect}
\end{ex}

The stability of the representation obtained by a NN depends
on
its inherent (systemic, epistemic) stochasticity of the system during the runtime (as opposed to the training time) as well as
the stochasticity of the environment (statistical, aleatoric).
Thus, \emph{any} symbol grounding system potentially suffers from
the symbol stability problem.
As for the stochasticity of the environment,
in many real-world tasks, it is common to obtain stochastic observations
due to external interference, e.g., vibrations of the camera caused by the wind.
As for the stochasticity of the network,
both
VAEs \cite{kingma2013auto,jang2017categorical,higgins2017beta} used in Latplan
and
GANs (Generative Adversarial Networks) \cite{goodfellow2014generative} used in Causal InfoGAN \cite{kurutach2018learning}
rely on sampling processes.

The stability of the learned representation is orthogonal to the robustness of the autoencoder
because unstable latent variables do not always cause bad reconstruction accuracy.
The reason that random latent variables do not harm the reconstruction accuracy is that
the decoder pipeline of the network learns to ignore the random neurons by assigning a negligible weight.
This indicates that \emph{unstable} propositions are also ``unused'' and ``uninformative.''

While this issue is similar to \emph{posterior collapse}
\cite{dinh2014collapse,bowman2016generating},
which is also caused by ignored latent variables,
an important difference is that the latter degrades the reconstruction (makes it blurry)
because there are too many ignored variables.
In contrast, the symbol stability issue is significant even if the reconstruction is successful.

\subsection{Addressing the Systemic Uncertainty: Removing the Run-Time Stochasticity}
\label{sec:argmax}

The first improvement we made from the original GS-VAE/BC-VAE is that we can
disable the stochasticity of the network in the test time.
After the training, we replace the Gumbel-softmax activation with
a pure argmax of class probabilities, which makes the network fully deterministic.
\begin{align}
 \GS(\vl) \rightarrow \argmax(\vl). \quad(\vl \in \R^{F\times C})
\end{align}
Again note that we assume $\argmax$ returns a one-hot representation rather than the index of the maximum value.
In this form, there is no systemic uncertainty due to the Gumbel noise.
In Binary Concrete, this is equivalent to using Heaviside step function:
\begin{align}
 \BC(\vl) \rightarrow \function{step}(\vl). \quad(\vl \in \R^{F})
\end{align}
Similarly, there is no systemic uncertainty due to the Logistic noise in this form.

\subsection{Addressing the Statistical Uncertainty: Selecting the Appropriate Prior}
\label{sec:bcvae-prior}

The prior distribution of the original Gumbel-Softmax VAE is a uniform random categorical distribution $\cat(\1/C)$,
and the prior distribution of the original Binary-Concrete VAE is a uniform random Bernoulli distribution $\bern(0.5)$.
By optimizing the ELBO with stochastic gradient descent algorithms,
the KL divergence $\KL(q(\vz|\vx)\Mid p(\vz))$ in the ELBO moves
the encoder distribution $q(\vz|\vx)$ closer to the target distribution (prior distribution) $p(\vz)=\bern(0.5)$.
Observe that this objective encourages the latent representation to be \emph{more random},
as $\bern(0.5)$ is the most random distribution among $\bern(p)$ for all $p$.

\begin{figure}[ht]
 \centering
 \includegraphics{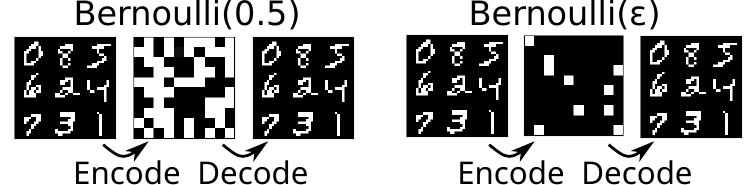}
 \caption{
Autoencoding results of an MNIST 8-puzzle state using a vanilla BC-VAE and
a proposed, non-uniform Bernoulli prior with 100 propositions.
The latter obtains a sparse/compact representation with fewer true bits.
}
 \label{fig:zsae-overview}
\end{figure}

To tackle the instability of the propositional representation,
\emph{we propose a different prior distribution $p(\vz)$ for the latent random variables:} $p(\vz)=\bern(0)$ (\refig{fig:zsae-overview}).
Its key insight is to penalize the latent propositions for unnecessarily becoming true
while preserving the propositions that are absolutely necessary for maintaining the reconstruction accuracy.
In practice, however, since computing the KL divergence with $p(\vz)=\bern(0)$
causes a division-by-zero error in the logarithm, we instead use $p(\vz)=\bern(\epsilon)$
for a small $\epsilon$ (e.g., 0.1, 0.01).

We formalize the notions above as follows.
Consider a case where $\vz$ is a 1-dimensional scalar $z$.
The KL divergence $\KL(q(z|\vx)\Mid p(z))$ with $p(z)=\bern(0.5)$
and $q=q(z=1|\vx)=\sigmoid(\encode(\vx))$ is as follows (\refsec{sec:discrete-vae}):
\begin{align*}
\sum_{k\in\{0,1\}} q(z=k|\vx) \log\frac{q(z=k|\vx)}{p(z=k)} =
q\log\frac{q}{0.5}+(1-q)\log\frac{(1-q)}{(1-0.5)} &=-H(q) + \log 2.
\end{align*}
$H(q)$ is an \emph{entropy} of the probability $q$, which models the randomness of $q$.
Therefore, minimizing the KL divergence is equivalent to maximizing the entropy,
which encourages the latent vector $z\sim q(z|\vx)$ to be more random.
The KL divergence with $p(z)=\bern(\epsilon)$
results in a form
\begin{align}
q\log\frac{q}{\epsilon}+(1-q)\log\frac{(1-q)}{(1-\epsilon)} &=-H(q) + \alpha \cdot q - \log (1-\epsilon) \label{eq:zsae}
\end{align}
for $\alpha=\log (1-\epsilon)-\log \epsilon$.
When $\alpha > 0 \ (\epsilon<\frac{1}{2})$,
this form shows that minimizing the KL divergence
has an additional regularization pressure to move $q$ toward 0.

Intuitively, this optimization objective assumes a variable to be false
when there is not enough ``evidence'' from the input image to flip a variable to true.
The evidence in the data detected as a combination of complex patterns by the neural network
declares some variables to be true if the evidence is strong enough.
This resembles \emph{closed-world assumption} \cite{reiter1981closed}
where propositional variables (generated by grounding a set of first order logic formulae with terms)
are assumed to be false unless it is explicitly declared or proven to be true.
The difference is whether the declaration is implicit (performed by the encoder using the input data)
or explicit (encoded manually by human, or proved by a proof system).

In contrast, the optimization objective made by the original prior distribution $p(z)=\bern(0.5)$
makes a latent proposition highly random when there is neither the evidence for it to be true nor the evidence for it to be false.
Since a uniform binary distribution $\bern(0.5)$ carries no information about either the truthfullness or the falsifiability of the proposition,
thus a variable having this distribution can be seen as in an ``unknown'' state.
Again, this resembles an open-world assumption where
a propositional variable that is neither declared true or false are put in an unknown state.

\dnote{
We discuss the difference from an ad-hoc method presented in the conference version \citep{Asai2019a}
in appendix \refsec{sec:zsae-conference-version-difference}.
}

\section{\texorpdfstring{AMA$_1$}{AMA1}: An SAE-based Translation from Image Transitions to Ground Actions}
\label{sec:ama1-overview}

An SAE by itself is sufficient for a minimal approach to generating a PDDL definition for a grounded 
STRIPS planning problem from image data.
Given a pair of images $\xtr=(\xboth)$ which represents a valid transition in the domain,
we can apply the SAE to $\rxbefore$ and $\rxafter$ to obtain $\zbefore$ and $\zafter$,
the latent-space propositional representations of these states.

The transition $(\zboth) \in \Ztr$ can be directly translated to an action $\action$ as follows:
each bit $\zbefore[i,j]_f$ ($f\in 1..F, j\in 0..1$) in a boolean vector $\zbefore[i,j]$ is
mapped to a proposition \texttt{(z$_f$)} when the value is 1,
or to its negation \texttt{(not (z$_f$))} when the value is 0.
The elements of a bitvector $\zbefore$ are directly used as the preconditions of action $\action$ using negative precondition extension of STRIPS.
The add/delete effects of the action are obtained from the bitwise differences between $\zbefore$ and $\zafter$.
For example, when $(\zbefore, \zafter)=(0011,0101)$, 
the action definition would look like
\begin{minted}{common-lisp}
 (:action action-0011-0101
  :preconditions (and (not (z0)) (not (z1)) (z2) (z3))
  :effects       (and (z1) (not (z2))))
\end{minted}

Given a set of image pairs $\Xtr$,
AMA$_1$ directly maps each image transition to a STRIPS action as described above.
The initial and the goal states are similarly created by applying the SAE to the initial and goal images
and encoding them into a PDDL file, which can be input to an off-the shelf planner.
{\it If $\Xtr$ contains images for a sufficient portion of the state space containing a path from the initial and goal states}, then a planner can find a satisficing solution. If $\Xtr$ includes image pairs for {\it all} valid transitions, then an optimal solution can be found using an admissible search algorithm.
However, collecting such a sufficient of transitions $\Xtr$ is impractical in most domains.
Thus, AMA$_1$ is of limited practical utility, as it only provides the planner with a grounded model of state transitions which have already been observed -- this lacks the key ability to generalize from grounded observations and predict/project  {\it previously unseen} transitions, i.e., AMA$_1$ lacks the ability to generate more general action schemas.

\section{\texorpdfstring{AMA$_2$}{AMA2}: Action Symbol Grounding with an Action Auto-Encoder (AAE)}
\label{sec:ama2}

\begin{figure}[tbp]
 \centering
 \includegraphics[width=\linewidth]{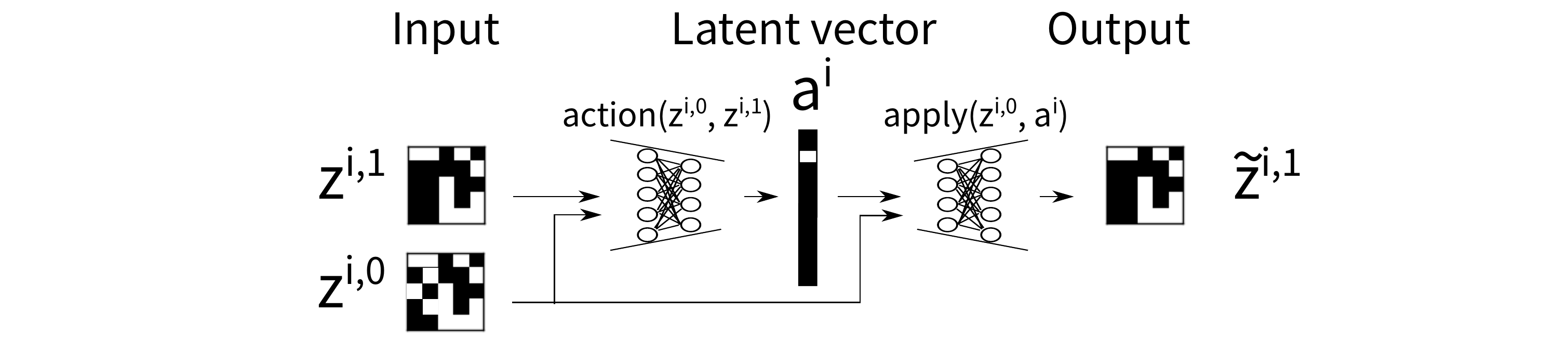}
 \caption{Action Autoencoder.}
 \label{fig:aae}
\end{figure}

Next, we propose an unsupervised approach to acquiring an action model from a limited set of examples (image transition pairs).
AMA$_2$ contains a neural network named \emph{Action Autoencoder} (AAE, \refig{fig:aae}).
The AAE jointly learns action symbols and the corresponding effects
and provides an ability to enumerate the successors of a given state, i.e., it can be used as a black-box successor function for a forward state space search algorithm.
The key idea is that the successor predictor can be trained along with the action predictor by a variant of Gumbel-Softmax VAE.
\begin{itemize}
 \item The encoder $\aaee(\zbefore,\zafter)$ samples an action label $\action$ for a state transition $(\zbefore,\zafter)$.
 \item The decoder $\aaed(\action,\zbefore)$ applies $\action$ to $\zbefore$ and samples a successor $\zafterrec$.
       Thus the decoder network can be seen as modeling the \emph{effect} of each action,
       which represents what change should be made to $\zbefore$ due to the application of the action.
 \item It takes a concatenated vector $\zbefore;\zafter$ of a state transition pair $(\zbefore,\zafter)$
       and returns a reconstruction $\zafterrec$.
       The output layer is activated by a sigmoid function.
       The training is performed by minimizing the loss $\loss(\zafter,\zafterrec) + \KL$.
       The reconstruction loss $\loss$ is a Binary Cross Entropy loss
       because predicting a binary successor state is equivalent to multinomial classification.
       The KL divergence is a standard one for Gumbel-Softmax.
 \item Unlike typical Gumbel-Softmax VAEs, which contain multiple ($n$-way) one-hot vectors in the latent space,
       AAE has a single (1-way) one-hot vector of $A$ classes, representing an \textbf{action label} $\action\in \B^A$.
       $A$ is a hyperparameter for the maximum number of action labels to be learned.
\end{itemize}

The number of labels $A$ serves as an upper bound on the number of action symbols learned by the network.
We tend to use a large value for $A$
because
(1) too few labels make AAE reconstruction loss fail to converge to zero, and
(2) even if a large $A$ is given,
AAE tends to use the available set of labels $1..A$ frugally and leave the excess labels unused.
We remove those unused labels after the training by iterating through the dataset
and checking which labels were used.

Our earlier conference paper showed that AMA$_2$ could be used to successfully learn an action model for several image-based planning domains, and that a forward search algorithm using AMA$_2$ as the successor function could be used to solve planning instances in those domains \cite{Asai2018}.

AMA$_2$ provides a non-descriptive, black-box neural model as the successor generator, instead of a descriptive, symbolic model (e.g., PDDL model).
A black-box action model can be useful for exploration-based search algorithms such as Iterated Width \cite{frances2017purely}.
On the other hand, the non-descriptive, black-box nature of AMA$_2$ has a drawback:
The lack of a descriptive model prevents the use of existing, goal-directed heuristic search techniques,
which are known to further improve the performance when combined with exploration-based search \cite{lipovetzky2017bwfs}.

To overcome this problem, we could try to translate/compile such a black-box model into a descriptive model usable by a standard symbolic problem solver.
However, this is not trivial. Converting an AAE learned for a standard STRIPS domain to PDDL can result in a PDDL file which is several orders of magnitude larger than typical PDDL benchmarks.
This is because the AAE has the ability to learn an arbitrarily complex state transition model.
The traditional STRIPS progression $\aaed(s,a)= (s \setminus \dele(a)) \cup \adde(a)$
\emph{disentangles} the effects from the current state $s$,
i.e., the effect $\adde(a)$ and $\dele(a)$ are defined entirely based on action $a$.
In contrast, the AAE's black-box progression $\aaed(\action,\zbefore)$ does not offer such a separation,
allowing the model to learn arbitrarily complex conditional effects.
For example, we found that a straightforward logical translation of the AAE with a rule-based learner
(e.g., Random Forest \cite{ho1998random}) results in a PDDL that cannot be processed by modern classical planners
due to the huge file size and exponential compilation of disjunctive formula \cite{Asai2020b}.

\section{\texorpdfstring{AMA$_3$/AMA$_3^+$}{AMA3/AMA3+}: Descriptive Action Model Acquisition with Cube-Space AutoEncoder}
\label{sec:ama3}

We now propose the \emph{Cube-Space AutoEncoder} architecture (CSAE),
which can generate a standard PDDL model file from a limited set of observed image transitions for domains which conform to STRIPS semantics.
There are two major technical advances from AMA$_2$.
First, CSAE jointly learns a state representation and an action model, whereas AMA$_2$ learns a state representation first and learns the action model on the fixed state representation.
As often pointed out in the machine learning literature, the joint learning is expected to improve the performance.
Second, CSAE constrains the action model to the STRIPS semantics.
Thus we can trivially extract STRIPS/PDDL-compatible action effects from the trained CSAE,
providing a grounded PDDL that is immediately usable by off-the-shelf planners.
We implement the restriction as a structural prior that replaces the MLP decoder $\aaed$ in the AAE.
Due to the joint training, the state space shapes itself to satisfy the constraints of the action model.

We will refer to the previously published version of Cube-Space AE \cite{Asai2020} as AMA$_3$,
and the updated version presented here as AMA$_3^+$.
The architecture and the objective function of AMA$_3$ were neither theoretically analyzed in \cite{Asai2020},
nor were designed with theoretical justification in mind.
AMA$_3^+$ revises AMA$_3$ by providing a theoretically sound optimization objective and architecture.

This section proceeds as follows:
In \refsec{sec:vanilla-space-ae},
we introduce Vanilla Space AutoEncoder, which performs a joint training of the state and action models in AMA$_2$,
but without the novel structure in CSAE that restricts the action model.
We define its training objective which is theoretically justified as
a lower bound of the log likelihood of observing the data.
In \refsec{sec:cube-space-ae},
we formally introduce CSAE by inserting a structural prior to Vanilla Space AE.
\refsec{sec:eq-cube-btl} shows that the CSAE generates a STRIPS model.
\refsec{sec:ama3-effect-extraction} briefly discuss how to extract the action effects and preconditions from CSAE.
In the following, we always omit the dataset index $^i$.

\subsection{Vanilla Space AutoEncoder}
\label{sec:vanilla-space-ae}

\renewcommand{\defaultindex}{}
\renewcommand{\defaultcomma}{}

\begin{figure}[tbp]
 \centering
 \includegraphics[width=\linewidth]{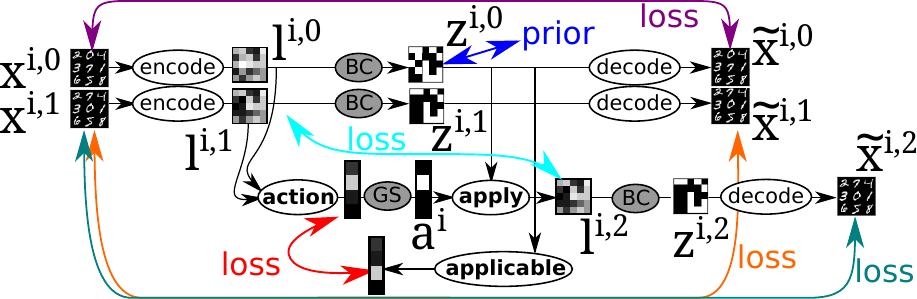}
 \caption{AMA$_3^+$, a vanilla Space AutoEncoder architecture that we propose in this paper.
The inputs for $\aaee$ can be any deterministic value computed from $\xboth$ (including $\xboth$ itself),
and we use $\lbefore,\lafter$ because some features are supposed to be already extracted in them.
Each color corresponds to each formula in \refeq{eq:vanilla-elbo8-main}.
}
 \label{fig:ama3}
\end{figure}

\begin{figure}[tbp]
\begin{align}
 \text{(input)}                                         & & \xbefore, \xafter & &                               \nn
 \text{(encoded logits)}                       & & \lbefore[k]    & = \encode(\xbefore[k]).     & k \in \braces{0,1} \nn
 \text{(sampled binary representations)}       & & \zbefore[k]    & = \BC(\lbefore[k]).         & k \in \braces{0,1} \nn
 \text{(logit for actions)}                    & & \laction       & = \aaee(\lbefore, \lafter). &                    \nn
 \text{(sampled discrete action label)}        & & \action        & = \GS(\laction).            &                    \nn
 \text{(logit for the result of progression)}  & & \lafteralt     & = \aaed(\zbefore, \action)  &                    \nn
 \text{(sampled binary result of progression)} & & \zafteralt     & = \BC(\lafteralt).          &                    \nn
 \text{(reconstructions)}                      & & \xbeforerec[k] & = \decode(\zbefore[k]).     & k \in \braces{0,1,2}
 \label{fig:vanilla-network-definition}
\end{align}
\begin{align}
 \log p(\xboth) \geq\ &\text{ELBO}(\xboth)\nn
 =& \log \violet{p(\xbefore \mid \zbefore)} + \frac{1}{2}\log \orange{p(\xafter\mid\zafter)} + \frac{1}{2}\log \teal{p(\xafter\mid\zafteralt)} &\text{(Reconstruction losses)}      \nn
 & - \beta_1\KL(\blue{q(\zbefore\mid\xbefore) \Mid p(\zbefore)})                                                                              &\text{(Stabilize \refex{example:stability-state}.)}       \nn
 & - \beta_2\KL(\red{q(\action\mid\xboth) \Mid p(\action\mid\zbefore)})                                                                       &\text{(Action must be applicable.)}        \nn
 & - \frac{1}{2}\beta_3\KL(\cyan{q(\zafter\mid\xafter) \Mid p(\zafteralt\mid\zbefore,\action)}).                                              &\text{(Stabilize \refex{example:stability-effect}.)}
 \label{eq:vanilla-elbo8-main}
\end{align}
\begin{align}
  & p(\zbefore)  = \bern(\epsilon). & & \text{\refsec{sec:bcvae-prior}.} \nn
  & p(\action \mid \zbefore) = \cat(\aaep(\zbefore)).& & \nn
  & p(\xbefore[k]\mid \zbefore[k]) = \mathcal{N}(\xbefore[k]\mid\xbeforerec[k],\sigma). & & k \in \braces{0,1}        \nn
  & p(\xafter\mid \zafteralt) = \mathcal{N}(\xafter\mid\xafteraltrec,\sigma). & &         \nn
  & p(\zafteralt \mid \zbefore, \action) = \bern(\qafteralt),\ \text{where}\ \qafteralt=\sigmoid(\lafteralt).
 \label{eq:ama3-generative-main}
\end{align}
\begin{align}
  & q(\zbefore[k] \mid \xbefore[k]) = \bern(\qbefore[k]), \text{where}\ \qbefore[k] = \sigmoid(\lbefore[k]). && k \in \braces{0,1} \nn
  & q(\action  \mid \xbefore,\xafter)   = \cat (\qaction), \text{where}\ \qaction  = \softmax(\laction).
 \label{eq:ama3-variational-main}
\end{align}
\caption{Main data pipeline, ELBO, generative model, and variational model of \ama3.}
\end{figure}

Vanilla Space AE (\refig{fig:ama3}) is a network representing a function of two inputs $\xboth$ and three outputs $\xbothrec,\xafteraltrec$.
It consists of several subnetworks: $\encode,\decode,\aaee,\aaed,\aaep$.
$\aaep$ is a new subnetwork that was missing AMA$_3$
and is necessary for using a theoretically justifiable optimization objective (ELBO).
The data pipeline from the input to the output is defined in \refig{fig:vanilla-network-definition}.
$\aaep$ is not used in this main pipeline because
it is used only for regularizing the network during the training, and is not used in the test time
where we generate a PDDL output.
The network is trained by maximizing $\text{ELBO}(\xboth)$ defined in \refeq{eq:vanilla-elbo8-main},
where each probability is obtained from the corresponding neural network
as defined in the generative model (\refeq{eq:ama3-generative-main}) and the variational model (\refeq{eq:ama3-variational-main}).

\paragraph{Theoretical Property}

Unlike AMA$_3$, the loss function $\text{ELBO}(\xboth)$ is theoretically justified as
the lower bound of the log likelihood $\log p(\xboth)$ of observing a pair of images $(\xboth)$,
following Maximum Likelihood Estimation (MLE) and variational inference framework (\refsec{sec:aevae}).
An important implication of this theoretical property is that,
\emph{in theory}, when $p(\xboth)$ converges to the ground-truth distribution,
Latplan never generates visualizations containing invalid states and invalid transitions.
In the ground-truth distribution,
$p(\xboth)=0$ if $(\xboth)$ is an invalid transition, or if either $\xbefore$ or $\xafter$ are invalid states.
MLE achieves this by maximizing $p(\xboth)$ for real data,
which reduces $p(\xboth)$ for invalid data
because a probability distribution sums/integrates into 1 ($\int p(\xboth)d(\xboth)=1$).
Thus, following the MLE framework, which is the current dominant theoretical foundation of machine learning models,
converging a lower bound (ELBO) of the log likelihood $\log p(\xboth)$ to the ground-truth distribution
guarantees the correctness of the learned results as well as the planning results (due to the soundness of the planner being used).

In practice, this convergence is not achieved
and visualized plans do not have a correctness guarantee in a traditional sense.
However, note that this is not necessarily a disadvantage of our paradigm,
since the use case of this system is where human supervision and manual modeling are not available.
Moreover, even human could write a PDDL model with a bug that results in an incorrect modeling.

\paragraph{Interpretation}

\refeq{eq:vanilla-elbo8-main} is not only theoretically justified,
but also has a clear interpretation.
The first three terms represent reconstruction losses for $\zboth$.
The remaining KL divergence terms, which regularizes the training, have the following interpretations:
$\KL(\blue{q(\rzbefore\mid\rxbefore) \Mid p(\rzbefore)})$
addresses the stability of propositional symbols learned by SAE due to noise,
as discussed in \refsec{sec:unstable}, \refex{example:stability-state}.
$\KL(\red{q(\raction\mid\rxboth) \Mid p(\raction\mid\rzbefore)})$
addresses applicability of actions (precondition learning).
$\KL(\cyan{q(\rzafter\mid\rxafter) \Mid p(\rzafter\mid\rza)})$
addresses effects of actions and the stability of propositional symbols due to effects,
as discussed in \refsec{sec:unstable}, \refex{example:stability-effect}.

$\KL(\red{q(\raction\mid\rxboth) \Mid p(\raction\mid\rzbefore)})$
compares $q(\raction\mid\rxboth)$ and $p(\raction\mid\rzbefore)$, which are outputs of $\aaee$ and $\aaep$, respectively.
$q(\raction\mid\rxboth)$ is a distribution of an action predicted by observing the states both before and after the transition.
In contrast,
$p(\raction\mid\rzbefore)$ is a distribution predicted without observing the result (successor state) of the action.
The latter is thus intrinsically ambiguous, and returns a distribution of \emph{applicable} actions (hence the name of the network).
The former, in contrast, can be seen as a distribution of the \emph{actual} action that happened,
having access to the consequence of the action.

By minimizing the KL divergence on the training data, two things happen:
First, the distribution of the actual action ($q(\ldots)$) gets closer to applicable actions ($p(\ldots)$) in each transition.
Second, the distribution of applicable actions ($p(\ldots)$) is stretched toward various actual actions ($q(\ldots)$)
by multiple transitions with the same $\rzbefore$ and different $\rzafter$.
For example, $p(\ldots)$ could be $1/3$ for each of three applicable actions, while $q(\ldots)$ is one-hot.
As a result, informally speaking, the KL divergence is forcing each actual action to be in a set of applicable actions.

$\KL(\cyan{q(\rzafter=\zafter\mid\rxafter) \Mid p(\rzafter=\zafteralt\mid\rza)})$
compares $q(\rzafter\mid\rxafter)$ and $p(\rzafter\mid\rza)$, which are both distributions of a latent successor state $\rzafter$.
$p(\rzafter\mid\rza)$ is a distribution of the result of symbolically \emph{applying} an action to a latent current state (thus is an output of $\aaed$),
and $q(\rzafter\mid\rxafter)$ is a distribution of the latent successor state obtained from an observation.
By matching them, we maintain the stability of the latent propositions and the correctness of the effects at the same time.

To reduce the length of the paper,
more in-depth discussion on the design decision and the derivation of this loss function was moved to the appendix (\refsec{sec:loss-derivation}).
Readers who are not familiar with machine learning may find the section useful
because we describe not only our mathematical derivation,
but also the general strategy of statistical modelling
which allows one to derive a machine learning model with a principled theoretical justification.

\renewcommand{\defaultindex}{i}
\renewcommand{\defaultcomma}{,}

\subsection{Cube-Like Graph}

The state space modeled by AMA$_2$, and Vanilla Space AE as its end-to-end version,
are not specifically designed for generating a compact STRIPS action model,
therefore its STRIPS compilation may be exponentially large \cite{Asai2020b}.
To illustrate this issue,
we consider the problem of assigning action labels to each edge in the latent space
based on node embedding differences caused by action effects.
This yields a graph class called \emph{Cube-Like Graph}, which inspired the name of our new architecture:

\begin{defi}
A \emph{cube-like graph} \citep{payan1992chromatic} is a simple\footnote{No duplicated edges between the same pair of nodes}
undirected graph $G(S,D)=(V,E)$ defined by sets $S$ and $D$.
Each node $v\in V$ is a finite subset of $S$, i.e., $v\subseteq S$.
The set $D$ is a family of subsets of $S$,
and for every edge $e = (v,w) \in E$, the symmetric difference between the connected nodes belongs to $D$,
i.e., $d = v\oplus w = (v\setminus w) \cup (w\setminus v) \in D$.
\label{def:cube-like}
\end{defi}

\begin{figure}[htb]
 \centering
 \includegraphics[width=\linewidth]{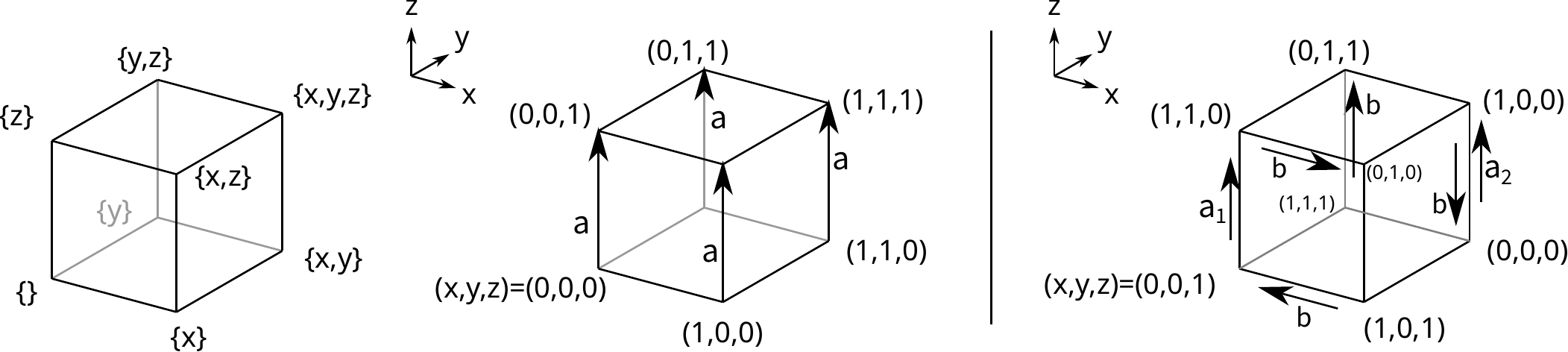}
 \caption{
(Left) A graph representing a 3-dimensional cube can be made into a cube-like graph by assigning a set to each vertex.
(Center) Its bit-vector representation.
(Right) The same graph with a randomly shuffled node embedding.
}
 \label{fig:cube-like}
\end{figure}

For example, a unit cube becomes a cube-like graph if we assign a set to each vertex appropriately as in \refig{fig:cube-like},
i.e.,
\begin{align*}
S&=\braces{x,y,z},&
V&=\braces{\emptyset,\braces{x},\ldots \braces{x,y,z}},\\
E&=\braces{(\emptyset,\braces{x}),\ldots (\braces{y,z},\braces{x,y,z})},&
D&=\braces{\braces{x},\braces{y},\braces{z}}.
\end{align*}
The set-based representations can alternatively be represented as bit-vectors, e.g.,
\begin{align*}
    V=\braces{(0,0,0),(0,0,1),\ldots (1,1,1)}.
\end{align*}

We observed a similarity between STRIPS action models and cube-like graphs,
with the difference being that STRIPS action models consider a directed version, i.e.,
for every edge $e = (v,w) \in E$,
their asymmetric differences $(d^+,d^-)=(w \setminus v, v \setminus w)$
satisfy $w=(v \setminus d^-) \cup d^+$,
which is typically denoted as $w=(v \setminus \dele(a)) \cup \adde(a)$.

Consider the number of actions required in two node embeddings in \refig{fig:cube-like}.
The first embedding (left and center) requires 3 labels (6 labels if directed),
where each label is assigned to 4 parallel edges which share the same node embedding differences.
The set of node embedding differences corresponds to the set $D$,
and each element of $D$ represents an action,
such as $a$ whose difference is $(1,0,0)$ in the figure.

In contrast, the graph on the right has node embeddings that are randomly shuffled.
Despite having the same topology and the same embedding size,
this graph lacks the patterns we saw on the left,
thus it needs more action labels,
i.e., it lacks a compact STRIPS action model.
For example,
action $a_1$ represents a difference $(1,1,0)-(0,0,1)=(1,1,-1)$ and
action $a_2$ represents $(1,0,0)$, requiring different effects.
While some edges may share the effects,
such as the action $b$ with difference $(-1,0,0)$,
there are 9 differences in this graph, thus it needs 9 actions (18 if directed).

\subsection{Cube-Space AE (CSAE)}
\label{sec:cube-space-ae}

\renewcommand{\defaultindex}{}
\renewcommand{\defaultcomma}{}

Based on this intuition, we named our new architecture Cube-Space AE.
Cube-Space AE modifies the $\aaed$ network so that it directly predicts the effects
 without taking the current state as the input and logically computes the successor state
based on the predicted effect and the current state.

\begin{figure}[htbp]
 \centering
 \includegraphics[width=0.8\linewidth]{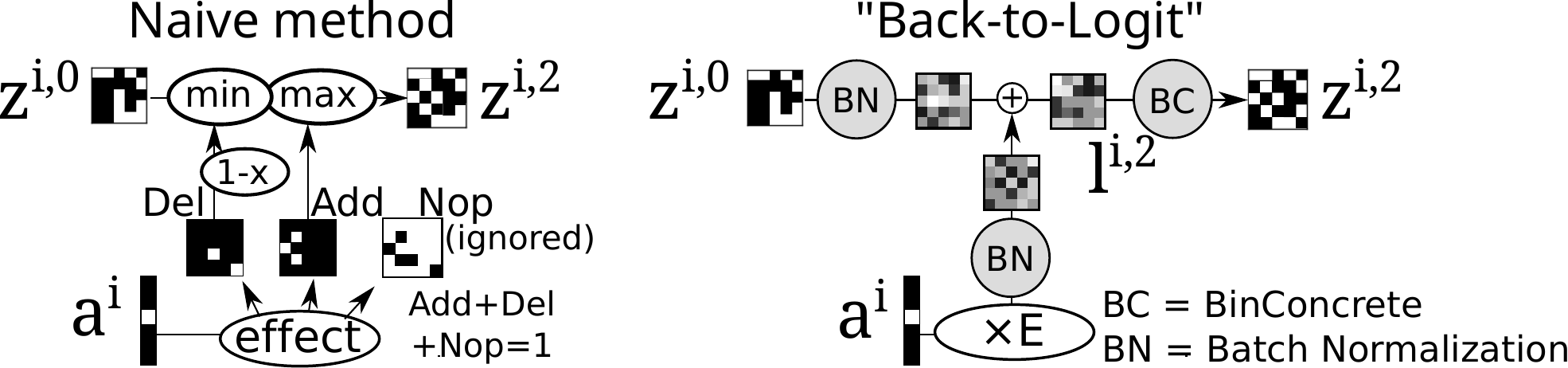}
 \caption{A naive and a Back-to-Logit implementation of the $\aaed$ module of the CSAE.}
 \label{fig:cube}
\end{figure}

\paragraph{Naive implementation:}
To illustrate the idea, we first present
an ad-hoc, naive and impractical implementation of such a network in \refig{fig:cube} (left),
which directly simulates STRIPS action application with vector operations.
The \function{effect} network predicts a binary tensor of shape $F\times 3$
using $F$-way Gumbel-Softmax of 3 categories.
Each Gumbel Softmax corresponds to one bit in the $F$-bit latent space
and 3 classes correspond to the add effect, delete effect, and NOP,
only one of which is selected by the one-hot vector.
The effects are applied to the current state
by max/min operations.
Formally, the naive CSAE is formulated as follows:
\begin{align}
 \B^F\ni\zafteralt =&\max (\min (\zbefore,1-\dele(\action)),\adde(\action)),\quad\text{where}
\end{align}
\begin{align}
 \effect(\action) &=\GS(\text{MLP}(\action)) \in \B^{F\times 3}, & \adde(\action)          &=\effect(\action)_0, \nn
 \dele(\action)   &=\effect(\action)_1,                          & \function{nop}(\action) &=\effect(\action)_2.
\end{align}

While intuitive, we found these naive implementations extremely difficult to train in AMA$_3$.
Moreover, $\zafteralt$ in these models is not produced by Binary Concrete using a logit $\lafteralt$,
making it impossible to compute the KL divergence
$\KL(\cyan{q(\zafter\mid\xafter) \Mid p(\zafteralt\mid\zbefore,\action)})$
where $q(\zafter\mid\xafter)=\sigmoid(\lafter)$ and $p(\zafteralt\mid\zbefore,\action)=\sigmoid(\lafteralt)$,
which is required by the theoretically motivated ELBO objective of AMA$_3^+$.
We thus abandoned this idea in favor of a theoretically justifiable alternative.

\paragraph{Back-to-Logit:}
Our contribution to the architecture is \emph{Back-to-Logit} (BTL, \refig{fig:cube}, right),
\emph{a generic approach that computes a logical operation in the continuous logit space}.
We re-encode a logical, binary vector back to a continuous, logit representation
by an element-wise monotonic function $m$.
This monotonicity preserves the order between true (1) and false (0) even after transformed into real numbers.
We then apply the given action by adding a \emph{continuous effect vector} to the continuous current state.
The effect vector is produced by applying an MLP named $\function{effect}$ to the action vector $\action$.
After adding the continuous vectors, we re-discretize the result with Binary Concrete.
Formally,
\begin{align}
 \zafteralt = \BC(\aaed(\za)) = \BC(m(\zbefore)+\effect(\action)). \label{eq:btl}
\end{align}

\paragraph{Batch Normalization as $m$:}
We found that an easy and successful way to implement $m$
is \emph{Batch Normalization} \cite{ioffe2015batch}, a method that was originally developed for
addressing \emph{covariate shift} in deep neural networks.
For simplicity, we consider a scalar operation, which can be applied to vectors element-wise.
During the batched training of the neural network,
Batch Normalization layer $\BN(x)$ takes a minibatch input $B=\braces{x^{1}\ldots x^{|B|}}$,
computes the mean $\mu_B$ and the variance $\sigma_B^2$ of $B$,
then shifts and scales each $x^i$ so that the resulting batch has a mean of 0 and a variance of 1.
It then shifts and scales the results by two trainable scalars $\gamma$ and $\beta$, which are shared across different batches.
Formally,
\begin{align}
 \forall x^i\in B;\ \BN(x^i) = \frac{x^i-\mu_B}{\sigma_B} \gamma + \beta. \label{eq:batchnorm}
\end{align}
After the training, $\BN$ behaves deterministically.
At the end of the training,
$\mu_B$ and $\sigma_B$ are set to the statistics of the entire training dataset $\mathcal{X}$, i.e.,
$\mu_\mathcal{X}$, $\sigma_\mathcal{X}$.

\paragraph{Final definition:}
Furthermore, since $\action$ is a probability vector over $A$ action ids and
$\action$ eventually converges to a one-hot vector due to Gumbel-Softmax annealing,
the additional MLP can be merely a linear embedding, i.e., $\effect(\action)=\mE\action$, where $\mE \in \R^{F\times A}$.
It also helps the training if we apply batch normalization on the effect vector. Therefore, a recommended implementation is:
\begin{align}
\zafteralt = \BC(\aaed(\za)) = \BC(\BN(\zbefore)+\BN(\mE\action)).  \label{eq:btl-implementation}
\end{align}

Similar to the original Vanilla Space AE,
we can interpret it from a Bayesian generative modeling perspective as well.
We discuss the interpretation in Appendix \refsec{sec:btl-bayesian-interpretation}.

\renewcommand{\defaultindex}{i}
\renewcommand{\defaultcomma}{,}

\paragraph{Equivalence to STRIPS:}
\label{sec:eq-cube-btl}

Consider an ideal training result where $\forall i; \zafter\equiv\zafteralt$, i.e.,
\renewcommand{\defaultindex}{}%
\renewcommand{\defaultcomma}{}%
\[
 \KL(\cyan{q(\rzafter=\zafter\mid\xafter) \Mid p(\rzafter=\zafteralt\mid\zbefore,\action)})=0.
\]
\renewcommand{\defaultindex}{i}%
\renewcommand{\defaultcomma}{,}%

States learned by BTL have the following property:
\begin{theo}
For state transitions $\zboth$ whose action is $\action[]$,
the transitions satisfy the following conditions.
For each bit $j$, it is either:
\begin{align}
 (\text{add:})\ & \forall i; (\zbefore_j,\zafter_j)\in \braces{(0,1), (1,1)},\\
\text{or}\ (\text{del:})\ & \forall i; (\zbefore_j,\zafter_j)\in \braces{(1,0), (0,0)},\\
\text{or}\ (\text{nop:})\ & \forall i; (\zbefore_j,\zafter_j)\in \braces{(0,0), (1,1)}.
\end{align}
\label{thm:monotonicity}
\end{theo}

In other words, they are bitwise monotonic ($\zbefore_j\leq\zafter_j$, $\zbefore_j=\zafter_j$, or $\zbefore_j\geq\zafter_j$),
deterministic ($\zbefore_j \mapsto \zafter_j$ is a function),
and these three modes never mix up for the same action.
The proof is straightforward from the monotonicity of $m$ and Binary Concrete (See Appendix \refsec{sec:proof-monotonicity}).

This theorem guarantees that
each action deterministically sets a certain bit on and off in the binary latent space.
Therefore, the actions and the transitions satisfy
the STRIPS state transition rule $s' = (s \setminus \dele(a)) \cup \adde(a)$,
thus enabling a direct translation from neural network weights to PDDL modeling language,
as depicted in \refig{fig:15puzzle}.

\paragraph{Effect extraction:}
\label{sec:ama3-effect-extraction}

To extract the effects of an action $\action[]$ from CSAE,
we compute $\function{add}(\action[])=\aaed(\action[],\0)$ and $\function{del}(\action[])=1-\aaed(\action[],\1)$
for each action $\action[]$,
where $\0,\1 \in \B^F$ are vectors filled by zeros/ones and has the same size as the binary embedding.
Since $\aaed$ deterministically sets values to 0 or 1, feeding these vectors is sufficient to see which bit it turns on and off.
For each $j$-th bit that is 1 in each result, a corresponding proposition is added to the add/delete-effect, respectively.

There is one difference between theory and practice.
As mentioned in \refeq{eq:batchnorm}, batch normalization has a trainable parameter $\gamma$ for each dimension.
This $\gamma$ must be positive in order for a batch normalization layer to be monotonic.
The value of $\gamma$ is typically initialized by a positive constant (e.g., 1 in Keras)
and typically stays positive during the training.
However, in rare cases, it is possible that $\gamma$ turns negative in some bits, which requires special handling.

We call this monotonicity violation as XOR semantics,
as the resulting bitwise value follows a pattern $\braces{(0,1), (1,0)}$ that is not allowed under STRIPS.
Detecting XOR semantics from the extracted effects is easy
because these problematic bits have both add- and delete-effects in the same action.
We compile them away by splitting the action into two versions, appending appropriate preconditions.
While the compilation is exponential to the number of violations,
violations are usually rare. We will report the number of violations in the empirical evaluation.

\paragraph{Precondition extraction:}
To extract the preconditions of an action $\action[]$,
we propose a simple ad-hoc method that is similar to \cite{Wang94} and \cite{amado2018goal}.
It looks for bits that always have the same value when $\action[]$ is used.
Let $\mathcal{Z}^0(\action[])$ be the set of states that we can observe the usage of an action $\action[]$ in the dataset,
i.e., $\braces{\zbefore \mid (\zboth)\in\Ztr; \argmax(\aaee(\zbefore,\zafter))=\action[]}$. Then
the positive and the negative preconditions of an action $\action[]$ are defined as follows:
\begin{align}
 \posp(\action[])=\{f\mid\forall \zbefore \in \mathcal{Z}^0(\action[]);\zbefore_f=1\}, \negp(\action[]) = \{f\mid\forall \zbefore \in \mathcal{Z}^0(\action[]); \zbefore_f=0\}.
\end{align}

This ad-hoc method suffers from a serious lack of accuracy because the propositions whose values differ between different samples in $\mathcal{Z}^0(\action[])$ are
always treated as ``don't care'' even if they could consist of several  patterns.
Suppose the ground-truth generator of $\mathcal{Z}^0(\action[])$ is a random vector $\rvz = [0, \ra, \lnot \rb, \lnot \rb, \rb, 1, \rc]$
of independent Bernoulli(0.5) random variables $\ra, \rb, \rc$ and constants 0, 1.
The random vector can be alternatively represented by a logical formula
$\lnot \rvz_0 \land \rvz_5 \land (( \lnot \rvz_2 \land \lnot \rvz_3 \land \rvz_4 ) \lor ( \rvz_2 \land \rvz_3 \land \lnot \rvz_4 ))$.
The ad-hoc method will only recognize $\lnot \rvz_0 \land \rvz_5$, failing to detect the disjunctions.

While we could develop a divide-and-conquer method which recursively extracts disjunctive conditions,
DSAMA \cite{Asai2020b} showed that it is impractical.
It learned them with decision trees applied to $\mathcal{Z}^0(\action[])$ obtained from AMA2,
then compiled the tree into a disjunctive precondition in PDDL.
Disjunctive conditions are typically compiled away in modern planners at the cost of exponential blowup in the number of ground actions,
which a deeper and more accurate tree suffers from.
On the other hand, if we stop the recursion at a certain depth to suppress the explosion,
such a shallow decision tree suffers from the limited accuracy.

These issues are naturally caused by the fact that
the network and the state representation are not conditioned to learn a compact action model with conjunctive preconditions.
To address these issues, we propose \ama4, an improved architecture which also restricts the actions to have  fully conjunctive preconditions.

\section{\texorpdfstring{AMA$_4^+$}{AMA4+}: Learning Preconditions as Effects Backward in Time}
\label{sec:ama4}

While AMA$_3^+$ managed to model the STRIPS action effects,
it requires an ad-hoc precondition extraction with limited accuracy because
the CSAE lacks an explicit mechanism for action preconditions.
We address this problem in a \emph{Bidirectional Cube-Space AE} (BiCSAE):
a neural network that casts precondition learning as
\emph{effect learning for regression planning}.
To our knowledge, no existing work has addressed precondition learning in this manner.
Modeling the precondition learning as a type of effect learning requires us to
treat the problem in a time-symmetric manner.
However, it causes a unique issue in that we have an expectation that regressed states are complete.

\subsection{Complete State Regression Semantics}
\label{sec:complete-state-regression}

Regression planning is a group of search methods that solve a classical planning problem
by looking for the initial state backward from the goal \cite{bonet2001planning,alcazar2013revisiting}.
It typically operates on a partial state / condition as a search node,
which specifies values only for a subset of state variables.
For a classical planning problem $\brackets{P,A,I,G}$,
it starts from a node that contains the goal condition $G$,
unlike usual forward search planners, which starts from the initial state $I$.
State transitions on partial states are performed by \emph{regression},
which infers the properties of a state prior to applying an action.

In general, regression produces partial states. This is so, \textit{even if we regress a complete state}. Because of the nature of our learned latent representation, we must maintain this complete state property. Consider
\reftbl{tab:regression}: a list of possible transitions of a propositional variable $p\in\braces{0,1}$
when a predecessor state $s$ transitioned to a successor state $t$ using a STRIPS action.
In the precondition / effect columns,
+ indicates a positive precondition and an add-effect,
- indicates a negative precondition and a delete effect,
0 indicates that the action does not contain either effects,
* indicates that the action does not contain either preconditions.
When the precondition is not specified (*) for a bit that is modified by an effect,
there is ambiguity in the source state, indicated by ``?'' (line 3, 6 in \reftbl{tab:regression})
--- the previous value could be either 0 or 1.
In other words, regressing such an action is non-deterministic,
and the default STRIPS semantics for unspecified precondition is disjunctive (true or false).

\begin{table}[htbp]
 \centering
 \begin{minipage}[t]{0.48\linewidth}
 \begin{tabular}{ccccc}
    & $s$   & precondition & effect & $t$   \\\toprule
  0 & $p=1$ & +            & +      & $p=1$ \\
  1 & $p=1$ & +            & 0      & $p=1$ \\
  2 & $p=1$ & +            & -      & $p=0$ \\\midrule
  \rowcolor{gray}
  3 & $p=?$ & *            & +      & $p=1$ \\
  4 & $p=0$ & *            & 0      & $p=0$ \\
  5 & $p=1$ & *            & 0      & $p=1$ \\
  \rowcolor{gray}
  6 & $p=?$ & *            & -      & $p=0$ \\\midrule
  7 & $p=0$ & -            & +      & $p=1$ \\
  8 & $p=0$ & -            & 0      & $p=0$ \\
  9 & $p=0$ & -            & -      & $p=0$ \\\bottomrule
 \end{tabular}
 \end{minipage}
 \begin{minipage}[t]{0.48\linewidth}
 \begin{tabular}{ccccc}
    & $s$   & precondition & effect & $t$   \\\toprule
  0 & $p=1$ & +            & +      & $p=1$ \\
  1 & $p=1$ & +            & 0      & $p=1$ \\
  2 & $p=1$ & +            & -      & $p=0$ \\\midrule\rowcolor{gray}
  3 & $p=1$ & 0            & +      & $p=1$ \\
  4 & $p=0$ & 0            & 0      & $p=0$ \\
  5 & $p=1$ & 0            & 0      & $p=1$ \\\rowcolor{gray}
  6 & $p=0$ & 0            & -      & $p=0$ \\\midrule
  7 & $p=0$ & -            & +      & $p=1$ \\
  8 & $p=0$ & -            & 0      & $p=0$ \\
  9 & $p=0$ & -            & -      & $p=0$ \\\bottomrule
 \end{tabular}
 \end{minipage}
 \caption{
 (Left) Possible transitions of a propositional value with a STRIPS action.
 (Right) Possible transitions of a propositional value with a STRIPS action,
 modeled by a prevail condition (precondition=0).
}
 \label{tab:regression}
 \label{tab:prevail}
\end{table}

One way to address this uncertainty / non-determinism / disjunctiveness in complete state is
to reformulate the action model
using \emph{prevail condition} introduced in SAS+ formalism \cite{backstrom1995complexity},
a condition that is satisfied when the value is \emph{unchanged}.
In this form, a precondition for a propositional variable is either +, -, or 0 (unchanged),
instead of +, -, * (don't-care).
This modification rules out the ambiguous columns as seen in \reftbl{tab:prevail} (right),
making regression deterministic, and making preconditions conjunctive.
Note that certain combinations of preconditions and effects are equivalent.
For example, lines 0, 1, 3, as well as lines 6, 8, 9 in \reftbl{tab:prevail} are redundant.
We refer to this restricted form of regression as \emph{complete state regression}.

The semantics of preconditions with positive, negative, and prevail conditions
exactly mirrors the semantics of effects (add, delete, NOP),
with the only difference being the direction of application.
For example, BTL is able to model NOP, which does not change the value of a propositional variable.
In complete state regression, this corresponds to a prevail condition.

\subsection{Learning with Complete State Regression in Bidirectional Cube-Space AE (BiCSAE)}
\label{sec:bidirectional}

Complete state regression semantics provides a theoretical background for
modeling a precondition learning problem as an effect learning problem.
Based on this underlying semantics, we now propose \emph{Bidirectional Cube-Space AE} (BiCSAE),
a neural network that can learn the effects and the preconditions at the same time.
Since the semantics of complete state regression and STRIPS effects are equivalent,
we can apply the same Back-to-Logit technique to learn a complete state regression model.
As a result, the resulting network contains a symmetric copy of Cube-Space AE (AMA$_3^+$).

We build BiCSAE by augmenting CSAE with
a network $\aaer(\zafter,\action)$ that uses the same BTL mechanism for predicting the outcome of action regression,
i.e., predict the current state $\zbeforealt$ from a successor state $\zafter$ and a one-hot action vector $\action$.
A BiCSAE also has $\aaeq(\zafter)$, a symmetric counterpart of $\aaep(\zbefore)$
which returns a probability over actions and is used for regularizing $\action$ via KL divergence.
The generative model is also a symmetric copy of AMA$_3^+$.
The main data pipeline excluding the networks for regularization ($\aaeq$, $\aaep$) is formalized in \refig{fig:ama4-network-definition}.
The full network is depicted in \refig{fig:ama4}.

\begin{figure}[htbp]
\begin{align*}
 \relsize{-0.5}
 \text{(input)}                                         & & \xbefore, \xafter       & \\
 \text{(encoder and encoded logits)}                    & & \lbefore, \lafter       & = \encode(\xbefore), \encode(\xafter) \\
 \text{(sampled binary representations)}                & & \zbefore, \zafter       & = \BC(\lbefore), \BC(\lafter)         \\
 \text{(action assignment)}                             & & \action                 & = \GS(\aaee(\lbefore, \lafter))       \\
 \text{(logits for progression / forward dynamics)}     & & \lafteralt              & = \aaed(\zbefore, \action)            \\
 \text{(sampled binary representation for progression)} & & \zafteralt              & = \BC(\lafteralt)                     \\
 \text{(logits for regression / backward dynamics)}     & & \lbeforealt             & = \aaer(\zafter, \action)             \\
 \text{(sampled binary representation for regression)}  & & \zbeforealt             & = \BC(\lbeforealt)                    \\
 \text{(reconstructions)}                               & & \xbeforerec, \xafterrec & = \decode(\zbefore), \decode(\zafter) \\
 \text{(reconstruction based on forward dynamics)}      & & \xafteraltrec           & = \decode(\zafteralt)                 \\
 \text{(reconstruction based on backward dynamics)}     & & \xbeforealtrec          & = \decode(\zbeforealt).               \\
\end{align*}
 \caption{Main data pipeline of \ama4.}
 \label{fig:ama4-network-definition}
\end{figure}

\begin{figure}[htbp]
 \centering
 \includegraphics[width=0.8\linewidth]{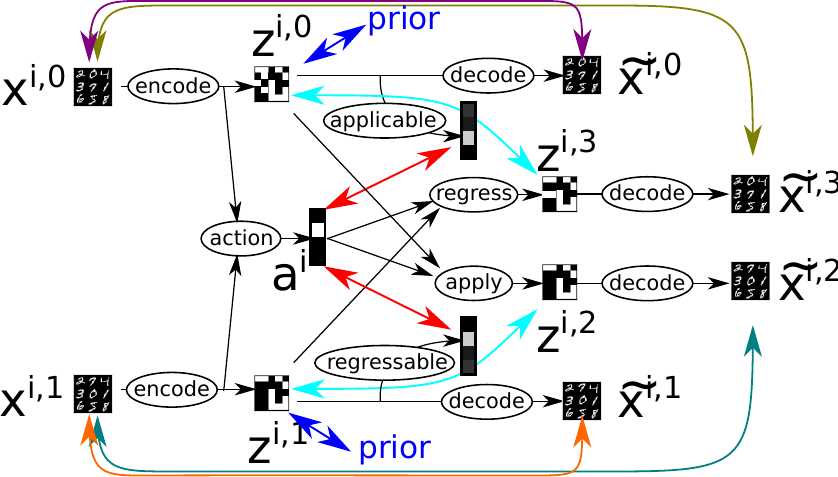}
 \caption{An illustration of Bidirectional Cube-Space AE for AMA$_4^+$.}
 \label{fig:ama4}
\end{figure}

Since $\action$ is learned unsupervised, action replication happens automatically during the training.
$\aaer$ is formalized with a trainable matrix $\mP$ (for preconditions) as:
\begin{align}
 \zbeforealt = \BC(\aaer(\zafter,\action))
 &=\BC(\BN(\zafter)+\BN(\mP\action)).
\end{align}

In $\function{regress}$, add-effects and delete-effects now
correspond to \emph{positive preconditions} and \emph{negative preconditions}.
While negative preconditions are (strictly speaking) out of
STRIPS formalism, it is commonly supported by modern classical planners that
participate in the recent competitions.

BiCSAE restricts the preconditions to be strictly conjunctive.
Note that STRIPS planning assumes deterministic environments, thus its effects are conjunctive:
For example, if an action contained a disjunctive effect such as ``flipping a coin makes it tail \emph{or} head'',
then the state transition is non-deterministic.
As seen in \reftheo{thm:monotonicity}, this does not happen in CSAE.
By using the same mechanism for learning the regression (preconditions),
preconditions learned by BiCSAE do not contain disjunctions,
its regression is deterministic, and it achieve complete regression semantics, as promised.

To extract the preconditions from the network,
we apply the same method used for extracting the effects from the progressive/forward dynamics with two modifications.
First,
when an action contains a prevail condition for a bit $j$ and either an add-effect or a delete-effect for $j$,
we must convert a prevail condition for $j$ into a positive / negative precondition for $j$, respectively.
Otherwise, it underspecifies a set of transitions.
Second, when both the effect and the precondition of a bit $j$ have the XOR semantics,
we should duplicate the action only once (resulting in 2 copies of actions), not separately (resulting in 4 copies of actions).

\label{sec:ama4-elbo}

\renewcommand{\defaultindex}{}
\renewcommand{\defaultcomma}{}

Given a BiCSAE network as shown in \refig{fig:ama4},
we must define its optimization objective as a variational lower bound similar to that of CSAE/AMA$_3^+$.
This is straightforward given the ELBO of CSAE,
because the backward network is a symmetric copy of CSAE.
We obtain an identical loss function for the forward and the backward network, then optimize the average of these objectives.

As seen in \ama3,
each uni-directional model already individually captures
both the preconditions and the effects by itself.
We therefore have duplicate components in the forward and the backward model:
Preconditions are learned by both $\aaep$ and $\aaer$, effects are learned by both $\aaed$ and $\aaeq$.
Two models are individually theoretically justified and we merely trained them at once.
Due to this decoupling,
the training objectives in the forward and the backward direction do not need any special term designed for bidirectionality.
Note that, while two models are trained at once,
they share $\encode$, $\decode$ and $\aaee$, therefore they use the same state encoding and the same set of actions.
In other words, this weight sharing ties two uni-directional models together.

Having duplicate components does not affect the theoretical validity of the model.
This is because, at the limit of convergence where
the lower-bound (ELBO) matches the ground truth ($\log p(\xboth)$) in both forward/backward models,
the variational distributions (distributions using $q$) matches
the ground-truth generative distributions (distributions using $p$), as explained in \refsec{sec:aevae}.
Therefore the forward model eventually matches the backward model and vice versa.

\renewcommand{\defaultindex}{i}
\renewcommand{\defaultcomma}{,}

\section{Training Performance}
\label{sec:training-evaluation}

In this section, we empirically and thoroughly analyze the networks proposed in this paper.
We compare several metrics (overall accuracy, latent state stability,
successor prediction accuracy, monotonicity violation, precondition
accuracy) between different latent space priors, \ama3 and \ama4, and
various hyperparameters.
The primary objective of the comparisons made in this section is to verify the following factors:
\begin{enumerate}
 \item The effect of hyperparameter tuning on the overall accuracy, especially the impact of tuning $\beta_{1..3}$.
 \item The effect of non-standard prior $\bern(\epsilon), \epsilon=0.1$ proposed in \refsec{sec:bcvae-prior},
       compared to the standard uniform Bernoulli prior $\epsilon=0.5$.
 \item The successor prediction accuracy of Back-to-Logit proposed in \refsec{sec:bcvae-prior}.
 \item The number of monotonicity violations described in \refsec{sec:ama3-effect-extraction} in the learned results.
\end{enumerate}
In addition, we provide a training curve plot in \refsec{sec:curve} in the appendix
to facilitate the readers' replication of our result.

We first describe the experimental setup including
dataset generation \refsec{sec:domains} and
training/architecture details \refsec{sec:network-specifications}.

\subsection{Experimental Domains}
\label{sec:domains}

\begin{figure}
 \includegraphics[width=0.16\linewidth]{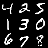}
 \includegraphics[width=0.16\linewidth]{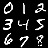}
 \includegraphics[width=0.16\linewidth]{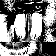}
 \includegraphics[width=0.16\linewidth]{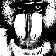}
 \includegraphics[width=0.16\linewidth]{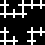}
 \includegraphics[width=0.16\linewidth]{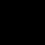}
 \includegraphics[width=0.16\linewidth]{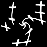}
 \includegraphics[width=0.16\linewidth]{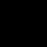}
 \includegraphics[width=0.16\linewidth]{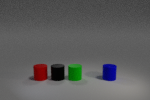}
 \includegraphics[width=0.16\linewidth]{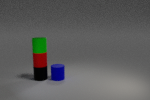}
 \includegraphics[width=0.16\linewidth]{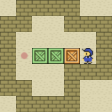}
 \includegraphics[width=0.16\linewidth]{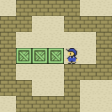}
 \caption{An example of problem instances (initial and goal images) for each domain.
 In Blocksworld and Sokoban, the images shown are high-resolution images,
 while the actual training and planning experiments are performed on smaller resized images.}
 \label{fig:example}
\end{figure}

We evaluate Latplan on 6 domains. Example visualizations are shown in \refig{fig:example}.
\begin{itemize}
 \item
\textbf{MNIST 8-Puzzle} \citep{Asai2018}
is a 42x42 pixel, monochrome image-based version of the 8-Puzzle.
Tiles contain hand-written digits (0-9) from the  MNIST database \citep{lecun1998gradient},
which are shrunk to 14x14 pixels so that each state of the puzzle is a 42x42 image.
Valid moves swap the ``0'' tile  with a neighboring tile,
i.e., the ``0'' serves as the ``blank'' tile in the classic 8-Puzzle.
8-Puzzle has 362880($=9!$) states and 967680 transitions,
while half of the states are unreachable.
The longest shortest path in 8-puzzle contains 31 moves \cite{Reinefeld93}.
From any specific goal state, the reachable
number of states is 181440 ($9!/2$).  Note that the same image is used
for each digit in all states, e.g., the tile for the ``1'' digit is the
same image in all states.
 \item
The \textbf{Scrambled Photograph (Mandrill) 15-puzzle}
cuts and scrambles a real photograph, similar to the puzzles sold in stores.
We used a ``Mandrill'' image taken from the USC-SIPI benchmark set \cite{weber1997usc}.
These differ from the MNIST 8-puzzle in that ``tiles'' are \textit{not} cleanly separated by black regions
(we re-emphasize that \latentplanner has no built-in notion of square or movable region).
Unlike 8-puzzle, this domain can increase the state space easily by dissecting the image 4x4 instead of 3x3,
generating 15-puzzle instances whose state space is significantly larger than that of 8-puzzle.
The image consists of 56x56 pixels.
It was converted to greyscale, and then
histogram-normalization and contrast enhancement was applied.
 \item
A recreation of \textbf{LightsOut} \citep{anderson1998turning} by Tiger Electronics in 1995,
or sometimes also known as ``Magic Square'' game mode of \emph{Merlin The Electronic Wizard}.
It is an electronic puzzle game where a grid of lights is in some on/off configuration ($+$: On),
and pressing a light toggles its state as well as the states of its neighbors.
The goal is to turn all lights Off.
Unlike previous puzzles, a single operator can flip 5 locations at once and
removes some ``objects'' (lights).
This demonstrates that \latentplanner is not limited to domains with highly local effects and static objects.
$n \times n$ LightsOut has $2^{n^2}$ states and $n^22^{n^2}$ transitions.
In this paper, we used 5x5 configuration (33554432 states)
whose images consist of 45x45 pixels.
 \item
\textbf{Twisted LightsOut} distorts images in LightsOut by a swirl effect available in scikit-image package,
showing that \latentplanner is not limited to handling rectangular ``objects''/regions.
It has the same image dimensions as the original LightsOut.
The effect is applied to the center, with strength=3, linear interpolation,
and radius equal to 0.75 times the dimension of the image.
 \item
\textbf{Photo-realistic Blocksworld} \citep{Asai2018b} is a dataset that consists of 100x150 RGB images rendered by Blender 3D engine.
The image contains complex visual elements such as reflections from neighboring blocks.
We modified the rendering parameter to simplify the dataset:
We limited the object shapes to be cylinders and removed the jitters (small noise) in the object placements.
The final images are resized to 30x45 pixels.
As shown in the figures, Blocksworld contains 4 objects and maximum 5 towers,
which results in $4! \times 5^4 = 15000$ logical states and
 each state has a branching factor from 4 (where all blocks are stacked in a single tower)
 to 20 (where all blocks are on the table).
 For each logical state, there are infinite number of images
 due to different random jitter and raytracing rendering noise.

 \item
\textbf{Sokoban} \citep{culberson1997sokoban,junghanns2001sokoban} is a PSPACE-hard puzzle domain
whose 112x112 pixel visualizations are obtained from the PDDLGym library \citep{silver2020pddlgym}
and are resized into 28x28 pixels.
Unlike other domains, this domain is irreversible and dead-ends exist.
We used p006-microban-sequential instance as the visualization source.
p006-microban-sequential has 23363 states reachable from the initial state.
 States have varying branching factors from 1 to 4.
 The dataset is generated by generating all reachable states with an exhaustive Dijkstra search and their corresponding transitions,
 then sampling 20000 valid transitions.
\end{itemize}

In 8-Puzzle and two Lightsout domains, we sampled 5000 transitions / 10000 states.
In 15-Puzzle, Blocksworld, and Sokoban, we sampled 20000 transitions / 40000 states because
they are visually challenging domains
that contain more objects, more noise, and more complex modes of state transitions.
All datasets are generated by domain-specific generators.
Except sokoban, the generators generate a pair of states at a time:
One uniformly sampled configuration of the environment and its uniformly sampled successor state.
In Sokoban, the dataset is uniformly sampled from the whole state space enumerated by an exhaustive search.
These datasets are divided into 90\%,5\%,5\% for the training set, validation/tuning set, and testing set, respectively.
During the development of the network models, we look at the validation set to tune the network
while only in the final report we evaluate the network on the test set and report the results.
While the number of states (images) used for visually challenging domains may appear to be reaching the interpolation rather than generalization,
note that there are much more transitions (multiplied by the branching factor) in the environment that the system must generalize to.

The pixel values in the image datasets are normalized to mean 0, variance 1 for each pixel and channel, i.e.,
for an image pair $(\vx^{i,0},\vx^{i,1})$ where $\vx^{i,j}\in \R^{H,W,C}$
(with $H,W,C$ being height, width, and color channel respectively),
$\forall k,l,m; \E_{i,j} [\vx^{i,j}_{klm}] = 0$ and $\Var_{i,j} [\vx^{i,j}_{klm}] = 1$.

All experiments are performed on a distributed compute
cluster equipped with nVidia Tesla K80 / V100 / A100 GPUs and Xeon E5-2600 v4.
With V100 (a Volta generation GPU),
training a single instance of \ama3 model on 8-Puzzle takes 2 hours,
and \ama4 model takes 3 hours due to additional networks.
15-Puzzle, Blocksworld, and Sokoban take more runtime varying from 5 hours to 15 hours
due to more data points and larger images.
The system is implemented on top of Keras library \cite{chollet2015keras}.

\subsection{Network Specifications}
\label{sec:network-specifications}

We trained the \ama3 and the \ama4 models as discussed in \refsecs{sec:ama3}{sec:ama4}.
We do not include training results of previous models AMA$_{1..3}$
due to the lack of STRIPS export capability and their ad-hoc natures.
The ad-hoc nature indeed prevents us from defining an effective metric
that is independent from the implementation of the neural network.
For example, while we can directly compare the loss values of the \ama3 and the \ama4
because both values are lower bounds of the likelihood $p(\rxboth)$,
we cannot derive a meaningful conclusion by comparing the loss values of AMA$_{1..3}$.

Detailed specifications of each network component of \ama3 and \ama4 are listed in \reftbl{tab:architecture}.
The encoder and the decoders consist of
three convolutional layers \cite{fukushima1980neocognitron,lecun1989backpropagation} each,
while action-related networks primarily consist of fully connected networks.
We use Rectified Linear Unit ($\function{Relu}(x)=\max(0,x)$) as the activation function of each layer,
which performs better in deep networks because it avoids gradient vanishing compared to a $\sigmoid$ function.
Each activation is typically followed by Batch Normalization ($\BN$) \cite{ioffe2015batch} and Dropout \cite{hinton2012improving} layers,
both of which are generally known to improve the performance and allow for training the network with a higher learning rate.
A layer that is immediately followed by a $\BN$ layer does not need the bias weights because $\BN$ has its own bias weights.
Following the theoretical and empirical finding by \citet{he2015delving},
all layers that are immediately followed by a ReLU activation are initialized by Kaimin He's uniform weight initialization \cite{he2015delving},
as opposed to other layers that are initialized by the Glorot Xavier's uniform weight initialization \cite{glorot2010understanding},
which is the default in Keras library.
In addition, the input is augmented by a GaussianNoise layer which adds a fixed amount of Gaussian noise,
making the network an instance of Denoising AutoEncoder \cite{vincent2008extracting}.

For each dataset,
we evaluated 30 hyperparameter configurations using grid search,
varying the latent space size $F$ and the $\beta_1$, $\beta_3$ coefficients of KL divergence terms ($\beta_2$ was kept at 1).
Parameter ranges are listed in \reftbl{tab:hyper}.
We tested numerous hyperparameters during the development (using validation set)
to narrow down the range of hyperparameters to those listed in this table.
Naturally, a single hyperparameter does not work for all domains because, for example,
a more complex domain (e.g., 15 Puzzle) requires a larger latent space than a simpler domain (e.g., 8 Puzzle).
The values in this table have a sufficient range for covering all domains we tested.

\begin{table}[p]
\centering
\begin{tabular}{|cc|}\hline
\emph{Subnetworks} & \emph{Layers}                                   \\\hline
\multirow{4}{*}{$\encode$}
          & GaussianNoise(0.2), $\BN$,                     \\
                   & Conv(5,32), ReLU, $\BN$, Dropout(0.2),  \\
                   & Conv(5,32), ReLU, $\BN$, Dropout(0.2),  \\
                   & Conv(5,32)                                   \\\hline
\multirow{4}{*}{$\decode$}          & $\BN$,       \\
                   & Conv(5,32), ReLU, $\BN$, Dropout(0.2),  \\
                   & Conv(5,32), ReLU, $\BN$, Dropout(0.2),  \\
                   & Conv(5,32)                                    \\\hline
$\aaee$            & $\sigmoid$, fc(1000), ReLU, $\BN$, Dropout(0.2), fc(6000) \\\hline
$\aaed,\aaer$      & \refeq{eq:btl-implementation}                   \\\hline
$\aaep,\aaeq$      & fc(6000)                                        \\\hline
\end{tabular}
\caption{
Detailed overview of the subnetworks of \ama3 and \ama4.
``Conv$(k,c)$'' denotes a convolutional layer with kernel size $k$ and channel width $c$,
and ``fc$(w)$'' denotes a fully-connected layer with width $w$.
We omit reshaping operators such as ``flatten.''
To maintain the notational consistency with \refig{fig:ama3},
this table does not include $\BC$ activation in $\encode$ and
$\GS$ activation in $\aaee$.
}
\label{tab:architecture}
\end{table}

\begin{table}[p]
\centering
\begin{tabular}{|ll|}
\hline
\emph{Training parameters} &                                \\
Optimizer                  & Rectified Adam \cite{liu2019radam} \\
Training Epochs            & 2000                           \\
Batch size                 & 400                            \\
Learning rate              & $10^{-3}$                      \\
Gradient norm clipping     & 0.1                            \\
\hline
\emph{Gumbel Softmax / Binary Concrete annealing parameters} &                                                                                                         \\
Initial annealing temperature $\tau_{\text{max}}$            & 5                                                                                                       \\
Final annealing temperature $\tau_{\text{min}}$              & 0.5                                                                                                     \\
Annealing schedule $\tau(t)$ for epoch $t$                   & $\tau_{\text{max}} \parens{\frac{\tau_{\text{min}}}{\tau_{\text{max}}}}^\frac{\min(t,1000)}{1000}$ \\
\hline
\emph{Network shape parameters}    &     \\
 Latent space dimension $F$        & $F\in\braces{50,100,300}$ \\
 Maximum number of actions $A$     & $A=6000$ \\
\hline
\emph{Loss and Regularization Parameters}                                                &                                       \\
$\sigma$ for all reconstruction losses (cf. \refsec{sec:aevae})                          & $0.1$                                 \\
$\beta_1$ for $\KL(\blue{q(\zbefore\mid\xbefore) \Mid p(\zbefore)})$                     & $\beta_1\in\braces{1,10}$                                  \\
$\beta_2$ for $\KL(\red{q(\action\mid\xboth) \Mid p(\action\mid\zbefore)})$              & $1$                                   \\
$\beta_3$ for $\KL(\cyan{q(\zafter\mid\xafter) \Mid p(\zafteralt\mid\zbefore,\action)})$ & $\beta_3\in\braces{1,10,100,1000,10000}$ \\
\hline
\end{tabular}
\caption{List of hyperparameters.}
\label{tab:hyper}
\end{table}

\subsection{Accuracy Measured as a Lower Bound of Likelihood}
\label{sec:elbo}

We first evaluate the ELBO, the lower bound of the likelihood $\log p(\rxboth)$ of observing the entire dataset (the higher, the better).
The summary of the results is shown in \reftbl{tab:elbo}.
We report the ELBO values obtained by setting $\beta_{1..3}$ to 1 after the training.
The original maximization objectives with $\beta_{1..3}\geq 1$ used for training are lower bounds of the ELBO with $\beta_{1..3}=1$
because all KL divergences have a negative sign (\refeq{eq:vanilla-elbo8})
and KL divergences themselves are always positive (\refeq{sec:aevae}).
Resetting $\beta_{1..3}$ to 1 improves / increases the lower bound while still being the lower bound of the likelihood.
Since ELBO is a model-agnostic metric, we can directly compare the values of different models.

The actual numbers we report are loss values -ELBO (negative value of ELBO) that are minimized (lower the better).
In addition, the numbers do not include a constant term $C_2 = \log \sqrt{2\pi\sigma^2}$ with $\sigma=0.1$
multiplied by the number of pixels (cf. \refsec{sec:aevae}).
While the constant should be added to obtain the true value of -ELBO,
and -ELBO+Constant has an intuitive interpretation as a sum of scaled squared errors and KL divergences.
The constant does not affect the training either.
The drawback is that the -ELBO we report should not be compared between different domains
because different datasets have different numbers of pixels, resulting in a different amount of constant offsets.

\dnote{
Note that the best accuracy/ELBO alone
is \emph{not} the sole factor that improves the likelihood of success in solving visual planning problems.
It is affected by multiple factors, including all metrics we evaluate in the following sections.
For example, even if the overall accuracy (-ELBO, which is a sum of reconstruction and KL losses) is good,
the planning is not likely to succeed
if its action model does not predict the latent successor state accurately
because it overly focuses on improving the reconstruction accuracy
and sacrifices successor prediction accuracy modeled by
$\beta_3\KL(\cyan{q(\zafter\mid\xafter) \Mid p(\zafteralt\mid\zbefore,\action)})$.
}

\begin{table}[htb]
\centering
\begin{adjustbox}{width=\linewidth,keepaspectratio}
\begin{tabular}{|r|rrrr|rrrr|rrrr|}
\hline
 & \multicolumn{ 4}{c|}{\textbf{kltune}} & \multicolumn{ 4}{c|}{notune} & \multicolumn{ 4}{c|}{default} \\
 & \multicolumn{ 4}{c|}{$\epsilon=0.1$} & \multicolumn{ 4}{c|}{$\epsilon=0.1, \beta_{1..3}=1$} & \multicolumn{ 4}{c|}{$\epsilon=0.5$} \\
domain & -ELBO & $\beta_1$ & $\beta_3$ & $F$ & -ELBO & $\beta_1$ & $\beta_3$ & $F$ & -ELBO & $\beta_1$ & $\beta_3$ & $F$ \\
\hline
 \multicolumn{ 13}{|c|}{\ama3} \\
\hline
Blocks & \textbf{\textit{6.26E+03}} & 10 & 1 & 300 & 6.55E+03 & 1 & 1 & 300 & \textit{6.32E+03} & 1 & 100 & 300 \\
LOut & 1.97E+03 & 1 & 1000 & 50 & 2.07E+03 & 1 & 1 & 100 & \textbf{1.90E+03} & 1 & 1000 & 300 \\
Twisted & 1.95E+03 & 10 & 1000 & 50 & 2.03E+03 & 1 & 1 & 50 & \textbf{1.93E+03} & 10 & 1000 & 300 \\
Mandrill & \textbf{2.65E+03} & 10 & 10 & 300 & 3.39E+03 & 1 & 1 & 100 & \textit{3.09E+03} & 10 & 10 & 300 \\
MNIST & \textbf{1.21E+03} & 10 & 10 & 300 & 1.41E+03 & 1 & 1 & 100 & \textit{1.21E+03} & 1 & 10 & 300 \\
Sokoban & \textbf{\textit{1.45E+03}} & 10 & 1 & 50 & 1.60E+03 & 1 & 1 & 50 & 1.50E+03 & 1 & 1 & 50 \\
\hline
 \multicolumn{ 13}{|c|}{\ama4} \\
\hline
Blocks & 7.43E+03 & 1 & 1 & 100 & 7.61E+03 & 1 & 1 & 100 & \textbf{7.25E+03} & 10 & 10 & 100 \\
LOut & \textbf{\textit{1.82E+03}} & 10 & 10 & 300 & 1.89E+03 & 1 & 1 & 300 & \textit{1.85E+03} & 10 & 1 & 300 \\
Twisted & \textbf{\textit{1.82E+03}} & 10 & 1 & 300 & 1.93E+03 & 1 & 1 & 300 & \textit{1.85E+03} & 10 & 1 & 300 \\
Mandrill & \textbf{\textit{2.58E+03}} & 10 & 100 & 300 & 3.26E+03 & 1 & 1 & 100 & 3.18E+03 & 10 & 100 & 100 \\
MNIST & \textbf{1.31E+03} & 10 & 1 & 300 & 1.50E+03 & 1 & 1 & 300 & 1.34E+03 & 10 & 1 & 300 \\
Sokoban & 1.53E+03 & 10 & 10 & 300 & 1.55E+03 & 1 & 1 & 100 & \textbf{\textit{1.12E+03}} & 10 & 10 & 300 \\
\hline
\end{tabular}
\end{adjustbox}
 \caption{
Best negative ELBO (lower the better) of three different configurations (kltune, notune, default)
of \ama3 and \ama4 models and the hyperparameters which achieved it.
The best results among three configurations are highlighted in \textbf{bold}.
Also, the better result among \ama3 and \ama4 are highlighted in \textit{italic}.
Overall, in terms of -ELBO, which represents training accuracy,
\textbf{kltune is better than notune}, \textbf{kltune and default are comparable},
and \textbf{\ama4 and \ama3 are comparable}.
}
 \label{tab:elbo}
\end{table}

In \reftbl{tab:elbo},
we first assess the effectiveness of $\beta$-VAE objective with tuned $\beta_{1..3}$.
Comparing the best hyperparameters (``kltune'') and the baseline
that uses $\beta_{1..3}=1$ during the training (``notune''),
we observed that tuning these parameters is important for obtaining accurate models.
Note that the latent space size $F$ is still tuned in both cases.

\begin{figure}[p]
 \centering
 \includegraphics[width=0.4\linewidth]{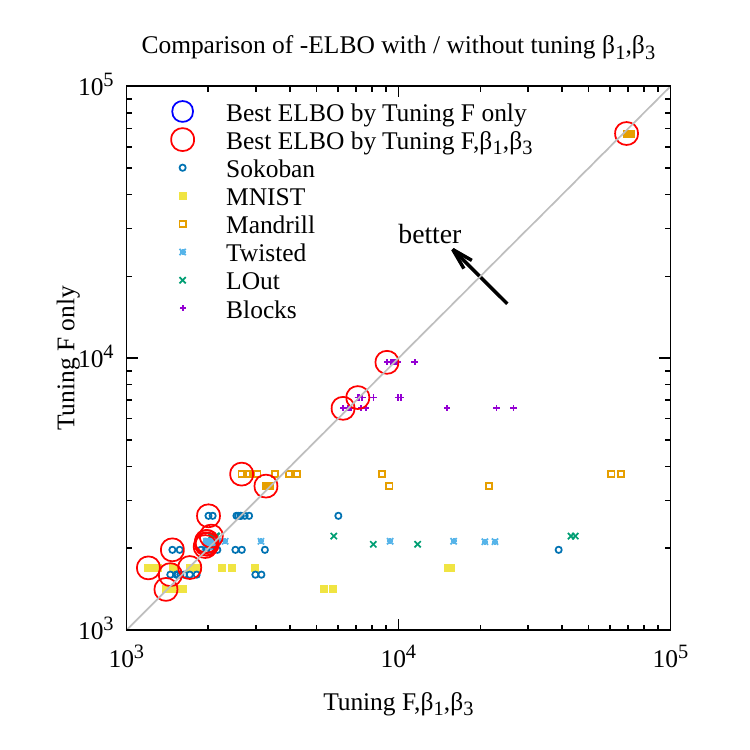}
 \includegraphics[width=0.4\linewidth]{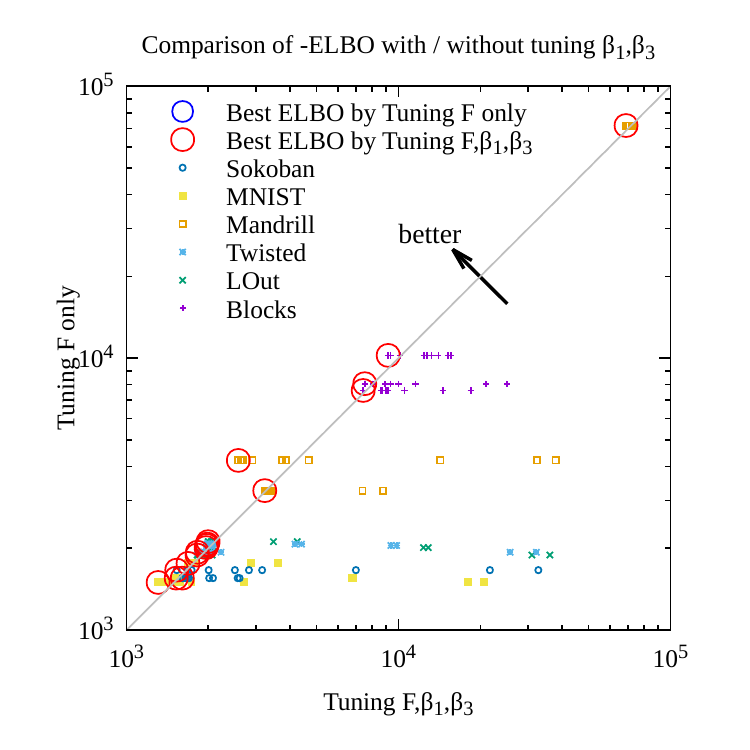}
 \caption{
 (Left: \ama3, Right: \ama4.)
 Comparing the effect of tuning $\beta_{1..3}$ on -ELBO.
 Each point represents a pair of configurations with the same $F$,
 where the $y$-axis always represents the result from $\beta_1=\beta_3=1$,
 while the $x$-axis represents the results from various $\beta_1,\beta_3$.
 For each $F$, we highlighted the best configuration of $\beta_1,\beta_3$ with blue circles.
 While bad parameters will significantly degrade the accuracy (below the diagonal),
 tuning the value appropriately will improve the accuracy (above the diagonal).
 }
 \label{fig:vs-notune}
\end{figure}

We also plotted the results with all matching hyperparameters (same $F$) in \refig{fig:vs-notune}.
The scatter plot provides richer information on how these hyperparameters affect the ELBO.
It emphasizes the importance of hyperparameter tuning:
Bad hyperparameter of $\beta_{1..3}$ could quite negatively affect the result
while tuning the value appropriately yields a better result.

\begin{figure}[p]
 \centering
 \includegraphics[width=0.4\linewidth]{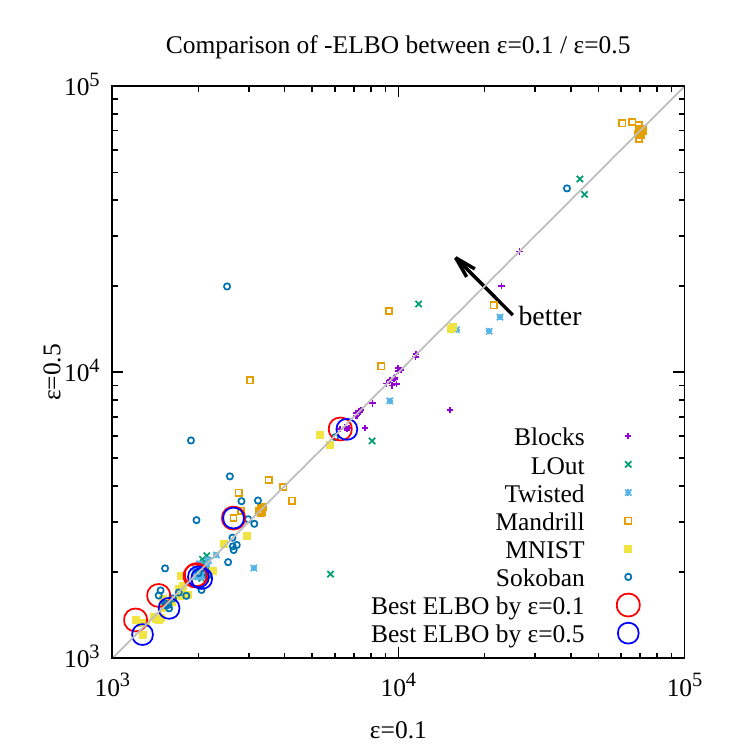}
 \includegraphics[width=0.4\linewidth]{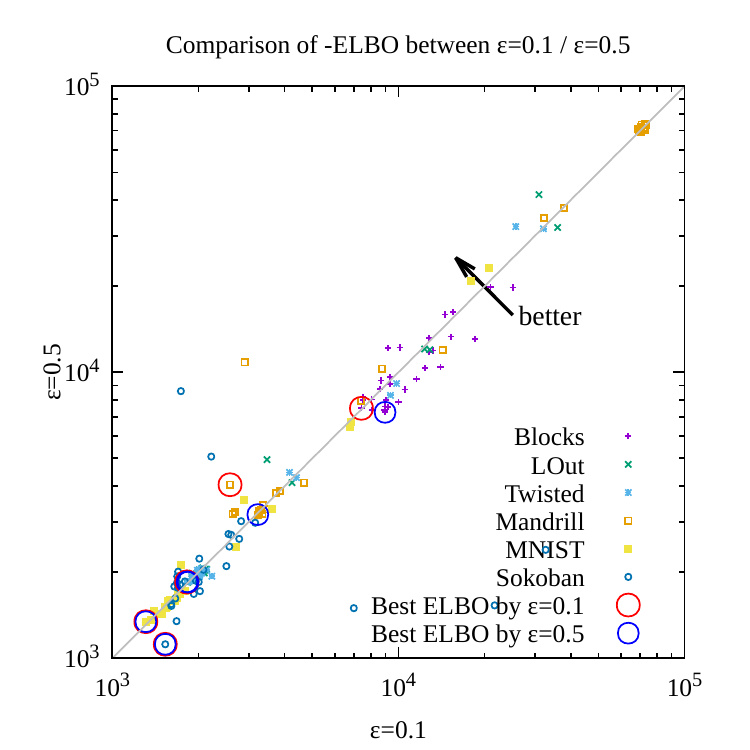}
 \caption{
 (Left: \ama3, Right: \ama4.)
 Comparing the effect of the prior distribution on -ELBO.
 Each point represents a pair of configurations with the same $F,\beta_1,\beta_3$,
 where the $x$-axis represents a result from $\epsilon=0.1$,
 while the $y$-axis represents a result from $\epsilon=0.5$.
 As expected, we do not see a significant difference of ELBO between these two configurations.
 }
 \label{fig:vs-nozsae}
\end{figure}

We next compare the results obtained by training the networks with a standard $\bern(\epsilon=0.5)$ prior
and with a proposed $\bern(\epsilon=0.1)$ prior for the latent vector $\vz$.
As we already discussed in \refsec{sec:unstable},
we observed comparable results; therefore it does not affect the \emph{accuracy}.
For the scatter plot, please refer to \refig{fig:vs-nozsae}.
This is not surprising, as we are merely training the network with two different underlying assumptions:
 $\bern(\epsilon=0.5)$ assumes an \emph{open-world assumption},
while $\lim_{\epsilon\to 0}\bern(\epsilon)$ assumes a \emph{closed-world assumption} (\refsec{sec:bcvae-prior}).

\begin{figure}[htbp]
 \centering
 \includegraphics[width=0.4\linewidth]{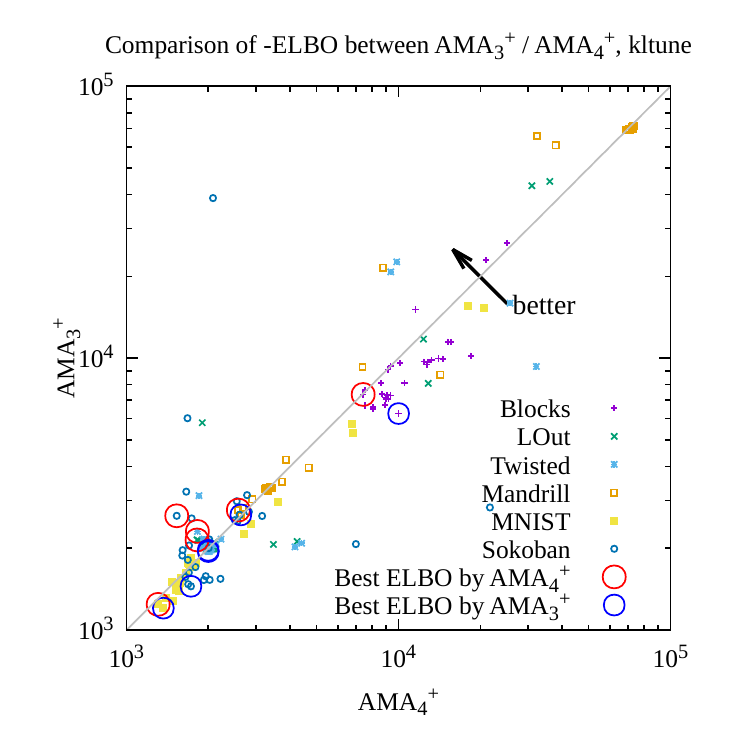}
 \includegraphics[width=0.4\linewidth]{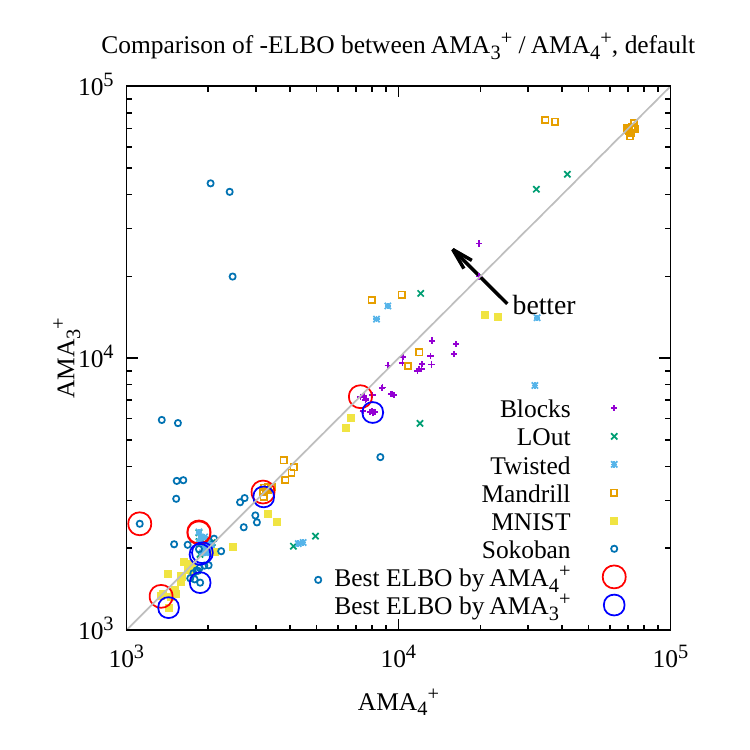}
 \caption{
 Comparing the accuracy (-ELBO) between \ama3 and \ama4 (Left: $\epsilon=0.1$, Right: $\epsilon=0.5$).
 Each point represents a pair of configurations with the same hyperparameter tuple $(F,\beta_1,\beta_3)$,
 where the $x$-axis represents a result from \ama3,
 while the $y$-axis represents a result from \ama4.
 We do not see a significant difference of ELBO between them.
 }
 \label{fig:ama3-ama4}
\end{figure}

We next compare \ama3 and \ama4 models by their ELBO.
\reftbl{tab:elbo} and \refig{fig:ama3-ama4} show that the accuracy of \ama3 and \ama4 are comparable.

\subsection{Stability of Propositional Symbols Measured as a State Variance}
\label{sec:stability}

For each hyperparameter which resulted in the best ELBO in each domain and configuration,
we next measured other effects of using $p(\rzbefore[0])=\bern(\epsilon=0.1)$ (``kltune'')
against the default $p(\rzbefore[0])=\bern(\epsilon=0.5)$ configuration (``default'').
We focus on the stability of a latent vector $\zbefore$ on perturbed inputs,
whose role in a symbol grounding process was discussed in \refsec{sec:unstable}.

The stability is measured by the \emph{state variance} for the noisy input, i.e.,
the variance of the latent vectors $\zbefore=\BC(\encode(\xbefore+\vn))$ where $\vn$
is a noise following a Gaussian distribution $\gaussian(\mu=0,\sigma=0.3)$.
We compute the variance by iterating over 10 random vectors, then
average the results over $F$ bits in the latent space and the dataset index $i$.
Formally,
\[
 \text{State Variance} = \E_{f\in 0..F}\E_{i} \Var_{j\in 0..10}[\BC(\encode(\xbefore+\vn^j)_f)].
\]

\begin{figure}[htbp]
 \centering
 \includegraphics[width=0.4\linewidth]{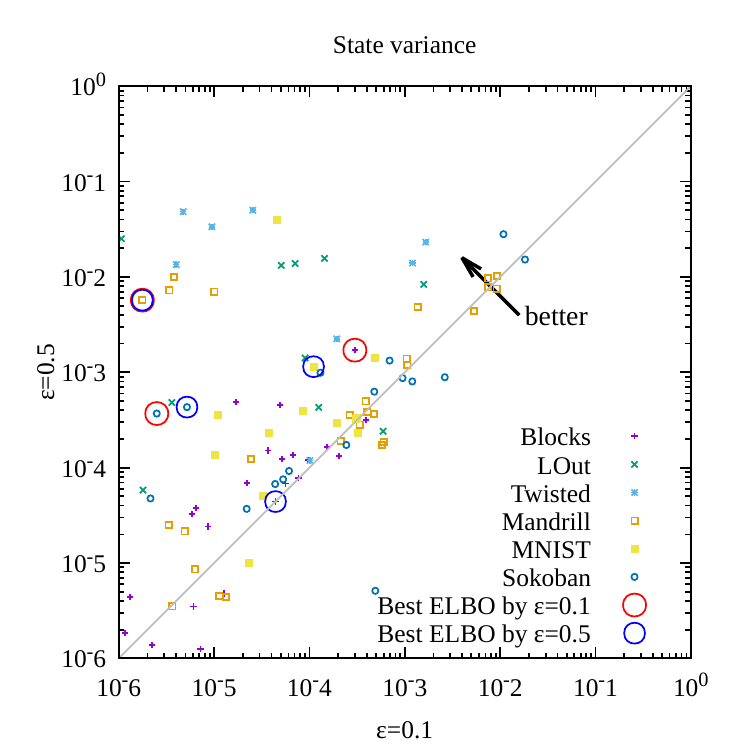}
 \includegraphics[width=0.4\linewidth]{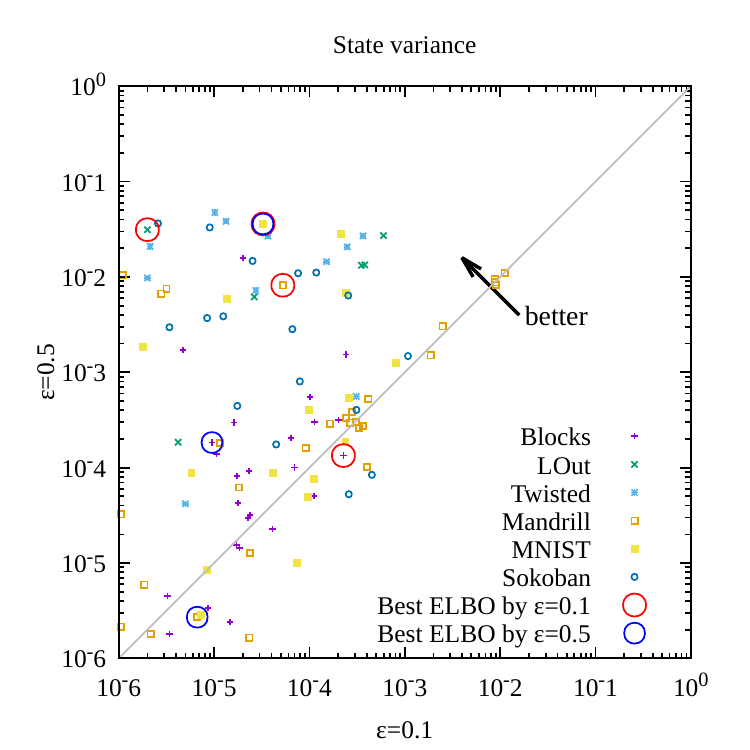}
 \caption{
 Each subfigure corresponds to \ama3 (left) and \ama4 (right).
 Results showing the variance of the latent vectors when the input image is corrupted by Gaussian noise,
 where the $x$-axis corresponds to the value in $\epsilon=0.1$ configuration and
 the $y$-axis shows the value in $\epsilon=0.5$ configuration, while the rest of the hyperparameters are the same.
 Additionally, we highlighted each configuration in a circle when it achieved the best ELBO in each domain.
 }
 \label{fig:variance}
\end{figure}

As we discussed in \refsec{sec:unstable} about the potential cause of unstable symbols,
this value is likely to increase when the latent space has an excessive capacity (larger $F$)
and could also be affected by other hyperparameters.
Therefore, we made a scatter plot where each point represents
a result of evaluating $\epsilon=0.1$ and $\epsilon=0.5$ configurations
with the same hyperparameter.
In \refig{fig:variance}, we observed that the network trained with $\bern(\epsilon=0.1)$ has a lower state variance,
confirming that the proposed prior makes the resulting symbolic propositional states more stable against aleatoric uncertainty.

\begin{figure}[htbp]
 \centering
 \includegraphics[width=0.4\linewidth]{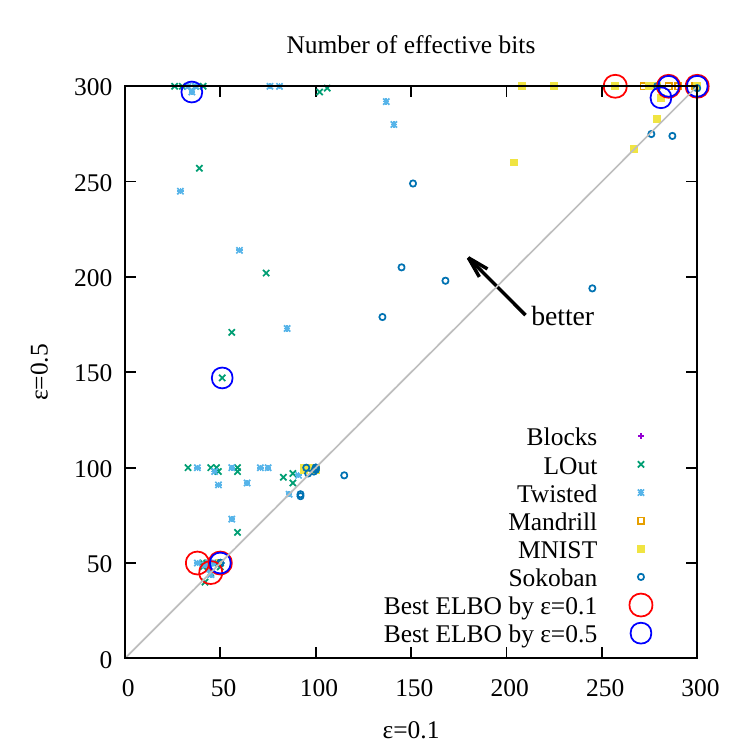}
 \includegraphics[width=0.4\linewidth]{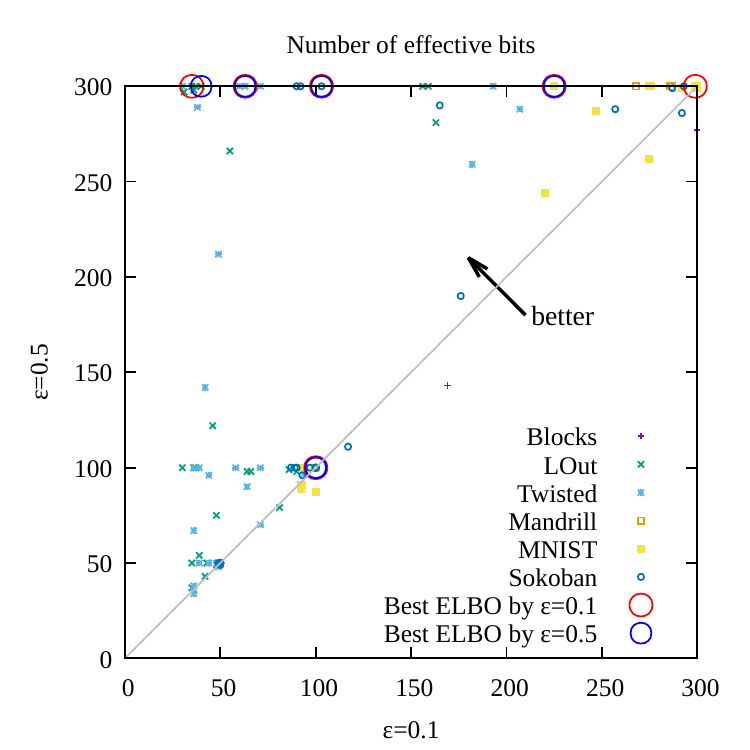}
 \caption{
 Each subfigure corresponds to \ama3 (left) and \ama4 (right).
 Results showing the number of effective bits in the test dataset
 where the $x$-axis corresponds to the value in $\epsilon=0.1$ configuration and
 the $y$-axis shows the value in $\epsilon=0.5$ configuration, while the rest of the hyperparameters are the same.
 Additionally, we highlighted each configuration in a circle when it achieved the best ELBO in each domain.
 $\epsilon=0.1$ has a significantly fewer number of effective bits.
 }
 \label{fig:effective}
\end{figure}

Another related metric we can evaluate is how many ``effective bits'' each model uses in the latent space.
An effective bit is defined as follows:
For each bit $f$ in the latent vector $\zbefore=\BC(\encode(\xbefore))$ where $\xbefore$ is from a test dataset,
we check if $\zbefore_f$ \emph{ever} changes its value when we iterated across different $i$ in the dataset.
If it changes, it is an ``effective bit.'' Otherwise, the bit is a constant and is not used for encoding the information in an image.
In \refig{fig:effective},
we observed that our proposed configuration (``kltune'') tends to have fewer effective bits;
thus each bit of the latent vectors changes its value less frequently across the dataset.

\begin{figure}[htbp]
 \centering
 \includegraphics[width=0.4\linewidth]{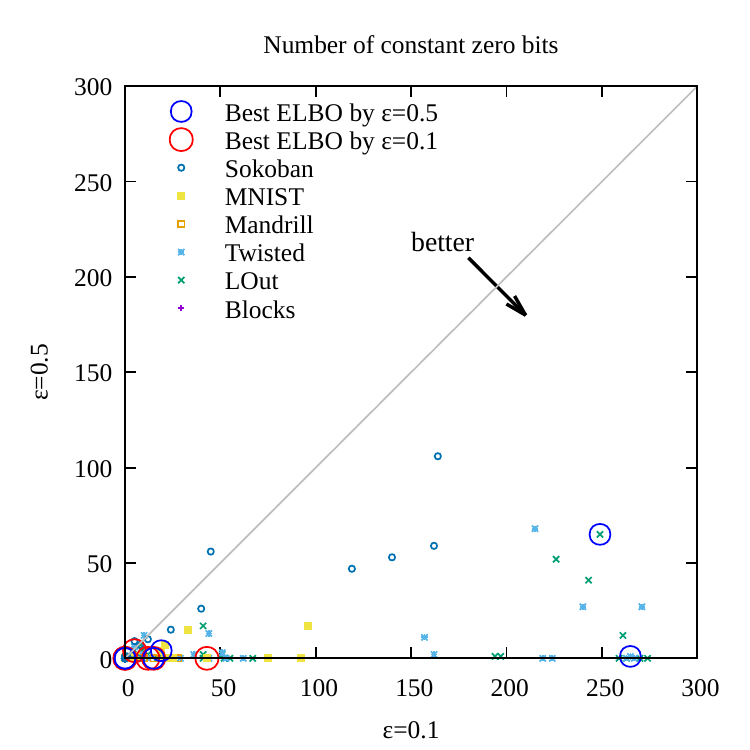}
 \includegraphics[width=0.4\linewidth]{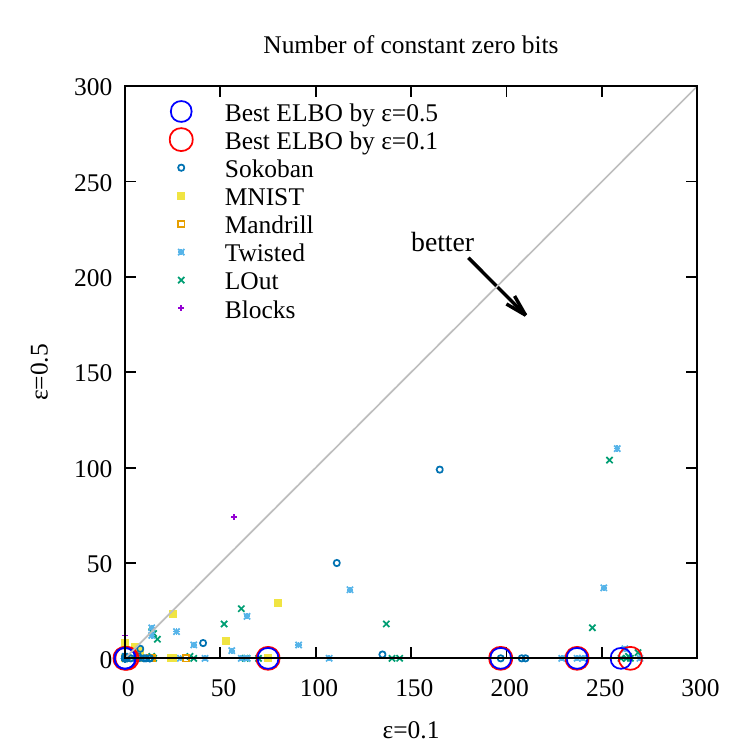}
 \caption{
 Each subfigure corresponds to \ama3 (left) and \ama4 (right).
 Results showing the number of constant zero bits in the test dataset
 where the $x$-axis corresponds to the value in $\epsilon=0.1$ configuration and
 the $y$-axis shows the value in $\epsilon=0.5$ configuration, while the rest of the hyperparameters are the same.
 Additionally, we highlighted each configuration in a circle when it achieved the best ELBO in each domain.
 $\epsilon=0.1$ has significantly more constant bits.
 }
 \label{fig:constantzero}
\end{figure}

Furthermore, we confirmed that those static bits tend to be 0 rather than 1.
We extracted the bits whose values are always 0 regardless of the input
by iterating over the test dataset.
\refig{fig:constantzero} shows that the modified prior induces more constant 0 bits in the latent space.

Note that better stability does not by itself imply
the overall ``quality'' of the resulting network and should always be considered in conjunction with other metrics.
For example, a failed training result that exhibits \emph{posterior collapse}
may have a latent space where all bits are always 0, and may fail to reconstruct an input image at all.
While such a latent space shows a perfect ``stability,'' the network is not useful for planning
because it does not map the input image to an informative latent vector.

\subsection{Evaluating the Accuracy of Action Models: Successor State Prediction}
\label{sec:effect}

We next measured the accuracy of the progression prediction (effect).
In this experiment, we measured $\E_{i,f} |\zafter_f-\zafteralt_f|$, i.e.,
the absolute error between $\zafter$ (encoded directly from the image) and $\zafteralt$ (predicted by the BTL mechanism)
averaged over latent bits and the test dataset.
We show the absolute error instead of the corresponding KL divergence loss
$\KL(\cyan{q(\zafter\mid\xafter) \Mid p(\zafteralt\mid\zbefore,\action)})$
which was used for training because
the average absolute error gives an intuitive understanding of ``how often it predicts a wrong bit.''

\begin{figure}[htbp]
 \centering
 \includegraphics[width=0.47\linewidth]{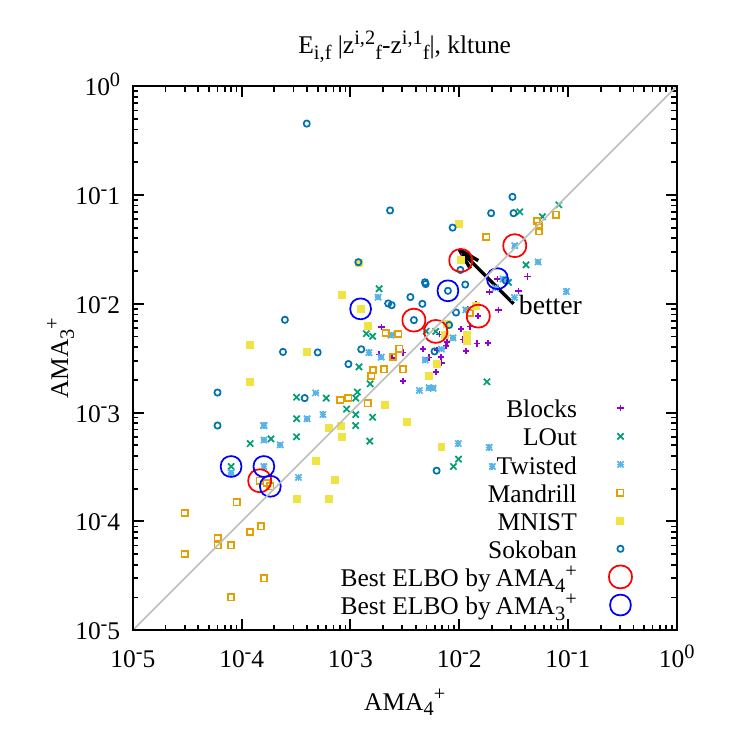}
 \includegraphics[width=0.47\linewidth]{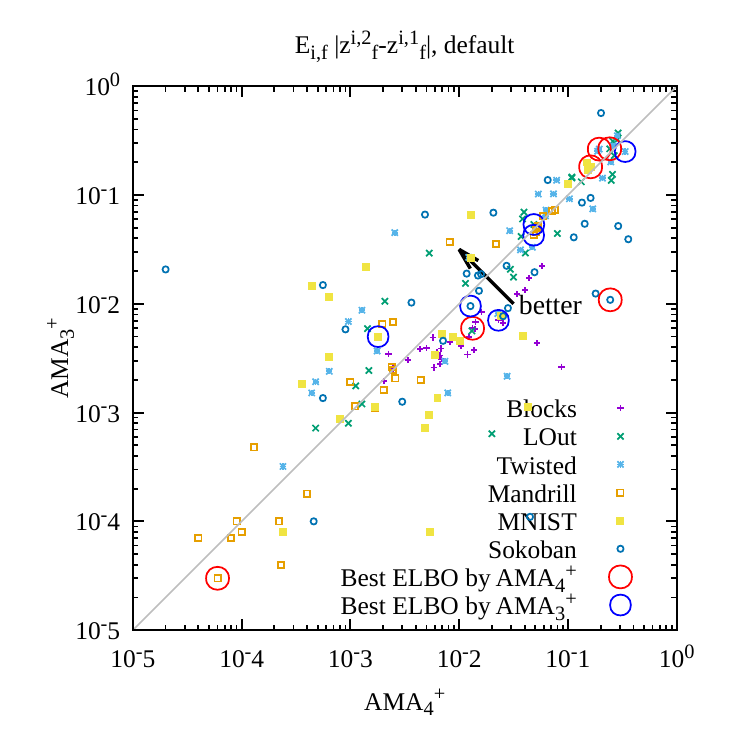}
 \caption{
 Both axes represent the
 absolute error $|\zafteralt-\zafter|$ between $\zafter$ obtained from the encoder distribution $q(\zafter|\xafter)$
 and $\zafteralt$ obtained from the AAE distribution $p(\zafteralt|\zbefore,\action)$
 (applying an action to the current state using a BTL mechanism).
 The values are averaged over the test dataset index $i$ and the latent dimension $f$.
 Each point $(x,y)$ represents a pair of (\ama4, \ama3) configurations with the same hyperparameter tuple $(F,\beta_1,\beta_3)$,
 where the $x$-axis represents a result from \ama4.
 The left figure is a result from networks trained with $\epsilon=0.1$,
 and the right figure is a result with $\epsilon=0.5$.
 With $\epsilon=0.1$, \ama4 tends to have a slightly better successor prediction accuracy.
 With $\epsilon=0.5$, we do not observe a significant difference between \ama3 and \ama4.
 Comparing $\epsilon=0.1$ (left) and $\epsilon=0.5$ (right),
 we noticed that the entire point cloud is moved toward the top right in $\epsilon=0.5$,
 indicating that the networks trained with $\epsilon=0.1$ tend to be more accurate.
 }
 \label{fig:action-model}
\end{figure}

\refig{fig:action-model} shows these metrics for the best hyperparameter configuration that achieved the best ELBO.
For those configurations, \ama3 and \ama4 models obtained comparable accuracy.
If we compare \ama3 and \ama4 having the same hyperparameter, the result seems slightly in favor of \ama4.

Comparing the results of different priors,
$\epsilon=0.1$ configurations (\refig{fig:action-model}, left) tend to have better successor prediction accuracy
than $\epsilon=0.5$ (\refig{fig:action-model}, right).
This is presumably because the state instability causes a non-deterministic state transition
that cannot be captured well by the BTL mechanism because BTL assumes STRIPS-compatible state transitions.

Again, note that neither the best ELBO or the best successor prediction accuracy alone
determine the likelihood of success in solving visual planning problems.
For example, in a collapsed latent space, successor prediction is quite easy
because not a single latent space bit will change its value.
Also, even if the overall accuracy (ELBO) is good,
the planning is not likely to succeed
if the network sacrifices successor prediction accuracy for reconstruction accuracy.

\subsection{Violation of Monotonicity in BTL}
\label{sec:xor}

As discussed in \refsec{sec:ama3-effect-extraction},
Batch Normalization used as a ``continualizer'' $m$ in BTL may violate the monotonicity of $m$.
This causes an action with \emph{XOR semantics} which always flips the value,
which is not directly expressible in the STRIPS semantics.

To quantify the harmful effect of these bits with XOR semantics,
we first counted the number of such violations averaged over all actions.
As we see in \reftbl{tab:xor}, the number of such bits is small compared to the
entire representation size.
We also compared the number of actions before and after compiling these XOR semantics away.
While the increase of STRIPS action is exponential to the number of violations in the worst case,
the empirical increase tends to be marginal because the number of violations was small for each action,
except for one instance of Sokoban in \ama4.

While the Sokoban result from the best ELBO hyperparameter of \ama4
resulted in a relatively large number of monotonicity violations,
this is not always the case across hyperparameters.
\reftbl{tab:xor} (Right) plots ELBO and the compiled number of actions, $A_2$,
which varies significantly between hyperparameters despite having a similar ELBO.
This is another case demonstrating that the ELBO alone does not necessarily characterize the detrimental effect on search performance.

Note that the increase depends on whether the violations are concentrated in one action.
For example, if 5 violations are found in one action, the action is compiled into $2^5=32$ variants.
However, if 1 violation is found in 5 separate actions, it only adds 5 more actions.

\begin{table}[htbp]
\centering
\begin{minipage}{0.6\linewidth}
\begin{adjustbox}{width=\linewidth}
\begin{tabular}{|r|cc|c|cc|}
\hline
 & \multicolumn{ 2}{c|}{XOR bits} &  & \multicolumn{ 2}{c|}{Actions} \\
domain & effects & precondition & $F$ & $A_1$ & $A_2$ \\
\hline
\multicolumn{ 6}{|c|}{\ama3} \\
\hline
Blocks & 0.00 & - & 300 & 1908 & 1917 \\
LOut & 0.00 & - & 50 & 877 & 877 \\
Twisted & 0.00 & - & 50 & 919 & 919 \\
Mandrill & 0.00 & - & 300 & 3265 & 3265 \\
MNIST & 0.09 & - & 300 & 974 & 1065 \\
Sokoban & 0.29 & - & 50 & 737 & 1028 \\
\hline
\multicolumn{ 6}{|c|}{\ama4} \\
\hline
Blocks & 0.00 & 0.02 & 100 & 2317 & 2364 \\
LOut & 0.00 & 0.00 & 300 & 941 & 941 \\
Twisted & 0.00 & 0.00 & 300 & 1069 & 1069 \\
Mandrill & 0.00 & 0.00 & 300 & 3450 & 3469 \\
MNIST & 0.19 & 0.10 & 300 & 1263 & 1664 \\
Sokoban & 2.70 & 2.81 & 300 & 1115 & 47451 \\
\hline
\end{tabular}
\end{adjustbox}
\end{minipage}
\begin{minipage}{0.39\linewidth}
 \includegraphics[width=\linewidth]{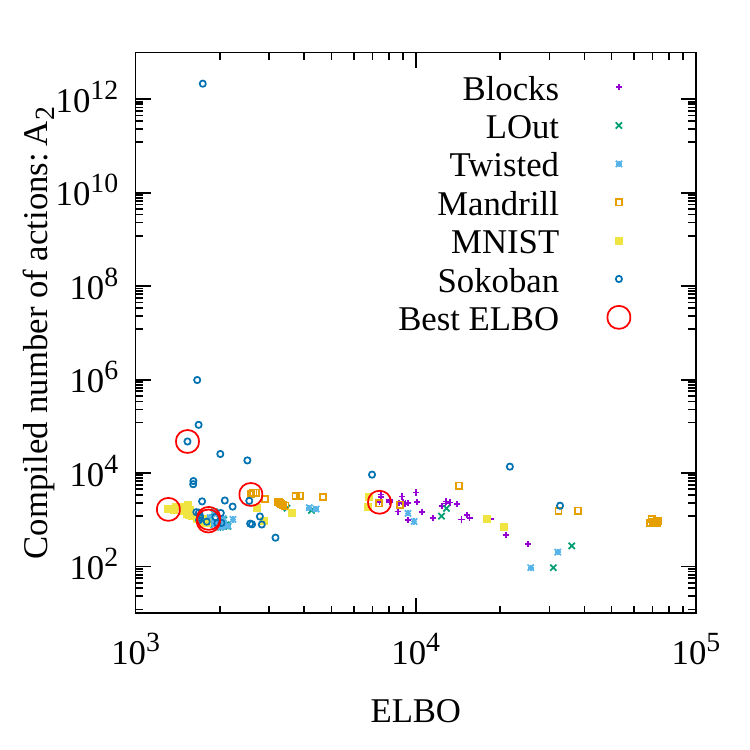}
\end{minipage}
\caption{
The XOR columns show the number of bits with XOR semantics averaged across actions
for a hyperparameter configuration that achieved the best ELBO in each domain.
$A_1$ column shows the number of actions before compiling XOR bits away,
and $A_2$ shows the number after the compilation.
Note that $A_1 \leq A = 6000$, where $A$ is the hyperparameter that specifies the length of one-hot vector $\action$,
thus serves as the maximum number of action labels that can be learned by the neural network.
In \ama3, we observed that the number of monotonicity violations is small,
and thus the empirical increase of the number of actions due to the compilation is marginal.
In \ama4, this is also the case except for the Sokoban domain that resulted in a larger number of XOR bits.
However, note that this is not always the case in all hyperparameters;
In the figure on the right,
we plotted the number of compiled actions in the $y$-axis against the ELBO in the $x$-axis.
The number of compiled actions varies significantly for a similar ELBO.
}
\label{tab:xor}
\end{table}

\subsection{Summary Statistics of the Learned PDDL Representation}

Finally, \reftbl{tab:statistics-summary} summarizes the statistics of
the PDDL obtained from our final model, AMA$_4^+$, with $\bern(\epsilon=0.1)$ prior.
This table shows
the number of effective bits (\refsec{sec:stability}),
the number of compiled actions (\refsec{sec:xor}),
average state differences between current and successor states,
and the average number of
delete/add effects, positive/negative preconditions.

State differences are the average number of different bits
between the current and the successor states over the dataset, i.e., $\E_{i,f} |\zbefore_f-\zafter_f|$.
It explains why
the PDDL files generally contain larger numbers of delete effects and negative preconditions:
The system recognized static predicates that take false values and added it to the preconditions/effects.

\begin{table}[htbp]
 \centering
\begin{tabular}{|l|ccc|cc|cc|}
\hline
domain & Propositions     & Actions        & State       & \multicolumn{2}{c|}{Effects}& \multicolumn{2}{c|}{Preconditions} \\
       & (effective bits) & (xor-compiled) & differences & $\adde$ & $\dele$ & $\posp$ & $\negp$ \\
\hline
Blocks   & 100 & 2366  & 23.88 & 20.62 & 21.44  & 19.96 & 22.04  \\
LOut     & 35  & 941   & 0.86  & 2.67  & 146.50 & 2.74  & 146.49 \\
Twisted  & 63  & 1069  & 2.08  & 10.03 & 145.35 & 10.06 & 145.33 \\
Mandrill & 299 & 3469  & 8.90  & 4.79  & 16.44  & 4.82  & 16.43  \\
MNIST    & 225 & 1665  & 7.46  & 5.85  & 53.08  & 5.83  & 53.63  \\
Sokoban  & 103 & 47495 & 2.84  & 5.00  & 219.29 & 4.93  & 220.03 \\
\hline
\end{tabular}
 \caption{
 A summary of the statistics of the PDDL files obtained from the AMA$_4^+$ model for each domain
 (hyperparameter with the best ELBO and with $\bern(\epsilon=0.1)$ prior).
 }
 \label{tab:statistics-summary}
\end{table}

\section{Planning Performance}
\label{sec:planning-evaluation}

We next evaluate the PDDL models produced by the networks
using randomly generated problem instances for each domain.
The objectives of the comparisons made in this section are threefold:
\begin{enumerate}
 \item Verifying the effectiveness of the preconditions generated by Bidirectional Cube-Space AE (\ama4)
       over the ad-hoc preconditions generated by Cube-Space AE (\ama3).
 \item Verifying the effect of \lsota heuristics on the search performance in the latent space generated by \ama3/\ama4.
 \item Verifying the effect of improved symbol stability on the planning performance,
\end{enumerate}

\subsection{Experimental Setup}
\label{sec:planning-setup}

We ran the off-the-shelf planner Fast Downward on the PDDL generated by our system.
To evaluate the quality of the domains, we similarly generate multiple problems
by encoding an initial and a goal state image $(\init,\goal)$ which are randomly generated using domain-specific code.
The images are normalized to mean-0, variance-1 using the statistics of the training dataset,
i.e., they are shifted and scaled by the mean and the variance of the training dataset.
Each normalized image is encoded by the encoder, then is converted into a PDDL encoding of the initial state and the goal state.
Formally,
when $\init\in [0,255]^{H,W,C}$ is an initial state image of width $H$, width $W$ and color channel $C\in\braces{1,3}$,
and $\vmu=\E_{i,j} [\xbefore[i,j]]$ and $\vsigma^2=\Var_{i,j} [\xbefore[i,j]]$ are the mean and the variance of the training dataset,
the latent propositional initial state vector is $\zinit=\BC(\encode(\frac{\init-\vmu}{\vsigma}))$,
just as all training is performed on the normalized dataset.

In 8-Puzzle, 15-Puzzle, LightsOut and Twisted,
we used a randomly sampled initial state image $\init$ and a fixed complete goal state $\goal$.
The goal states are those states that are normally considered to solve the puzzles,
e.g., for a Mandrill 15-Puzzle, the goal state is a state where all tiles are ordered correctly so that the whole configuration
recovers the original photograph of a Mandrill.
Initial states $\init$ are sampled from the frontier of a Dijkstra search, which was run backward from the goal.
The search stops when it exhausts the plateau at a specific $g$-value, at which point
the shortest path length from the goal state is obtained for each of those states.
We then randomly select a state from this plateau.
In each domain, we generated 20 instances for $g=7$ and $g=14$, resulting in 40 instances per domain.\footnote{$g$ is the measure of distance from the initial state to a node is the search graph.}

In 8-Puzzle,
we also added instances whose goal states are randomly generated using domain-specific code,
and the initial states are sampled with $g=7$ and $g=14$ thresholds.
The purpose of evaluating them is to see whether the system is generalized over the goals.

In Sokoban, the goal-based sampling approach described above does not work because the problem is not reversible.
We instead sampled goal states from the initial state of p006-microban-sequential instance
with $g=7$ and $g=14$ thresholds.
These goal states do not correspond to the goal states of p006-microban-sequential which solve the puzzle.

In Blocksworld,
we randomly generated an initial state and performed a 7-step or 14-step random walk to generate a goal state.
The number of steps does not correspond to the optimal plan length.

We tested
\astar with blind heuristics,
Landmark-Cut \cite[\lmcut]{Helmert2009},
Merge-and-Shrink (\mands) \cite{HelmertHHN14}, and
the first iteration of the satisficing planner LAMA \cite{richter2010lama}.
Experiments are run with the 10 minutes time limit and 8GB memory limit.
We tested the PDDL generated by \ama3 and \ama4 models,
each trained with different latent space priors $\bern(\epsilon=0.5)$ and $\bern(\epsilon=0.1)$.

We counted the number of instances in which a solution was \textbf{found}.
However, the solution found by our system is not always correct
because the correctness of the plan is guaranteed only with respect to the PDDL model
--- if the symbolic mapping produced by the neural network is incorrect,
the resulting visualization of the symbolic plan does not correspond to a realistically correct visual plan.
To address this issue, we wrote a domain-specific plan validator for each domain which
heuristically examines the visualized results and
we counted the number of plans which are empirically \textbf{valid}.
Finally, we are also interested in how often the solution is \textbf{optimal} out of the valid plans.
In domains where we generated the instances with Dijkstra-based sampling method,
we compared the solution length with the $g$ threshold which was used to generate the instance.

In the comparisons made in the following sections, we select the hyperparameter configuration
which maximizes the metric of interest (e.g., the number of valid solutions or the number of optimal solutions).
This selection is performed on a per domain basis.

\subsection{Bidirectional Models Outperform Unidirectional Models}
\label{sec:ama3-ama4}

We evaluated each hyperparameter configuration of \ama3 and \ama4 on the problem instances we generated.
We verified the solutions and \reftbl{tab:vanilla-ama3-vs-ama4} shows the numbers
obtained from configurations that achieved the highest number of valid solutions.
The result shows that \ama4 outperforms \ama3,
indicating that the precondition generated by the complete state regression in \ama4 is more effective
than those generated by an ad-hoc method in \ama3.
All results are based on $\bern(\epsilon=0.1)$ prior for the latent representation
(i.e., $\epsilon$ is not considered as a hyperparameter).

In particular, we note the significant jump in (1) coverage on arguably the most visually challenging domains (Blocks and Sokoban); and (2) the ratio of valid plans that are also optimal (\ama4 struggled with only MNIST in this regard).

\begin{table}[htbp]
\centering
\begin{adjustbox}{width=\linewidth,keepaspectratio}
\begin{tabular}{|r|ccc|ccc|ccc|ccc|}
\hline
 & \multicolumn{ 3}{c|}{Blind} & \multicolumn{ 3}{c|}{LAMA} & \multicolumn{ 3}{c|}{LMCut} & \multicolumn{ 3}{c|}{M\&S} \\
\hline
 & \textbf{found} & \textbf{valid} & {\textbf{optimal}} & {\textbf{found}} & {\textbf{valid}} & {\textbf{optimal}} & {\textbf{found}} & {\textbf{valid}} & {\textbf{optimal}} & {\textbf{found}} & {\textbf{valid}} & {\textbf{optimal}} \\
\hline
\multicolumn{ 13}{|c|}{\ama3} \\
\hline
Blocks & 1 & 1 & - & 1 & 1 & - & 1 & 1 & - & 1 & 1 & - \\
LOut & 40 & 40 & 35 & 40 & 39 & 0 & 20 & 20 & 18 & 40 & 40 & 35 \\
Twisted & 40 & 40 & 36 & 40 & 40 & 0 & \textbf{23} & 20 & 18 & 40 & 40 & 16 \\
Mandrill & \textbf{38} & \textbf{38} & 20 & \textbf{38} & \textbf{38} & 9 & 38 & \textbf{38} & 18 & 40 & \textbf{40} & 26 \\
MNIST & 39 & 39 & 5 & \textbf{39} & \textbf{39} & 4 & 39 & 39 & 5 & 39 & 39 & 5 \\
Random MNIST & \textbf{39} & \textbf{39} & 4 & \textbf{39} & \textbf{36} & 4 & \textbf{39} & \textbf{39} & 4 & \textbf{39} & \textbf{39} & 4 \\
Sokoban & 14 & 14 & 12 & 14 & 14 & 12 & 14 & 14 & 12 & 14 & 14 & 12 \\
\hline
\textbf{Total} & 211 & 211 & 112 & 211 & 207 & 29 & 174 & 171 & 75 & 213 & 213 & 98 \\
\hline
\multicolumn{ 13}{|c|}{\textbf{\ama4}} \\
\hline
Blocks & \textbf{33} & \textbf{32} & - & \textbf{19} & \textbf{19} & - & \textbf{34} & \textbf{34} & - & \textbf{34} & \textbf{33} & - \\
LOut & 40 & 40 & \textbf{40} & 40 & \textbf{40} & \textbf{1} & 20 & 20 & \textbf{20} & 40 & 40 & \textbf{40} \\
Twisted & 40 & 40 & \textbf{40} & 40 & 40 & \textbf{1} & 20 & 20 & \textbf{20} & 40 & 40 & \textbf{40} \\
Mandrill & 25 & 23 & \textbf{23} & 20 & 15 & \textbf{11} & 38 & 30 & \textbf{30} & 40 & 32 & \textbf{32} \\
MNIST & \textbf{40} & 39 & \textbf{6} & 35 & 35 & \textbf{16} & \textbf{40} & 39 & \textbf{6} & \textbf{40} & 39 & \textbf{6} \\
Random MNIST & 36 & 34 & \textbf{11} & 32 & 32 & \textbf{5} & 35 & 32 & \textbf{11} & 36 & 33 & \textbf{11} \\
Sokoban & \textbf{40} & \textbf{39} & \textbf{38} & \textbf{40} & \textbf{31} & \textbf{21} & \textbf{40} & \textbf{38} & \textbf{37} & \textbf{40} & \textbf{39} & \textbf{38} \\
\hline
\textbf{Total} & \textbf{254} & \textbf{247} & \textbf{158} & \textbf{226} & \textbf{212} & \textbf{55} & \textbf{227} & \textbf{213} & \textbf{124} & \textbf{270} & \textbf{256} & \textbf{167} \\
\hline
\end{tabular}
\end{adjustbox}
 \caption{Planning results.
We highlight the numbers in \textbf{bold} when \ama4 and \ama3 outperforms each other, except ties.
The results indicate that the regression-based precondition generation method in \ama4 is effective.
}
\label{tab:vanilla-ama3-vs-ama4}
\end{table}

\subsection{Classical Planning Heuristics Reduce the Search Effort in the Latent Space}
\label{sec:vs-blind}

\begin{figure}[p]
 \centering
 \includegraphics[width=0.47\linewidth]{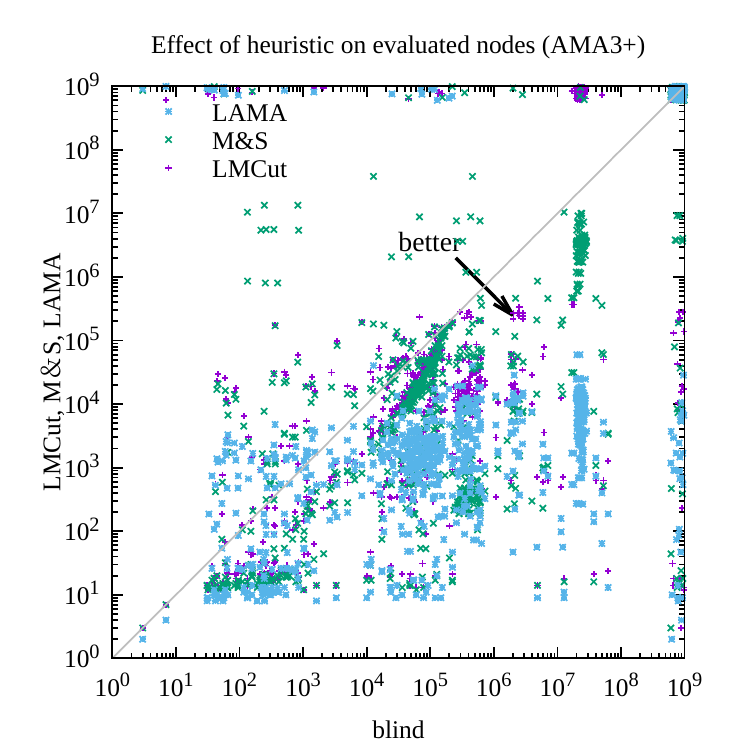}
 \includegraphics[width=0.47\linewidth]{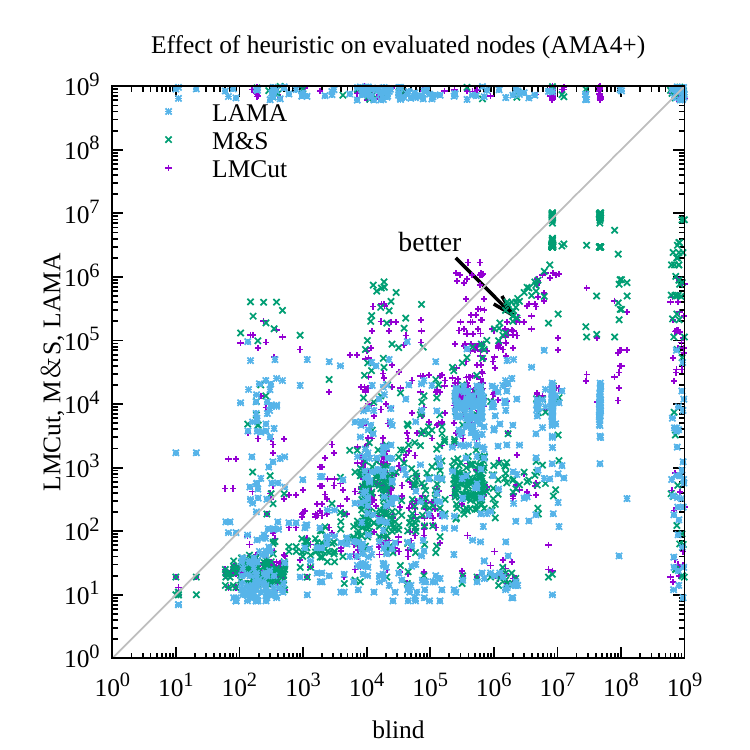}\\
 \includegraphics[width=0.47\linewidth]{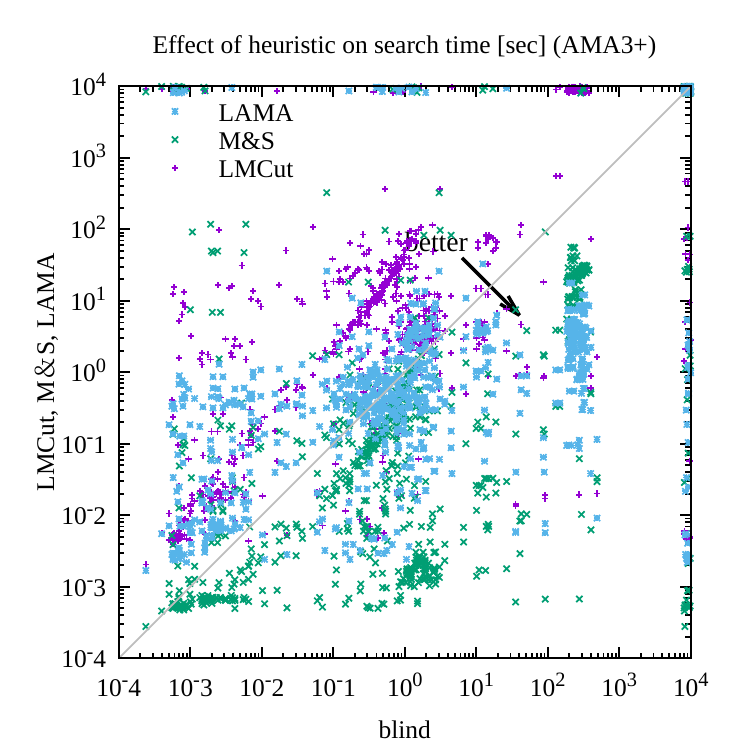}
 \includegraphics[width=0.47\linewidth]{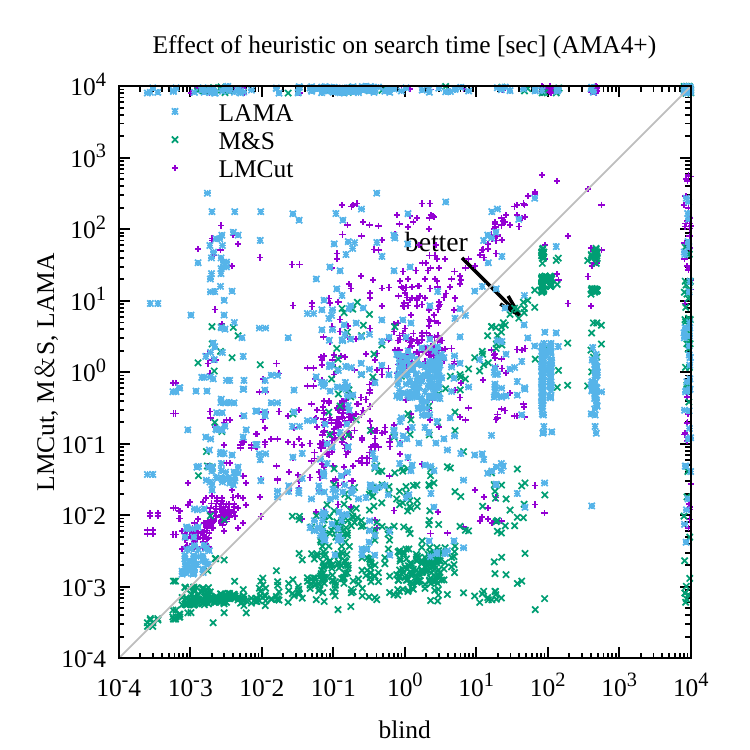}
 \caption{
 The left column shows \ama3 and the right column shows \ama4.
 The top row shows the number of node evaluations for $\blind$ heuristics ($x$-axis) and $\lmcut$, $\mands$ heuristics as well as LAMA ($y$-axis),
 for instances solved and validated by both.
 Unsolved instances are placed on the border.
 \sota heuristics developed in classical planning literature significantly reduces the search effort.
 The bottom row shows the search time.
 \mands still outperforms \blind.
 Due to the higher cost-per-node, \lmcut tends to consume more search time compared to \blind.
 Similarly, LAMA seems to struggle in easy instances. However, it tends to behave better in more difficult instances.
 For a plot colored by domains rather than heuristics, see appendix (\refsec{sec:heuristics-domainwise}).
 }
 \label{fig:heuristics}
\end{figure}

Domain-in\-dep\-endent heuristics in the planning literature are traditionally evaluated on hand-coded benchmarks,
 e.g., International Planning Competition instances \cite{McDermott00},
which contain a wide variety of domains and are regarded as a representative subset of the real-world tasks.
As a result, they assume a certain general class of structures that are commonly found in domains made by humans,
such as serializable subgoals and abstraction \cite{Korf85a}.
However, the latent space domains derived from raw pixels and neural networks may have completely different characteristics that could render them useless.
In fact, blind heuristic can sometimes outperform well-known heuristics
on some IPC domains (e.g., \pddl{floortile}) specifically designed to defeat their efficacy \cite{pdb-tribute}.
Also,
even though the image-based 8-puzzle domains we evaluate \emph{correspond} to the  8-puzzle domain used in the combinatorial search and classical planning literature,
the latent representations of the 8-puzzle states may be unlike any standard PDDL encoding of the 8-puzzle written by humans.
While the symbolic representation acquired by \latentplanner captures the state space graph of the domain,
the propositions in the latent space do not necessarily correspond to conceptual propositions in a natural, hand-coded PDDL model.
Thus, there is little \emph{a priori} reason to believe that the standard heuristics will be effective until we evaluate them.

We evaluated the effect of \lsota heuristic functions in the latent space generated by our neural networks.
\refig{fig:heuristics} compares node evaluations and search time between blind search and \lmcut, \mands heuristics as well as LAMA.
For each domain, we select the hyperparameter configuration with the largest number of valid solutions.
Results are based on $\bern(\epsilon=0.1)$.

Node evaluation plots confirm the overall reduction in the search effort.
In contrast, the results on search time are mixed.
\mands reduces the search time by an order of magnitude,
while \lmcut tends to be slow due to the high evaluation cost per node.
LAMA tends to be faster than a blind search in more difficult instances,
though a blind search is faster in easier instances that require less search ($<10$ seconds).

Despite the increased runtime, the reduced node evaluation
strongly indicates that either the models
we are inferring contain similar properties as the human-designed PDDL models,
or that powerful, domain-independent planning heuristics can induce effective search guidance even in less human-like PDDL models.

\subsection{Stable Symbols Contributes to the Better Performance}
\label{sec:vs-nozsae}

\refsec{sec:unstable} discussed three harmful effects of unstable symbols:
\textbf{(1)} disconnected search space,
\textbf{(2)} having many variations of latent states for a single real-world state,
\textbf{(3)} hyperparameter tuning for improving symbol stability.
We already evaluated the issue \textbf{(3)} in the previous sections,
where we verified that $\bern(\epsilon=0.1)$ automatically finds a latent space with smaller effective bits when a large $F$ is given.
In this section, we proceed to verify the effect of $\bern(\epsilon=0.1)$ on the first and the second issue.

Issue \textbf{(1)} (disconnected search space) would result in more \emph{exhausted instances},
i.e., the number of instances where Fast Downward exhausted the state space without finding a solution.
It would also result in a lower success ratio,
given that we set a certain time limit on the planner and not all unsolvable instances are detected.

Issue \textbf{(2)} (many variations of latent states) would result in more
\emph{suboptimal visualized plans that are optimal in the latent space / PDDL encoding.}
There are several possible reasons that a visualized solution becomes suboptimal.

The first and most obvious one is that satisficing planner configuration (LAMA) could return a suboptimal latent space solution,
which results in a suboptimal visual solution. Since this is trivial, we do not include LAMA during optimality evaluation.

Second, if a certain action is not correctly learned by the neural network
(e.g., missing or having unnecessary preconditions and effects), the optimal solution with regard to the PDDL model
may not be in fact optimal when visualized:
The planner may not be able to take certain otherwise valid transitions, thus resulting in a longer solution;
Or an action effect may alter the values of some bits that do not affect the image reconstruction,
and the planner tries to fix it with additional actions.
As discussed in \refex{example:stability-effect}, accurate successor prediction is a form of symbol stability.
In \refsec{sec:effect}, we already observed that $\epsilon=0.1$ tend to predict successors better than $\epsilon=0.5$
due to its stable state representation.

The third case is unstable init/goal states: When symbol instability is significant,
a noisy goal state image $\goal$ may map to multiple different latent states $\zgoal,\vz'^{G}$ depending on the input noise.
The encoded initial state $\zinit$ may be closer to $\vz'^{G}$ than to $\zgoal$ specified in PDDL,
thus reaching $\zgoal$ may require an extra number of steps, resulting in a suboptimal visual solution.
The same mechanism applies to the initial state too.

\begin{figure}
 \includegraphics[width=0.16\linewidth]{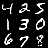}
 \includegraphics[width=0.16\linewidth]{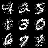}
 \includegraphics[width=0.16\linewidth]{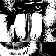}
 \includegraphics[width=0.16\linewidth]{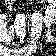}
 \includegraphics[width=0.16\linewidth]{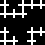}
 \includegraphics[width=0.16\linewidth]{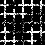}
 \includegraphics[width=0.16\linewidth]{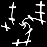}
 \includegraphics[width=0.16\linewidth]{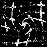}
 \includegraphics[width=0.16\linewidth]{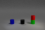}
 \includegraphics[width=0.16\linewidth]{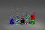}
 \includegraphics[width=0.16\linewidth]{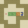}
 \includegraphics[width=0.16\linewidth]{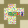}
 \caption{
Examples of initial state images before and after being corrupted by Gaussian noise.
The amount of noise is noticeable but is not as strong as making it difficult to understand the scene.
Note that unlike \refig{fig:example}, we used low-resolution images for Blocksworld and Sokoban
that were used for training and testing.
}
 \label{fig:noise}
\end{figure}

To investigate the effect of symbol stability,
we additionally evaluated our system on a set of initial/goal state images that are corrupted by a Gaussian noise $\mathcal{N}(0,1)$.
We added the noise to the images that are already normalized to mean-0, variance-1
using the statistics from the training dataset (see \refsec{sec:planning-setup}).
As a result, when de-normalized for visualization in \refig{fig:noise}, the noise appears missing in the static regions,
which have little variations across the dataset.

\begin{figure}[p]
 \centering
 \includegraphics[width=0.47\linewidth]{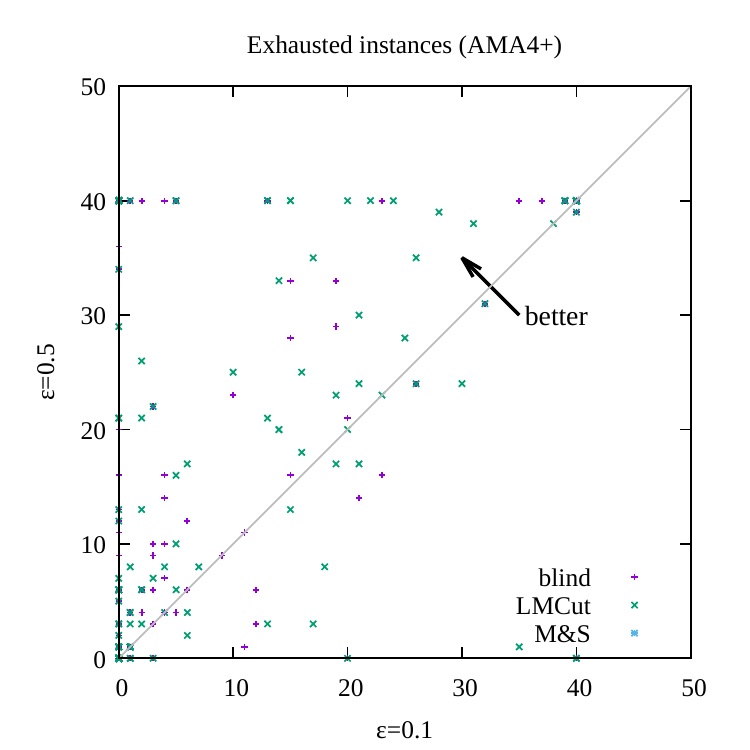}
 \includegraphics[width=0.47\linewidth]{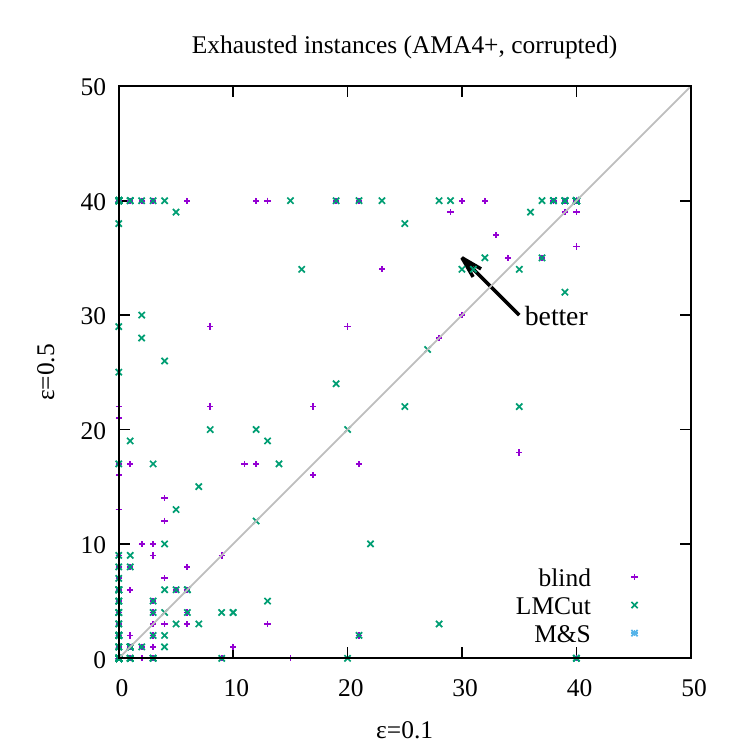}\\
 \caption{
 The left figure shows the noiseless instances, and the right figure shows the result from corrupted input images.
 Each point represents a tuple $(\text{domain}, \text{heuristic}, F,\beta_1,\beta_3)$.
 We evaluated different heuristics because they have different performances in terms of exhaustively searching the space within the time limit.
 For each point, each axis represents the number of unsolvable instances among the 40 instances in the domain,
 which are discovered by Fast Downward by exhaustively searching the state space.
 $x$-axis represents \ama4 networks with $\epsilon=0.1$ prior, and $y$-axis the $\epsilon=0.5$ prior.
 $\epsilon=0.1$ resulted in significantly fewer unsolvable instances.
 For a plot colored by domains rather than heuristics, see appendix (\refsec{sec:exhaust-domainwise}).
 }
 \label{fig:exhaust-vs-nozsae}
\end{figure}

\begin{table}[p]
\centering
\begin{adjustbox}{width=\linewidth,keepaspectratio}
\begin{tabular}{|r||*{4}{rr|}|*{4}{rr|}}
\hline
 & \multicolumn{ 8}{c||}{Clean input images} & \multicolumn{ 8}{c|}{Corrupted input images} \\
 & \multicolumn{ 2}{c|}{Blind} & \multicolumn{ 2}{c|}{LAMA} & \multicolumn{ 2}{c|}{LMCut} & \multicolumn{ 2}{c||}{M\&S}
 & \multicolumn{ 2}{c|}{Blind} & \multicolumn{ 2}{c|}{LAMA} & \multicolumn{ 2}{c|}{LMCut} & \multicolumn{ 2}{c|}{M\&S} \\
$\epsilon=$ & $0.1$ & $0.5$ & $0.1$ & $0.5$ & $0.1$ & $0.5$ & $0.1$ & $0.5$ & $0.1$ & $0.5$ & $0.1$ & $0.5$ & $0.1$ & $0.5$ & $0.1$ & $0.5$ \\
\hline
\multicolumn{ 17}{|c|}{\ama3} \\
\hline
Blocks       & 1           & 1           & 1           & 1           & 1           & 1           & 1           & 1           & \textbf{1}  & 0           & \textbf{1}  & 0           & \textbf{1}  & 0           & \textbf{1}  & 0           \\
LOut         & \textbf{40} & 27          & 39          & \textbf{40} & 20          & 20          & \textbf{40} & 28          & \textbf{40} & 30          & \textbf{39} & \textbf{33} & \textbf{20} & 19          & \textbf{40} & 31          \\
Twisted      & 40          & 40          & 40          & 40          & 20          & 20          & 40          & 40          & 40          & 40          & 40          & 40          & 20          & 20          & 40          & 40          \\
Mandrill     & 38          & \textbf{39} & 38          & \textbf{40} & 38          & 38          & 40          & 40          & \textbf{39} & 37          & \textbf{39} & 33          & \textbf{37} & \textbf{35} & \textbf{39} & 38          \\
MNIST        & 39          & \textbf{40} & 39          & \textbf{40} & 39          & \textbf{40} & 39          & \textbf{40} & 36          & \textbf{40} & 36          & \textbf{40} & 36          & \textbf{40} & 36          & \textbf{40} \\
Random MNIST & \textbf{39} & 38          & 36          & \textbf{38} & \textbf{39} & 38          & \textbf{39} & 38          & \textbf{36} & 36          & 33          & \textbf{36} & \textbf{36} & 36          & \textbf{36} & 36          \\
Sokoban      & \textbf{14} & 1           & \textbf{14} & 1           & \textbf{14} & 1           & \textbf{14} & 1           & \textbf{13} & 1           & \textbf{13} & 1           & \textbf{13} & 1           & \textbf{13} & 1           \\
\hline
\textbf{total} & \textbf{211} & 186 & \textbf{207} & 200 & \textbf{171} & 158 & \textbf{213} & 188 & \textbf{205} & 184 & \textbf{201} & 183 & \textbf{163} & 151 & \textbf{205} & 186 \\
\hline
\multicolumn{ 17}{|c|}{\ama4} \\
\hline
Blocks       & \textbf{32} & 31          & 19          & \textbf{23} & \textbf{34} & 29          & \textbf{33} & 31          & 25          & \textbf{27} & 10          & \textbf{12} & 24          & \textbf{26} & 24          & \textbf{27} \\
LOut         & 40          & 40          & 40          & 40          & 20          & 20          & 40          & 40          & 40          & 40          & 40          & 40          & 20          & \textbf{23} & 40          & 40          \\
Twisted      & 40          & 40          & 40          & 40          & 20          & 20          & 40          & 40          & 40          & 40          & 40          & 40          & 20          & 20          & 40          & 40          \\
Mandrill     & 23          & \textbf{24} & \textbf{15} & 10          & 30          & \textbf{33} & 32          & \textbf{36} & \textbf{19} & 17          & 7           & \textbf{10} & 27          & \textbf{28} & 26          & \textbf{30} \\
MNIST        & \textbf{39} & 37          & \textbf{35} & 26          & \textbf{39} & 36          & \textbf{39} & 34          & 36          & \textbf{37} & \textbf{26} & 21          & \textbf{36} & 36          & \textbf{36} & 34          \\
Random MNIST & 34          & \textbf{37} & \textbf{32} & 19          & 32          & \textbf{35} & 33          & \textbf{35} & 34          & \textbf{36} & \textbf{28} & 19          & 32          & \textbf{34} & 33          & \textbf{34} \\
Sokoban      & \textbf{39} & 33          & 31          & \textbf{34} & \textbf{38} & 31          & \textbf{39} & 32          & \textbf{39} & 32          & 31          & \textbf{33} & \textbf{38} & 31          & \textbf{39} & 31          \\
\hline
\textbf{total} & \textbf{247} & 242 & \textbf{212} & 192 & \textbf{213} & 204 & \textbf{256} & 248 & \textbf{233} & 229 & \textbf{182} & 175 & 197 & \textbf{198} & \textbf{238} & 236 \\
\hline
\end{tabular}
\end{adjustbox}
 \caption{
The number of instances where a valid solution is found.
Each number is the largest number among different hyperparameter configurations ($F,\beta_1,\beta_3$) in a single domain.
Networks trained with $\bern(\epsilon=0.1)$ latent space prior outperformed the networks trained with $\bern(\epsilon=0.5)$.
Numbers are highlighted in \textbf{bold} for the better latent space prior, except ties.
}
\label{tab:vs-nozsae}
\end{table}

\begin{table}[p]
\begin{adjustbox}{width=\linewidth,keepaspectratio}
\begin{tabular}{|r|*{3}{|*{3}{rr|}}}
\hline
 & \multicolumn{ 6}{c||}{Clean input images} & \multicolumn{ 6}{c||}{Corrupted input images} & \multicolumn{ 6}{c|}{Difference due to noise} \\
 & \multicolumn{ 2}{c|}{Blind}  & \multicolumn{ 2}{c|}{LMCut} & \multicolumn{ 2}{c||}{M\&S}
 & \multicolumn{ 2}{c|}{Blind}  & \multicolumn{ 2}{c|}{LMCut} & \multicolumn{ 2}{c||}{M\&S}
 & \multicolumn{ 2}{c|}{Blind}  & \multicolumn{ 2}{c|}{LMCut} & \multicolumn{ 2}{c|}{M\&S} \\
$\epsilon=$
& $0.1$ & $0.5$ & $0.1$ & $0.5$ & $0.1$ & $0.5$
& $0.1$ & $0.5$ & $0.1$ & $0.5$ & $0.1$ & $0.5$
& $0.1$ & $0.5$ & $0.1$ & $0.5$ & $0.1$ & $0.5$ \\
\hline
\multicolumn{19}{|c|}{\ama3} \\
\hline
LOut             & .88 & .93  & .95 & 1.00 & .88 & .93  & .88 & .90  & .95 & 1.00 & .88 & .90  & \textbf{.000}  & -.026         & .000          & .000          & \textbf{.000} & -.025         \\
Twisted          & .90 & .88  & .90 & .85  & .90 & .88  & .90 & .88  & .90 & .84  & .90 & .88  & .000           & .000          & \textbf{.000} & -.008         & .000          & .000          \\
Mandrill         & .81 & .77  & .83 & .79  & .79 & .75  & .79 & .70  & .84 & .74  & .79 & .68  & \textbf{-.016} & -.067         & \textbf{.005} & -.047         & \textbf{.005} & -.066         \\
MNIST            & .43 & .20  & .48 & .20  & .43 & .20  & .43 & .21  & .48 & .21  & .43 & .21  & .000           & \textbf{.007} & .000          & \textbf{.007} & .000          & \textbf{.007} \\
Random MNIST     & .24 & .28  & .24 & .28  & .24 & .28  & .29 & .28  & .28 & .28  & .28 & .28  & \textbf{.056}  & .002          & \textbf{.045} & .002          & \textbf{.045} & .002          \\
Sokoban          & .86 & 1.00 & .86 & 1.00 & .86 & 1.00 & .85 & 1.00 & .85 & 1.00 & .85 & 1.00 & -.011          & \textbf{.000} & -.011         & \textbf{.000} & -.011         & \textbf{.000} \\
\hline
\multicolumn{19}{|c|}{\textbf{\ama4}} \\
\hline
LOut             & 1.00 & 1.00 & 1.00 & 1.00 & 1.00 & 1.00 & 1.00 & 1.00 & 1.00 & 1.00 & 1.00 & 1.00 & .000          & .000  & .000          & .000          & .000          & .000  \\
Twisted          & 1.00 & 1.00 & 1.00 & 1.00 & 1.00 & 1.00 & 1.00 & 1.00 & 1.00 & 1.00 & 1.00 & 1.00 & .000          & .000  & .000          & .000          & .000          & .000  \\
Mandrill         & 1.00 & 1.00 & 1.00 & 1.00 & 1.00 & 1.00 & 1.00 & 1.00 & .96  & 1.00 & 1.00 & 1.00 & .000          & .000  & -.037         & \textbf{.000} & .000          & .000  \\
MNIST            & 1.00 & .78  & 1.00 & .81  & 1.00 & .79  & 1.00 & .78  & 1.00 & .81  & 1.00 & .79  & .000          & .000  & .000          & .000          & .000          & .000  \\
Random MNIST     & .91  & .92  & .94  & .91  & .94  & .91  & .95  & .92  & 1.00 & .91  & .95  & .91  & \textbf{.041} & -.002 & \textbf{.065} & -.003         & \textbf{.017} & -.003 \\
Sokoban          & .97  & .79  & .97  & .80  & .97  & .78  & .97  & .78  & .97  & .77  & .97  & .77  & \textbf{.000} & -.007 & \textbf{.000} & -.026         & \textbf{.000} & -.007 \\
\hline
\end{tabular}
\end{adjustbox}
\caption{
We calculated the rate of the number of optimal solutions found over the number of valid solutions found.
We then obtained the rate difference due to the input noise and compared the results between $\epsilon=0.1$ and $\epsilon=0.5$.
The results suggest that the input noise more negatively affects the optimality rate of $\epsilon=0.5$ than that of $\epsilon=0.1$.
}
\label{tab:opt-valid-ratio}
\end{table}

We first investigated the effect of \textbf{(1)}.
We evaluated \ama3 and \ama4 trained with different $\bern(\epsilon)$ priors.
\refig{fig:exhaust-vs-nozsae} plots the number of instances where Fast Downward exhaustively searched the state space
and failed to find a solution. The PDDL is generated by \ama4, and we compare the numbers between
$\bern(\epsilon=0.1)$ prior and the $\bern(\epsilon=0.5)$ prior.
We do not show \ama3 results because its preconditions are less accurate and ad-hoc,
and the precondition accuracy is crucial in this evaluation.
The figure clearly shows that more instances have disconnected search spaces when encoded by $\epsilon=0.5$ prior.

Next, to address the issue of insufficient runtime to prove the unsolvability,
\reftbl{tab:vs-nozsae} shows the number of valid solutions found.
For each domain, heuristic, and $\epsilon\in\braces{0.1,0.5}$,
we selected the hyperparameter which resulted in the largest number of valid solutions.
$\epsilon=0.1$ configuration outperforms $\epsilon=0.5$ on both clean and corrupted input images.

Recall that networks trained with different $\bern(\epsilon)$ priors had comparable results in terms of ELBO --- thus, ELBO (accuracy) alone is not
the sole factor that determines the likelihood of success in finding a solution.
In fact, $\epsilon=0.1$ configuration solves more instances than $\epsilon=0.5$ configuration does
in LightsOut (\ama3), Blocks, and Sokoban (\ama4) where $\epsilon=0.5$ had the better ELBO (\reftbl{tab:elbo}).

We next evaluate the effect of \textbf{(2)}, predicted to result in more suboptimal plans.
For each domain, heuristics, and prior distribution ($\epsilon\in\braces{0.1,0.5}$),
we selected the hyperparameter configuration which found the most valid optimal solutions.
Note that the absolute number of optimal plans is not a proper metric for evaluating the effect of \textbf{(1)}
because if a configuration finds fewer valid plans, the number of optimal plans found is also expected to decrease.
Instead, we measure the ratio between the number of optimal plans over the number of valid plans.
The results are shown in \reftbl{tab:opt-valid-ratio}.
\ama4 results show that $\epsilon=0.1$ is less affected by the noise than $\epsilon=0.5$ is.
While \ama3 results are mixed,
we attribute the mixed results to the lower optimality ratio of $\epsilon=0.5$ in MNIST (around 0.2, compared to around 0.43-0.48 in \ama4)
as well as the fact that \ama3 finds only a single valid instance of Sokoban, which was optimal (thus the optimality rate is 1).

\subsection{The Effect of Planner Preprocessing}

One may be interested in whether the planners are by themselves able to
prune more propositions/actions in the translation phase.
We compared the number of ``relevant propositions/actions'' as measured by Fast Downward translator
(translator variables / translator operators) against total number of propositions and actions.
For each domain,
we selected an \ama4 model that returned the largest number of valid solutions with M\&S
(we selected M\&S because it achieved the best overall performance).

As the total number of propositions,
we used the number of effective bits discussed in \refig{fig:effective},
rather than the number of propositions in a PDDL file,
because we do not filter propositions based on their effectiveness when writing them into a PDDL file.
As the total number of actions,
we directly used the number of actions in a PDDL file
because we already filter actions that is never mapped to by $\aaee$ in the tranining dataset (see AMA$_2$ section).
The number of actions in a PDDL file corresponds to $A_2$ after XOR compilation (\refsec{sec:ama3-effect-extraction}, \refsec{sec:xor}).

Results showed that Fast Downward did not reduce propositions/actions significantly.
For actions, over 95\% of original actions are retained.
The ratio $\frac{[\text{translator operators}]}{[\text{actions}]}$ was
blocks:99.8\%, lightsout:98.6\%, twisted:98.8\%, mandrill:96.2\%, mnist:95.4\%, sokoban:99.7\%.

The number of effective bits and the number of translator variables also closely matched:
The ratio $\frac{[\text{translator variables}]}{[\text{effective bits}]}$ was
blocks:100.0\%, lightsout:103.2\%, twisted:103.4\%, mandrill:100.0\%, mnist:100.0\%, sokoban:100.0\%.
The numbers can exceed 100\% because effective bits can be underestimated:
It is measured by encoding a dataset,
which contains only a subset of samples in the entire state space.

There were no variations in the translator variables / operators across problem instances
except instances where no solution was found.
This is unlike normal PDDL instances where these numbers differ due to different goal conditions.
In instances where no solution was found,
typically all actions and variables are detected as irrelevant and then removed.

\section{Related Work}
\label{sec:related}

In this section, we cover a wide range of related work from
purely connectionist approaches to purely symbolic approaches.
In between, there is a recent body of work which are
collectively called Neural/Neuro-Symbolic hybrid approaches.

\subsection{Previous Work on Action Model Acquisition and Symbol Grounding}

\begin{table}[tb]
\adjustbox{width=\linewidth}{
\begin{tabular}{r|llllll|lll|lll|l|l}
\toprule
 & \multicolumn{ 6}{c|}{Required Input} & \multicolumn{ 3}{c|}{Symbol Generation} & \multicolumn{ 3}{c|}{Symbol Grounding} & \multicolumn{1}{c|}{Optimization Method for} & Target \\
 & state ID & prop. & pred. & act. & obs. & noise & prop. & pred. & act. & prop. & pred. & act. & Likelihood Maximization  & language \\
\midrule
\cite{YangWJ07} & Y & Y & Y & Y & full & N & N & N & N & N & N & Y & MAXSAT & PDDL \\
\cite{MouraoZPS12} & Y & Y & Y & Y & \textbf{partial} & \textbf{Y} & N & N & N & N & N & Y & SVM & PDDL \\
\cite{zhuo2013action} & Y & Y & Y & Y & full & \textbf{Y} & N & N & N & N & N & Y & MAXSAT & PDDL \\
\cite{zhuo2019learning} & Y & Y & Y & Y & \textbf{dis.} & \textbf{Y} & N & N & N & N & N & Y & MAXSAT & PDDL \\
\midrule
\cite{cresswell2009acquisition} & Y & Y & Y & Y & full & N & N & N & N & N & N & Y & FSM & PDDL \\
\cite{CresswellG11} & Y & Y & Y & Y & full & N & N & N & N & N & N & Y & FSM & PDDL \\
\cite{CresswellMW13} & Y & Y & Y & Y & full & N & N & N & N & N & N & Y & FSM & PDDL \\
\cite{GregoryC15} & Y & Y & Y & Y & full & N & N & N & N & N & N & Y & FSM & PDDL \\
\midrule
\cite{aineto2018learning} & Y & Y & Y & \textbf{N} & \textbf{partial} & N & N & N & \textbf{Y} & N & N & Y & Classical Planning & PDDL \\
\midrule
\cite{bonet2020learning} & Y & \textbf{N} & \textbf{N} & Y & N/A & N & \textbf{Y} & \textbf{Y} & N & \textbf{Y} & \textbf{Y} & Y & MAXSAT & PDDL \\
\cite{rodriguez2021learning} & Y & \textbf{N} & \textbf{N} & Y & N/A & \textbf{Y} & \textbf{Y} & \textbf{Y} & N & \textbf{Y} & \textbf{Y} & Y & ASP & PDDL \\
\midrule
\cite{KonidarisKL14} & \textbf{N} & \textbf{N} & \textbf{N} & Y & full & N & \textbf{Y} & N & N & Y & N & Y & C4.5 & PDDL \\
\cite{KonidarisKL15} & \textbf{N} & \textbf{N} & \textbf{N} & Y & full & \textbf{Y} & \textbf{Y} & N & N & Y & N & Y & mixed & PPDDL \\
\cite{andersen2017active} & \textbf{N} & \textbf{N} & \textbf{N} & Y & full & \textbf{Y} & \textbf{Y} & N & N & Y & N & Y & mixed & PPDDL \\
\cite{KonidarisKL18} & \textbf{N} & \textbf{N} & \textbf{N} & Y & full & \textbf{Y} & \textbf{Y} & N & N & Y & N & Y & mixed & PPDDL \\
\midrule
\cite{BarbuNS10} & N & Y & Y & Y & N/A & Y & N & N & N & N & N & Y & ILP & Prolog \\
\cite{Kaiser12} & N & Y & Y & Y & N/A & Y & N & N & N & N & N & Y & ILP & GGP \\
\cite{ugur2015bottom} & \textbf{N} & \textbf{N} & \textbf{N} & Y & full & Y & \textbf{Y} & \textbf{Y*} & N & N & N & Y & mixed & PDDL \\
\cite{ahmetoglu2020deepsym} & \textbf{N} & \textbf{N} & \textbf{N} & Y & full & Y & \textbf{Y} & \textbf{Y*} & N & N & N & Y & mixed & PPDDL \\
\midrule
\cite{lindsay2017framer} & \multicolumn{ 6}{c}{\textbf{natural language}} & N & N & N & N & N & Y & k-medoid clustering & PDDL \\
\cite{miglani2020nltopddl} & \multicolumn{ 6}{c}{\textbf{natural language}} & N & N & N & N & N & Y & RL+LOCM2 & PDDL \\
\midrule
\textbf{Latplan} & \textbf{N} & \textbf{N} & \textbf{N} & \textbf{N} & full & \textbf{Y} & \textbf{Y} & N & \textbf{Y} & \textbf{Y} & N & Y & Deep Generative Model & PDDL \\
\bottomrule
\end{tabular}
}
\caption{
Comparison of various symbol grounding \& action model acquisition approaches.
Synopsis:
\emph{prop.}=propositional symbols,
\emph{pred.}=predicate symbols,
\emph{act.}=action symbols,
\emph{obs.}=observability,
\emph{dis.}=disordered observation.
(*) These systems obtain predicates, but they are limited to types (static predicates).
}
\label{tbl:ama-related}
\end{table}

\reftbl{tbl:ama-related} summarizes
a variety of action model acquisition approaches that have been proposed already \cite{jimenez2012review,arora2018review}.
They can be understood from three major axes:

\paragraph{Input:}
Traditionally, symbolic action learners tend to require a certain type
of human domain knowledge and have been situating itself merely as an
additional assistance tool for humans, rather than a system that builds knowledge from the scratch, e.g., from unstructured images.
Existing systems
\cite{YangWJ07,MouraoZPS12,zhuo2013action,zhuo2019learning,cresswell2009acquisition,CresswellG11,CresswellMW13,GregoryC15,aineto2018learning,bonet2020learning,rodriguez2021learning,KonidarisKL14,KonidarisKL15,andersen2017active,KonidarisKL18}
require some type of symbolic inputs as defined in \refsec{sec:processes4classical-palnning}
and exploit the categorical information and/or the structures provided by the input, as shown in the \reftbl{tbl:ama-related}.
The input format could vary, e.g.,
sequence of ground actions \cite{YangWJ07,zhuo2013action,zhuo2019learning,cresswell2009acquisition,CresswellG11,CresswellMW13,GregoryC15},
sequence of actions and intermediate states \cite{MouraoZPS12,KonidarisKL14,KonidarisKL15,KonidarisKL18,andersen2017active},
initial/goal states (and optionally plans and partial action models) \cite{aineto2018learning},
or a graph of atomic states (state IDs) and actions  \cite{bonet2020learning,rodriguez2021learning}.
Some systems accept varying amount of noise and partial observability
\cite{MouraoZPS12,zhuo2013action,zhuo2019learning,aineto2018learning,rodriguez2021learning}.
Approaches such as Framer \cite{lindsay2017framer} and cEASDRL \cite{miglani2020nltopddl}
obtain action models from natural language corpus.
They reuse the symbols found in the corpus, thus are grounding but not generating symbols,
though it contains a type of noise specific to natural language, e.g., case inflection and ambiguity.

\paragraph{Output:}
While existing systems do not generate all types of symbols,
some systems generate some symbols. For example,
\cite{bonet2020learning,rodriguez2021learning} generates predicate symbols from state IDs and action symbols,
\cite{aineto2018learning} generates action symbols from predicate symbols,
and
\cite{KonidarisKL14,KonidarisKL15,KonidarisKL18,andersen2017active} generates propositional symbols from action symbols.

\paragraph{Optimization Method for Likelihood Maximization:}

Although it is not explicitly acknowledged in some of the previous work,  
machine learning problems can be cast as optimization problems through Maximum-Likelihood Estimation framework (\refsec{sec:aevae}),
thus they are implicitly operating under the same framework as our paper,
with the difference being the implementation, differentiability, and joint training of the optimizer.

Each system tackles a learning problem using different optimization approaches.
The learning problem is encoded into Maximum Satisfiability \cite[MAX-SAT]{vazirani2013approximation} in \cite{YangWJ07,zhuo2013action,zhuo2019learning,bonet2020learning},
classical planning in \cite{aineto2018learning}, or
Answer Set Programming \cite[ASP]{gebser2011potassco} in \cite{rodriguez2021learning}.
Approaches based on Finite State Machines (FSM) \cite{cresswell2009acquisition,CresswellG11,CresswellMW13,GregoryC15}
tackle the learning problem by generating a large number of candidate FSMs and pruning those which contradict data,
effectively maximizing the number of data points that follow the set of FSMs,
i.e., $\argmax_{\text{FSMs}} p(x)$.

Many approaches \cite{KonidarisKL14,KonidarisKL15,KonidarisKL18,andersen2017active,ugur2015bottom,ahmetoglu2020deepsym,lindsay2017framer,miglani2020nltopddl}
combine several machine learning subprocedures, where each module individually optimize conditional likelihood,
which is similar to how AMA$_2$ learns a state representation and an action model separately.
Examples include
C4.5 Decision Tree in \cite{KonidarisKL14,ugur2015bottom},
Support Vector Machine, DBSCAN, and
Kernel density estimation in \cite{KonidarisKL15},
Bayesian sparse Dirichlet-categorical model and
Bayesian Hierarchical Clustering in \cite{andersen2017active},
sparse Principal Component Analysis in the robotic task using point clouds in \cite{KonidarisKL18},
Binary-Concrete VAE (also used in our paper in AMA$_1$ and AMA$_2$) \cite{ahmetoglu2020deepsym},
CoreNLP language model \cite{lindsay2017framer}, or
Reinforcement Learning and LOCM in \cite{miglani2020nltopddl}.
While they maximize individual conditional likelihoods (e.g., $q(\vz\mid\vx)$) separately,
they are not formally motivated as such, nor are they guaranteed to maximize the likelihood of observation.
In contrast, Latplan \ama3 and \ama4 jointly maximize the likelihood of observation through differentiable neural network.

\subsection{Neural Networks as Search Guidance, World Models}

Existing work
\cite{ArfaeeZH10,ArfaeeZH11,thayer2011learning,SatzgerK13,yoon2006learning,yoon2008learning,ferber-et-al-ecai2020,toyer2018action,shen2020learning}
has combined symbolic search and machine learning
by learning a function that provides search control knowledge,
e.g., domain-specific heuristic functions for
the sliding-tile puzzle and Rubik's Cube \cite{ArfaeeZH11},
classical planning \cite{SatzgerK13},
or the game of Go \cite{alphago}.
Some approaches use supervised learning from the precomputed target values (e.g., optimal solution cost) or expert traces,
while others use Reinforcement Learning \cite[RL]{sutton2018reinforcement}.

Model-free reinforcement learning has solved complex problems,
including video games in Arcade Learning Environment \cite[ALE]{bellemare2013arcade}
where the agent communicates with a simulator through images \cite[DQN]{dqn}.
The system avoids symbol grounding
by using the subsymbolic states directly and
assuming a set of action symbols that are given \emph{apriori} by a simulator.
For example, the ALE provides agents its discrete action labels such as 8-directional joysticks and buttons.
Another limitation of RL approaches is that
they require a carefully designed dense reward function to perform efficient exploration.
When the available reward function is sparse,
RL approaches suffer from sample efficiency issues
where the agents require millions or billions of interactions with the simulator to learn meaningful behavior.

Latplan does not learn domain-control knowledge / search heuristics from data,
but instead offloads the derivation of heuristic functions to the underlying off-the-shelf classical planner such as Fast Downward
by passing a learned PDDL model.
This framework enables the direct application of the large body of work on symbolically derived heuristics,
e.g., delete-relaxation heuristics including $\lmcut$ \cite{hoffmann01,Helmert2009} and abstraction heuristics including $\mands$ \cite{HelmertHHN14}.
It has an advantage over policy-learning approaches
because our system maintains all of the theoretical characteristics (optimality, completeness, etc.)
of the underlying off-the-shelf planner, with respect to the learned state space.
In contrast,
learned policies typically do not guarantee admissibility,
and only guarantees the convergence to the optimal policy in the limit.
Greedy algorithms typically used by RL agents during the evaluation also contribute to the lack of reliability.
Such a lack of the completeness/optimality guarantee is problematic for critical real-world applications.
Moreover, while systems that learn search control knowledge require
supervised signals, reward signals, or simulators that provides action symbols,
\latentplanner only requires a set of unlabeled image pairs (transitions), and
does not require a reward function, expert solution traces, simulators, or predetermined action symbols.

Model-based RL \cite{muzero,kaiser2020model} additionally learns a transition function in the state space to
facilitate efficient exploration.
World-model literature \cite{ha2018world} learns latent representations of states and their black-box transition functions,
i.e., a representation learning portion of model-based RL.
The learned transition function is typically a black-box function
that is not compatible with symbolic model analysis that are necessary for deriving heuristic functions.
AMA$_2$ (\refsec{sec:ama2}) and Vanilla Space AE (\refsec{sec:vanilla-space-ae}) can be seen as an instance of such a black-box world model.
As another example,
Causal InfoGAN  \cite{kurutach2018learning},
which was published after the conference version of this paper \cite{Asai2018},
learns binary latent representations and transition functions similar to Latplan
but with Generative Adversarial Networks \cite[GANs]{goodfellow2014generative}
augmented with Mutual Information maximization \cite[InfoGAN]{chen2016infogan}.
A significant limitation of this approach is that their transition function lacks the concept of actions:
The successor generation relies on sampling from a transition function which is expected to be multi-modal.
Unlike symbolic transition models, this does not guarantee that all logically plausible successors are enumerated
by a finite number of samples from a successor function, making the search process potentially incomplete.

More recently, several papers proposed ideas that resembles Cube-Space AE,
although outside the context of STRIPS model learning.
Alchemy \cite{wang2021alchemy} is a benchmark environment and an analysis toolkit for meta-reinforcement learning,
where the environment contains multiple stones which change shapes, colors, and sizes etc.\ due to actions.
Environment observations are provided either in a form of 3D rendering or in a symbolic encoding.
Its underlying dynamics assumes that each action has a consistent effect that resembles STRIPS actions.
Bisimulation metrics \cite{zhang2021learning} proposed
to improve the quality of the latent space learned by model-based RL.
The metric also resembles the assumptions in STRIPS, although the additional loss function
as well as the entire loss for representation learning are not justified as a lower bound of the likelihood.

\subsection{Neural Networks that Directly Models the Problems}

There is a large body of work using NNs to directly solve combinatorial tasks
by modeling the problem as the network itself,
starting with the well-known TSP solver by Hopfield and Tank  \citeyear{hopfield1985neural}.
Neurosolver represented a search state as a node in NN
and solved Tower of Hanoi \cite{bieszczad2015neurosolver}. 
However, they assume a symbolic input that is converted to the nodes in a network.
The neural network is merely used as a vehicle that carries out optimization.

\subsection{Novelty-Based Planning without Action Description}

While there are recent efforts in handling a complex state space without
having its action description 
\cite{frances2017purely}, action models could be used for other purposes
such as Goal Recognition \cite{ramirez2009plan},
macro-actions \cite{BoteaB2015,ChrpaVM15},
or plan optimization \cite{chrpa2015exploiting}.
Moreover, goal-directed heuristics based on descriptive action models are complementary
to novelty-based search techniques, and has their combination has been shown 
to achieve \lsota results on IPC domains \cite{lipovetzky2017bwfs}.

\section{Discussion and Conclusion}

\label{sec:discussion}

We proposed  \latentplanner, an integrated architecture for learning and planning which,
given only a set of unlabeled image pairs and no prior knowledge, generates a classical planning problem,
solves it with a symbolic planner,
and presents the plan as a human-comprehensible sequence of images.
We empirically demonstrated its feasibility using image-based versions of planning/state-space-search problems
(Blocksworld, 8-puzzle, 15-puzzle, Lights Out, Sokoban),
provided a theoretical justification for the training objectives from the maximum-likelihood standpoint,
and analyzed the model complexity of STRIPS action models from the perspective of graph coloring and Cube-Like Graphs.

Our technical contributions are
(1) \emph{State Auto-Encoder, which leverages the Binary-Concrete technique to learn a bidirectional mapping between raw images and propositional symbols compatible with symbolic planners}.
For example, on the 8-puzzle, the SAE can robustly compress the ``gist'' of a training image into a propositional vector representing the essential information (puzzle configuration) presented in the images.
(2) \emph{Non-standard prior distribution $\bern(\epsilon=0.1)$ for Binary Concrete,
which improves the stability of propositional symbols and helps symbolic search algorithms operate on latent representations.}
The non-standard prior encodes a closed-world assumption, which assumes that propositions are False by default,
unlike the standard prior $\bern(\epsilon=0.5)$ which assigns random, stochastic bits
to unused dimensions in the latent space,
i.e., ``unknown,'' much like in open-world assumption.
(3) \emph{Action Auto-Encoder, which grounds action symbols, i.e., identifies which transitions are ``same'' with regard to the state changes.}
By having a finite set of action symbols, agents can enumerate successor states efficiently during planning,
instead of exhaustively enumerating the entire state space or sampling successor states.
(4) \emph{Back-to-Logit, which enables learning and extracting descriptive, STRIPS-compatible action effects using a neural network.}
It restricts the hypothesis space of the action model with its structural prior,
and bounds the complexity of the action model through graph coloring on cube-like graphs.
(5) \emph{Complete State Regression Semantics, in which preconditions can be modeled as effects backward in time.}
-- preconditions are now modeled with prevail-conditions.
The network enables efficient and accurate extraction of preconditions from the trained network weights,
which resulted in better planning performance.
(6) \emph{Formalization of the training process under a sound likelihood maximization / variational inference framework.}
This results in a training objective that is less susceptible to overfitting due to various regularization terms
and the lower bounding characteristics.

The only key assumption we make about the input domain is that
it is fully observable and deterministic, i.e., that it is in fact a classical plannning domain.
We have shown that different domains can all be solved by the same system
without modifying any code or the NN architecture.
In other words, \emph{\latentplanner is a domain-independent, image-based classical planner}.
To our knowledge, this is the first system that completely automatically constructs a logical representation
\emph{directly usable by a symbolic planner} from a set of unlabeled image pairs for a diverse set of problems.

We demonstrated the feasibility of leveraging deep learning in order to enable
symbolic planning using classical search algorithms such as \astar,
when only image pairs representing action start/end states are available,
and there is no simulator, no expert solution traces, and no reward function.
Although much work is required to determine the applicability and scalability of this approach,
we believe this is an important first step in bridging the gap between symbolic and subsymbolic reasoning and
opens many avenues for future research.
In the following, we discuss several key limitations that Latplan must address in the future.

Latplan requires uniform sampling from the environment, which is nontrivial in many scenarios.
Automatic data collection via exploration and active learning \cite{burr2012active} is a major component of future work.

Next, in this paper, we evaluate \latentplanner as a high-level planner using puzzle domains such as the 8-puzzle.
Mapping a high-level action to low-level actuation sequences via a motion planner is beyond the scope of this paper.
Physically ``executing'' the plan is not necessary, as finding the
solution to the puzzles is the objective, so a ``mental image'' of the
solution (i.e., the image sequence visualization) is sufficient.
However,
in domains where actions have effects in the world, it will be necessary
to consider how actions found by \latentplanner (transitions between
latent bit vector pairs) can be mapped to actuations.
One direction of future work could be learning a continuous-discrete hybrid representation
that is more suitable for physical dynamical systems such as robotic tasks.

Latplan was evaluated in a noisy but fully observable and deterministic environment.
Representing states and state transitions with probabilistic belief states for partial observations
and stochastic state transitions would enable probabilistic planners to reason in such environments.

Latplan is also currently limited to tasks where a single goal state is specified.
Developing a method for specifying a set of goal states with a partial goal specification as in IPC domains is an interesting topic for future work.
For example, one may want to tell the planner ``the goal states must have tiles 0,1,2 in the correct places'' in a MNIST 8-puzzle instance.

Although we have shown that the Latplan architecture can be successfully applied to image-based versions of several relatively large standard puzzle domains (15-puzzle, Sokoban), some seemingly simple image-based domains may pose challenges for the current implementation of Latplan.
For example, in \refsec{sec:planning-hanoi} we discuss the result of applying Latplan to 4-disk, 4-tower Towers of Hanoi domain where the success ratio is quite low despite the fact that the domain has only 256 valid states, and thus is much ``simpler'' than domains such as the 15-puzzle.
We listed several potential reasons for failure in this domain, one of which is that some features are ``rare'', e.g., the smallest disk rarely appears in the top region of the image, resulting in an imbalanced dataset.
Such features are hard for the current SAE implementations to learn and generalize correctly, leading to incorrect plans.
Thus, methods for training a SAE on similar domains where images containing key features are very rare is a direction for future work.

Latplan's state representation is entirely propositional
and lacks first-order logic concepts such as predicates and parameters.
Historically, action model acquisition literature focused on such representations,
which provides generalization to environments with different numbers of objects.
As a result, adding an object to a Blocksworld environment currently requires training the model from the scratch.
One promising direction is using object-based representation, such as an ad-hoc approach pursued in \citep{harman2020learning},
in a more principled probabilistic manner.

Also, although we demonstrated that the SAE is somewhat robust to variations/noise,
it is not able to, for example, solve an instance (initial state image) of a  sliding-tile puzzle instance scrawled on a napkin by an arbitrary person.
Latplan will fail if, for example, some of the numbers in the initial state image for the 8-puzzle were rotated or translated, or
the appearance of the digits differed significantly from the those in the training data.
An ideal system would be robust enough to solve an instance (initial state image) of a  sliding-tile puzzle instance scrawled on a napkin by an arbitrary person.
Achieving this level of robustness will require improving the state encoder to the point that it is robust to styles, rotation, translation, etc.

Another direction for modifying the SAE is to combine the techniques in the autoencoders for the other types of inputs:
e.g., unstructured text \cite{li2015hierarchical} and audio data \cite{deng2010binary}.
Applying Gumbel-Softmax to these techniques, it may be possible for Latplan to perform language-based or voice-based reasoning.

In addition to the technical contributions,
this paper provides several conceptual contributions.
First and foremost, \emph{we provide the first demonstration that it is
possible to leverage deep learning quite effectively for classical planning,
which ``has been central to AI research since its inception.''} \cite[p396]{russell1995artificial}
We bridged the gap between connectionist and symbolic approaches by using the former as a perception system
generating the symbols for the latter resulting in a \emph{neuro-symbolic system}.

Second,
based on our observations of the problematic behavior of unstable propositions,
we defined the general \emph{Symbol Stability Problem} (SSP), a subproblem of symbol grounding.
We identified two sources of stochasticity which can introduce the instability:
(1) the inherent stochasticity of the network, and
(2) the external stochasticity from the observations.
This suggests that
SSP is an important problem that applies to any modern NN-based symbol grounding process
that operates on the noisy real-world inputs and
performs a sampling-based, stochastic process (e.g. VAEs, GANs) that are gaining popularity in the literature.
Thus, characterizing the aspect of SSP would help the process of designing a planning system operating on real-world input.
We demonstrated the importance of addressing the instability
by a thorough empirical analysis of the impact of instability on the planning performance.
An interesting avenue for future work is to extend our approach to InfoGAN-based
discrete representation of the environment \cite{kurutach2018learning}.

Finally, we demonstrated that  \lsota \domind search heuristics
provide effective search guidance in the automatically learned state spaces.
These \domind functions, which have been a central focus of the planning community in the last two decades,
provide search guidance without learning. This is in contrast to
popular reinforcement learning approaches that suffer from poor sample efficiency,
domain-dependence, and the lack of formal guarantees on admissibility.
We believe this finding stimulates further research into heuristic search
as well as reinforcement learning.

\clearpage
\renewcommand{\theHsection}{A\arabic{section}}
\appendix
\section{Probability Theory}
\label{sec:info}

A probability distribution $P(\rx)$ of a random variable $\rx$ defined on a certain domain $X$
is a function from a value $x\in X$ to some non-negative real $P(\rx=x)$.
The sum / integral over $X$ (when $X$ is discrete / continuous) is equal to 1,
i.e., $\sum_{x\in X} P(\rx=x) = 1$ (discrete domain) or $\int_{x\in X} P(\rx=x) dx = 1$ (continuous domain), respectively.
The function is also called a probability mass function (PMF) or a probability density function (PDF),
respectively.
Typically, each random variable is given a certain meaning,
thus two notations $P(\rx)$ and $P(\ry)$ denote different PMFs/PDFs
and the letter $P$ does not designate a function by itself,
unlike normal mathematical functions where $f(x)$ and $f(y)$ are equivalent under variable substitution.
For example, if $\rx$ is a boolean variable for getting a cancer and
$\ry$ is a boolean variable for smoking cigarettes,
then $P(\rx)$ and $P(\ry)$ denote completely different PMFs, and
we could write $P(\rx)=f(\rx)$, $P(\ry)=g(\ry)$, and $f\not=g$ to make it explicit.
When a value $x\in X$ is given, we obtain an actual value $P(\rx=x)=f(x)$,
which is sometimes abbreviated as $P(x)$.
To denote two different distributions for the same random variable,
an alternative letter such as $Q(\rx)$ is used.

A joint probability $P(\rx=x,\ry=y)$ is a function of two arguments which returns a probability of observing $x,y$ at once.
Conditional probability $P(\rx=x|\ry=y)$ represents a probability of observing $x$
when $y$ was already observed, and is defined as $P(\rx|\ry) = \frac{P(\rx,\ry)}{P(\ry)}$.
Therefore $P(\rx=x) = \sum_{y\in Y} P(\rx=x | \ry=y) P(\ry=y)$ holds.

For two probability distributions $Q(\rx)$ and $P(\rx)$ for a random variable $\rx$,
a \emph{Kullback-Leibler (KL) divergence} $\KL(Q(\rx)||P(\rx))$ is an expectation of their log ratio over $Q(\rx=x\in X)$:
\[
 \KL(Q(\rx)||P(\rx))=\E_{Q(\rx=x)}\brackets{\log \frac{Q(\rx=x)}{P(\rx=x)}}.
\]
This is $\sum_{x\in X} Q(x)\log \frac{Q(x)}{P(x)}$ for discrete distributions
and $\int_{x\in X} Q(x)\log \frac{Q(x)}{P(x)}dx$ for continuous distributions.
KL divergence is always non-negative, and equals to 0 when $P=Q$.
Conceptually it resembles a distance between distributions,
but it is not a distance because it does not satisfy the triangular inequality.

\section{Why do we Minimize L2-Norm / Square Errors ? Why not the \emph{Mean} Square Errors?}
\label{sec:l2-norm}

While ad hoc approaches have been successful in many machine learning systems,
a more principled approach to implementing machine learning algorithms is facilitated by a probabilistic interpretation \cite{murphy2012machine}.

The networks $f,g$ of an AE are optimized by minimizing a \emph{reconstruction loss} $||\vx-\hat{\vx}||$ under some norm,
which is typically a (Mean) Square Error / L2-norm.
Which norm to use is determined by the distribution of $\vx$ assumed by the model designer.
Assuming a 1-dimensional case,
let $x$ be a data point in the dataset, $z$ be a certain latent value,
and a probability distribution $p(x|z)$ be
what the neural network (and the model designer) believes is the distribution of $x$ given $z$.
Typical AEs for images assume that
$x$ follows a Gaussian distribution centered around the predicted value $\hat{x}=g(z)$, i.e.,
$p(x|z)=\mathcal{N}(x|\hat{x},\sigma) = \frac{1}{\sqrt{2\pi\sigma^2}} e^{-\frac{(x-\hat{x})^2}{2\sigma^2}}$ for an arbitrary fixed constant $\sigma$.
This leads to an analytical form of the negative log likelihood (NLL) $-\log p(x|z)$:
\begin{align}
 -\log \mathcal{N}(x|\hat{x},\sigma)
 = -\log \Big[\frac{1}{\sqrt{2\pi\sigma^2}} e^{-\frac{(x-\hat{x})^2}{2\sigma^2}}\Big]
 = \frac{(x-\hat{x})^2}{2\sigma^2} +\log \sqrt{2\pi\sigma^2} = C_1 (x-\hat{x})^2 + C_2 \label{eq:rec-loss}
\end{align}
for some constant $C_1>0$ and $C_2$, which is a scaled/shifted square error / L2-norm reconstruction loss.

For a multi-dimensional case,
by assuming that individual outputs (e.g., pixels) $\vx_i$, $\vx_j$ are independent for $i\not=j$,
we sum up \refeq{eq:rec-loss} across output dimensions $i$,
because $-\log p(\vx|\vz)=-\log \prod_i p(\vx_i|\vz) = \sum_i -\log p(\vx_i|\vz)$.
Finally, a \emph{mean} square loss ($C_1=1/D$, where $D$ is the number of output dimensions, e.g., pixels)
is obtained by arbitrarily setting $\sigma=\sqrt{D/2}$.

By minimizing the reconstruction loss $-\log p(\vx|\vz)$,
the training maximizes $p(\vx|\vz)$, the likelihood of observing $\vx$ --- the higher the probability,
the more likely we get an output $\hat{\vx}$ closer to the real data $\vx$.
The general framework that casts a machine learning task as a likelihood maximization task is
called \emph{maximum likelihood estimation}.

By assuming a different distribution on $\vx$, we obtain a different loss function.
For example, a Laplace distribution $\frac{1}{2b}\exp (-\frac{|\vx-\hat{\vx}|}{b})$ results in an absolute error loss $|\vx-\hat{\vx}|$.
Further, in many cases, a metric $d(x,\hat{x})$ automatically maps to a probability distribution
through $-\log p(x|\hat{x}) \propto d(x,\hat{x}) \Leftrightarrow p(x|\hat{x}) = C_1\exp (-C_2d(x,\hat{x}))$,
where $C_1,C_2$ are constants for maintaining $\int_x p(x|\hat{x})=1$.
While the choice of the loss function (thus the distribution of $\vx$) is arbitrary,
we typically choose Gaussian distribution because it is the \emph{maximum entropy distribution} among
continuous distributions on $\R$ with the same variance and the mean,
i.e.,
the Gaussian distribution has the highest entropy $\int -p(x)\log p(x)dx$, thus is most random,
therefore assumes the least about the distribution.

In many AE/VAE implementations, $C_2$ is ignored, and $C_1 = 1/D$ by assuming $\sigma=\sqrt{D/2}$.
Since this is an arbitrary value, it is sometimes necessary to tune $\sigma$ manually.
Alternatively, Bayesian NN methods \cite{nix1994estimating} learn to predict both $\sigma$ and the mean $\hat{x}$ of the Gaussian distribution
by doubling the size of the output of the network.
It trains the network by optimizing \refeq{eq:rec-loss} without omitting (now non-constant) $C_2$.

\section{Other Discrete Variational Methods}
\label{sec:other-discrete}

Other discrete VAE methods include VQVAE \citep{van2017neural},
DVAE++ \citep{vahdat2018dvae++}, and DVAE\# \citep{vahdat2018dvae}.
The difference between them is the training stability and accuracy.
They may contribute to stable performance, but
we leave the task of faster / easier training for future work.

A variant called Straight-Through Gumbel-Softmax (ST-GS) \cite{jang2017categorical}
combines a so-called Straight-Through estimator with Gumbel Softmax.
ST-GS is outperformed by standard GS \cite[Figure 3(b)]{jang2017categorical}, thus we do not use it in this paper.
However, we explain it in an attempt to cover as many methods as possible,
and also because this ST-estimator frequently appears in other discrete representation learning literature,
including ST-estimator for Heaviside $\function{step}$ function \cite{koul2018learning,bengio2013estimating}
and VQVAE \cite{van2017neural}.

A Straight-Through estimator is implemented by a primitive operation called \emph{stop gradient} \function{sg},
which is available in major deep learning / automatic differentiation frameworks.
\function{sg} acts as an identity in the forward computation but acts as a zero during the weight update / backpropagation
step of automatic differentiation in neural network training.
To understand how it works, it is best to see how ST-GS is implemented:
\begin{align}
 \STGS(\vl)=\function{sg}(\argmax(\vl)-\GS(\vl))+\GS(\vl) =
\left\{
 \begin{array}{ll}
 \argmax(\vl) & \text{(forward)}\\
 0+\GS  (\vl) & \text{(backward)}
 \end{array}
\right.
\end{align}
Notice that this function acts exactly as $\argmax(\vl)$ in the forward computation,
but uses only the differentiable $\GS(\vl)$ for backpropagation,
thus eliminating the need for a gradient of a non-differentiable function $\argmax$.
Several applications of ST-estimator exist.
ST estimator that combines Heaviside step function $\function{step}$ and a linear function $x$
was used in \cite{koul2018learning,bengio2013estimating}
but was outperformed by Binary Concrete \cite[Figure 3(a)]{jang2017categorical}.
Similarly, an ST-estimator can combine $\function{step}$ and Binary Concrete,
but this was also outperformed by the standard Binary Concrete \cite[Figure 3(a)]{jang2017categorical}.

\section{ELBO Computation of Gumbel Softmax / Binary Concrete VAE}
\label{sec:gsvae-elbo}

Having provided the overview,
we explain the details of ELBO computation of Gumbel Softmax VAE and then Binary Concrete VAE,
following \cite{jang2017categorical,MaddisonMT17}.
Those who are not interested in these theoretical aspects can safely skip this section,
but it becomes relevant in later sections where we analyze our contributions.
This section also discusses several ad-hoc variations of the implementations shared by the original authors (and their issues)
to sort out the information available in the field.

To compute ELBO, we need an analytical form of the reconstruction loss and the KL divergence.
We focus on the latter because
the choice of reconstruction loss $\E_{q(\vz\mid\vx)} [ \log p(\vx\mid\vz) ]$
is independent from the computation of the KL divergence $\KL(q(\vz\mid\vx)||p(\vz))$.

We denote a categorical distribution of $C$ classes as $\cat(\vp)$ with parameters $\vp\in \B^C$.
Here, $\vp$ is a probability vector for $C$ classes, thus it sums up to 1, i.e., $\sum_{k=1}^C \vp_k = 1$.
For example, when $\vp_k=1/6$ for all $k$ and $C=6$, it models a fair cube dice.

Gumbel-Softmax is a continuous relaxation of Gumbel-Max technique \citep{gumbel1954statistical,maddison2014sampling},
a method for drawing samples of categorical distribution $\cat(\vp)$ from a \emph{log-unnormalized probability} or a \emph{logit} $\vl$.
``Log-unnormalized'' imply that $\vl=\log \vp'$, where $\vp'$ is an unnormalized probability.
Since it is not normalized, $\vp'$ does not sum up to 1, but normalization is trivial ($\vp=\vp'_k/\sum_k\vp'_k$).
Combining these facts derives a following relation:
\begin{align}
 \vp_k = \vp'_k/\sum_k\vp'_k = \exp\vl_k/\sum_k\exp\vl_k = \softmax(\vl)_k.
\end{align}
Log-unnormalized probabilities $\vl$ are convenient for neural networks
because it can take an arbitrary value in $\R^C$.
Gumbel-Max draws samples from $\cat(\vp)$ using $\vl$ without computing $\vp$ explicitly:
\begin{align}
 \braces{0,1}^C\ni\function{GumbelMax}(\vl)=\argmax(\vl + \function{Gumbel}^C(0,1)) \sim \cat(\vp).
\end{align}
Again, note that we assume $\argmax$ returns a one-hot representation rather than the index of the maximum value.

Unlike samples generated by Gumbel-Max technique,
samples from Gumbel-Softmax function follows its own $\gsdist(\vl,\tau)$ distribution,
not the original $\cat(\vp)$ distribution:
\begin{align}
 \B^C\ni \vz=\GS(\vl)=\softmax\parens{\frac{\vl+\function{Gumbel}^C(0,1)}{\tau}} \sim \gsdist(\vl,\tau).
\end{align}
An obscure closed-form probability density function (PDF) of this distribution is available \cite{jang2017categorical,MaddisonMT17} as follows:
\begin{align}
 \gsdist(\vz\mid\vl,\tau) =\ (C-1)! \tau^{C-1} \prod_{k=1}^{C} \frac{\exp\vl_k\vz_k^{-(\tau+1)}}{\sum_{i=1}^C \exp\vl_i\vz_i^{-\tau}}.
\end{align}
The factorial $(C-1)!$ is sometimes denoted by a Gamma function $\Gamma(C)$ depending on the literature.

\subsection{A Simple Add-Hoc Implementation}
\label{sec:add-hoc-gsvae}

In practice,
an actual VAE implementation may avoid using this complicated PDF of $\gsdist(\vl,\tau)$
by computing the KL divergence based on $\cat(\vp)$ instead.
While it could potentially violate the true lower bound (ELBO), in practice, it does not seem to cause a significant problem.
This issue is discussed by \citet[Eq.21,22]{MaddisonMT17}.
The following derivation is based on the actual implementation shared by an author on his website%
\footnote{\url{https://blog.evjang.com/2016/11/tutorial-categorical-variational.html}},
which corresponds to Eq.22 in \citet{MaddisonMT17}.
It made two modifications to the faithful formulation based on the complicated PDF of $\gsdist(\vl,\tau)$
in order to simplify the optimization objective.
In this implementation,
it computes the KL divergence as if the annealing is completed ($\tau=0$),
treating the variable $\vz$ as a discrete random variable.
The implementation also uses $p(\vz)=\cat(\1/C)$
(i.e., a uniform categorical distribution) as a prior distribution.
The KL divergence in the VAE is thus as follows:
\begin{align}
 \int q(\vz\mid\vx) \log \frac{q(\vz\mid\vx)}{p(\vz)} d\vz
 &\approx \sum_{k\in\braces{1..C}} q(\rvz_k=1\mid\vx) \log \frac{q(\rvz_k=1\mid\vx)}{p(\rvz_k=1)}  \quad \because \text{$\vz$ is treated as discrete.} \nn
 &= \sum_{k\in\braces{1..C}} q(\rvz_k=1\mid\vx) \log \frac{q(\rvz_k=1\mid\vx)}{\frac{1}{C}}                                                     \nn
 &= \sum_{k\in\braces{1..C}} q(\rvz_k=1\mid\vx) \log q(\rvz_k=1\mid\vx) + \sum_{k\in\braces{1..C}} q(\rvz_k=1\mid\vx)(\log C)                                     \nn
 &= \sum_{k\in\braces{1..C}} q(\rvz_k=1\mid\vx) \log q(\rvz_k=1\mid\vx) + \log C \quad \because \text{$q$ sums up to 1.}
\end{align}
Since it assumes that the annealing is completed,
the distribution is also treated as if it is equivalent to $\cat(\vq)$
where $\vq=\softmax(\vl)$.
Therefore, this formula can be computed using $q(\rvz_k=1\mid\vx)=\vq_k=\softmax(\vl)_k$.

\section{Difference of the Non-Standard Bernoulli Prior (\refsec{sec:bcvae-prior}) from the Zero-Suppress SAE \cite{Asai2019a}}
\label{sec:zsae-conference-version-difference}

Our conference paper \citep{Asai2019a} proposed ZSAE (Zero-suppress SAE),
a method which uses an additional ad-hoc loss term called Zero-suppress loss defined as follows:
\begin{align}
  -H(q) + \alpha\cdot \BC(l) \label{eq:zsae-icaps19}
\end{align}
where $l$ is an output of the encoder network, which is a logit of $q$ (i.e., $q=\sigmoid(l)$).
Compare this with the KL divergence we defined in \refeq{eq:zsae} (repost):
\begin{align}
q\log\frac{q}{\epsilon}+(1-q)\log\frac{(1-q)}{(1-\epsilon)} &=-H(q) + \alpha \cdot q - \log (1-\epsilon) \label{eq:zsae-appendix}
\end{align}
Two losses are similar up to the constant difference $-\log (1-\epsilon)$ because
$\BC$ is a modified form of $\sigmoid$ (\refeq{eq:bc}),
therefore $\BC(l)$ in \refeq{eq:zsae-icaps19} and $q$ in \refeq{eq:zsae} behave similarly.

Although Zero-suppress loss was shown to work,
it is an ad-hoc loss that is not theoretically justified as an ELBO.
Moreover, ZSAE used the values \emph{after} the Binary Concrete activation $\BC(l)$:
This introduces unnecessary noise in the loss due to the logistic noise in $\BC$,
which makes the training slower and less stable than the theoretically justified formulation presented in this paper.

\section{Loss derivation for AMA\texorpdfstring{$_3^+$}{3+}}
\label{sec:loss-derivation}
\renewcommand{\defaultindex}{}
\renewcommand{\defaultcomma}{}

We define the statistical model of this network as a probability distribution $p(\rxboth)$
of observed random variables $\rxbefore$ and $\rxafter$, also known as a \emph{generative model} of $\rxbefore$ and $\rxafter$. 
Specifically, we model the probability distribution $p(\rxboth)$ by introducing latent variables corresponding to an action label $\raction$ (one-hot vector) and the current and successor propositional states, $\rzbefore$ and $\rzafter$ (binary vectors).
The model is trained by maximizing the log-likelihood $\log p(\rxboth)$ observing a tuple $(\rxboth)$.
Since it is difficult to compute the log-likelihood function of such a latent variable model,
we resort to a variational method to approximately maximize the likelihood, which leads to the autoencoder architecture depicted in \refig{fig:ama3}.

We start to model $p(\rxboth)$ by assuming a set of dependencies between variables
and a set of distributional assumptions for the variables. This process is often called \emph{statistical modeling}.
\emph{Following this process often helps to define a sound probabilistic model.}

\refeqs{eq:ama3-generative0}{eq:ama3-generative1} below define dependencies between variables,
where all summations are over the respective domains.
In \refeq{eq:ama3-generative1}, we assumed $\rxbefore$ and $\rxafter$ depend only on $\rzbefore$ and $\rzafter$, respectively,
because $\rxbefore$ and $\rxafter$ are visualizations of $\rzbefore$ and $\rzafter$, respectively.

\begin{align}
p(\rxboth)
 & = \sum_{\rzboth,\raction} p(\rxboth,\rzboth,\raction)                           \nn
 & = \sum_{\rzboth,\raction} p(\rxboth \mid \rzboth,\raction) p(\rzboth,\raction). \label{eq:ama3-generative0}\\
p(\rzboth,\raction)              & = p(\rzafter\mid\rzbefore,\raction)p(\raction\mid \rzbefore) p(\rzbefore). \label{eq:ama3-generative3}\\
p(\rxboth \mid \rzboth,\raction) & = p(\rxbefore \mid \rzbefore) p(\rxafter \mid \rzafter).                 \label{eq:ama3-generative1}
\end{align}

The dependencies also define the I/O interfaces of the subnetworks (e.g., $p(\rxbefore \mid \rzbefore)$ is modeled by a network that takes $\zbefore$ and returns $\xbeforerec$),
and which ones are prior distributions that take no inputs (e.g., $p(\rzbefore)$).
In general, when defining a generative model, it is better to avoid making unnecessary/unjustified independence assumptions.
For example, although it is possible, we generally avoid assuming $p(\rzboth,\raction)= p(\rzbefore) p(\rzafter) p(\raction)$,
i.e., $\rzboth$ and $\raction$ to be all independent,
unless there is a domain knowledge that justifies the independence.

\begin{figure}
 \centering
 \includegraphics{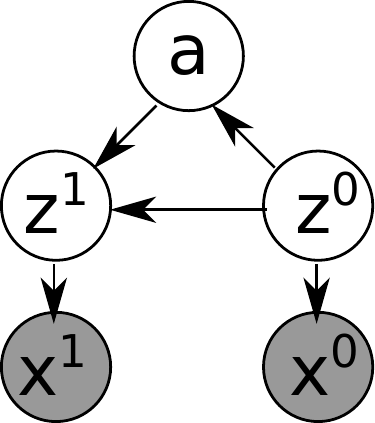}
 \caption{DAG notation (graphical model) of our statistical model
defined in \refeqs{eq:ama3-generative0}{eq:ama3-generative1}.
The dark nodes are for observed variables, and the light nodes are for latent variables.}
 \label{fig:graphical-model-vanilla}
\end{figure}

This process is also called \emph{graphical modeling} \cite[Chapter 14]{russell1995artificial}.
It visualizes the statistical model using
a directed acyclic graph (DAG) notation whose nodes are variables and edges are conditional dependencies.
For example, our model (\refeqs{eq:ama3-generative0}{eq:ama3-generative1})
can be visualized as \refig{fig:graphical-model-vanilla}.

After defining the dependencies, we assign an assumption to each distribution, as shown below.
Thanks to the process of defining our generative model,
we now realize that $p(\raction\mid \rzbefore)$ requires an additional subnetwork $\aaep(\zbefore[])$
that is missing in both the AMA$_2$ and the AMA$_3$ models.
This network models $p(\raction\mid \rzbefore)$ by taking a latent vector $\vz$ and returns a probability vector over actions.
Note that some distributional assumptions apply to multiple vectors in \refig{fig:ama3}.
For example, \refeq{eq:ama3-generative5} applies to both $\vz=\zafter$ and $\vz=\zafteralt$.
\begin{align}
  & p(\rzbefore)  = \bern(\epsilon). & & \text{\refsec{sec:bcvae-prior}.} \nn
  & p(\raction | \rzbefore=\zbefore[]) = \cat(\aaep(\zbefore[]))& & \label{eq:ama3-generative3.5}\\
  & p(\rxbefore| \rzbefore=\zbefore[]) = \mathcal{N}(\xbeforerec[],\sigma), \text{where}\ \xbeforerec[]    =\decode(\zbefore[]).         \label{eq:ama3-generative4}    \\
  & p(\rxafter | \rzafter =\zafter[]) = \mathcal{N}(\xafterrec[],\sigma), \text{where}\ \xafterrec[]       =\decode(\zafter[]).          \label{eq:ama3-generative5}    \\
  & p(\rzafter | \rzbefore=\zbefore[], \raction=\action[]) = \bern(\qafteralt[]), \text{where}\ \qafteralt[]   =\sigmoid(\lafteralt[]), \text{$\lafteralt[]=\aaed(\zbefore[],\action)$.} \label{eq:ama3-generative7}
\end{align}

Next, to compute and maximize $p(\rxboth)$, we should compute the integral/summation over latent variables $\rzboth,\raction$ (\refeq{eq:ama3-generative0}).
Such an integration is known to be intractable; 
in fact, the complexity class of computing the likelihood of observations (e.g., $p(\rxboth)$)
is shown to be \#P-hard \cite{dagum1993approximating,roth1996hardness,dagum1997optimal}
through reduction to \#3SAT.
Therefore, we have to resort to maximizing $p(\rxboth)$ approximately.

Variational modeling techniques, including VAEs, tackle this complexity by approximating the likelihood from below.
This approximated objective is called ELBO (Evidence Lower BOund)
and is defined by introducing a \emph{variational model} $q$,
an arbitrary distribution that a model designer can choose.
To derive an autoencoding architecture, we model the variational distribution with encoders.
\refeqs{eq:ama3-variational2}{eq:ama3-variational4} define dependencies and assumptions in $q$,
e.g., $\rzafter$ depends only on $\rxafter$.
\begin{align}
  & q(\rzbefore | \rxbefore=\xbefore[]) = \bern(\qbefore[]), \text{where}\ \qbefore[] = \sigmoid(\lbefore[]), \lbefore[]=\encode(\xbefore[]). \label{eq:ama3-variational2} \\
  & q(\rzafter  | \rxafter =\xafter []) = \bern(\qafter[]), \text{where}\ \qafter[]   = \sigmoid(\lafter[]),  \lafter[] =\encode(\xafter[]). \label{eq:ama3-variational3} \\
  & q(\raction  | \rxbefore=\vx,\rxafter =\vx')   = \cat (\qaction[]), \text{where}\ \qaction[]  = \softmax(\laction[]), \laction[]=\aaee(\vx,\vx'). \label{eq:ama3-variational4}
\end{align}

Using this model, we derive a lower bound by introducing variables one by one.
The first variable to introduce is $\rzbefore$ and
the bound is derived using the same proof used for obtaining VAE's ELBO (\refeq{eq:elbo}).
\begin{align}
\log p(\rxboth)
 & = \log \parens{\sum_{\rzbefore} \gray{p(\rxboth \mid \rzbefore)} \blue{p(\rzbefore)}} \nn
 & = \log \parens{\sum_{\rzbefore} \gray{p(\rxboth \mid \rzbefore)} \blue{\frac{p(\rzbefore)}{q(\rzbefore\mid\rxbefore)}} q(\rzbefore\mid\rxbefore)} \nn
 & = \log \parens{\E_{q(\rzbefore\mid\rxbefore)} \brackets{ \gray{p(\rxboth \mid \rzbefore)} \blue{\frac{p(\rzbefore)}{q(\rzbefore\mid\rxbefore)}}}} \quad \text{(Definition of expectations)} \nn
 & \geq \E_{q(\rzbefore\mid\rxbefore)} \brackets{\log \parens{ \gray{p(\rxboth \mid \rzbefore)} \blue{\frac{p(\rzbefore)}{q(\rzbefore\mid\rxbefore)}}}} \quad \text{(Jensen's inequality)}\nn
 & = \E_{q(\rzbefore\mid\rxbefore)}  \brackets{ \log \gray{p(\rxboth \mid \rzbefore)}} \nn
 & \quad - \KL(\blue{q(\rzbefore\mid\rxbefore) \Mid p(\rzbefore)}). \quad \text{(Definition of KL divergence)}
\label{eq:vanilla-elbo1}
\end{align}

Next, we decompose $\log \gray{p(\rxboth \mid \rzbefore)}$ by introducing $\raction$ and deriving the lower bound.
We are merely reapplying the same proof with a different set of variables and distributions:
\begin{align}
\log \gray{p(\rxboth \mid \rzbefore)}
 & \geq \E_{q(\raction\mid\rxboth)} \brackets{ \log \gray{p(\rxboth \mid \rza)}} - \KL(\red{q(\raction\mid\rxboth) \Mid p(\raction \mid \rzbefore)}). \label{eq:vanilla-elbo2}
\end{align}

Since $\rxbefore$ does not depend on $\rxafter$ and $\raction$ (\refeq{eq:ama3-generative1}),
\begin{align}
 \log \gray{p(\rxboth \mid \rza)}
 &= \log \violet{p(\rxbefore \mid \rzbefore)} \gray{p(\rxafter \mid \rza)}\nn
 & = \log \violet{p(\rxbefore \mid \rzbefore)} + \log \gray{p(\rxafter \mid \rza)}.
\label{eq:vanilla-elbo3}
\end{align}

We further decompose $\log \gray{p(\rxafter \mid \rza)}$
by deriving its variational lower bound:
\begin{align}
 \log \gray{p(\rxafter \mid \rza)}
 &= \log \sum_{\rzafter} p(\rxafter \mid \rzbefore,\rzafter,\raction) \cyan{\frac{p(\rzafter\mid\rzbefore,\raction)}{q(\rzafter\mid\rxafter)}} q(\rzafter\mid\rxafter)\nn
 &\geq \E_{q(\rzafter \mid \rxafter)} \log \black{p(\rxafter \mid \rzbefore,\rzafter,\raction)} - \KL(\cyan{q(\rzafter\mid\rxafter) \Mid p(\rzafter\mid\rzbefore,\raction)})\nn
 &=   \E_{q(\rzafter \mid \rxafter)} \log \orange{p(\rxafter \mid\rzafter)} - \KL(\cyan{q(\rzafter\mid\rxafter) \Mid p(\rzafter\mid\rzbefore,\raction)}).
\label{eq:vanilla-elbo5}
\end{align}

However, we can also decompose it by applying Jensen's inequality directly.
The reconstruction loss obtained from this formula corresponds to the reconstruction from the latent vector $\zafteralt$ (\refig{fig:ama3})
generated/sampled by the AAE, as the value is an expectation over $p(\rzafter\mid\rzbefore,\raction)$.
\begin{align}
\log \gray{p(\rxafter \mid \rzbefore,\raction)}
 &= \log \sum_{\rzafter} p(\rxafter \mid \rzbefore,\rzafter,\raction) p(\rzafter \mid \rzbefore,\raction)\nn
&\geq \E_{p(\rzafter \mid \rzbefore, \raction)} \log p(\rxafter \mid \rzbefore,\rzafter,\raction)\nn
&=   \E_{p(\rzafter \mid \rzbefore, \raction)} \log \teal{p(\rxafter \mid \rzafter)} \quad\text{\refeq{eq:ama3-generative1}}.
\label{eq:vanilla-elbo6}
\end{align}

We wish to optimize both of these objectives (\refeqs{eq:vanilla-elbo5}{eq:vanilla-elbo6}).
To understand the motivation, it is crucial to see the interpretation of these objectives.
First, $\KL(\cyan{q(\rzafter\mid\rxafter) \Mid p(\rzafter\mid\rzbefore,\raction)})$ maintains the symbol stability we discussed in \refsec{sec:unstable}
by making two distributions ($q(\rzafter\mid\rxafter)$ and $p(\rzafter\mid\rzbefore,\raction)$) identical.
Without this KL term, latent vectors encoded by the encoder (i.e., $q(\rzafter\mid\rxafter)$)
and by the AAE (i.e., $p(\rzafter\mid\rzbefore,\raction)$) may diverge,
despite them being the symbolic representation for the same raw observation $\rxafter$.
Second, $\E_{p(\rzafter \mid \rzbefore, \raction)} \log \teal{p(\rxafter \mid \rzafter)}$ ensures
that the latent vectors generated by the action model are reconstructable.
If we optimize \refeq{eq:vanilla-elbo5} alone, the stability is maintained, but the resulting vector $\rzafter$ is not guaranteed to be reconstructable.
If we optimize \refeq{eq:vanilla-elbo6} alone, the stability is not maintained.
Therefore, it is crucial to optimize both objectives at once.

To optimize both objectives at once, we combine them into a single objective.
There is some flexibility in how to achieve this:
For example, one could consider alternating two loss functions in each epoch,
or consider optimizing the maximum of two objectives.
Optimizing their Pareto front may also be an option.
However, since they both give a lower bound of $\log \gray{p(\rxafter \mid \rzbefore,\raction)}$,
a simple approach to optimizing them at once while maintaining the ELBO is to take a weighted sum between them, with equal weights 0.5.

Now that the log likelihood is decomposed into either reconstruction losses from a decoder or KL divergences,
we can finally discuss the total optimization objective.
Similar to $\beta$-VAE discussed in \refsec{sec:aevae}, we can apply coefficients $\beta_1,\beta_2,\beta_3 \geq 1$ to each of the KL terms.
This does not overestimate the ELBO because KL divergence is always positive,
and larger $\beta$ results in a lower bound of the original ELBO.
Combining \refeqs{eq:vanilla-elbo1}{eq:vanilla-elbo6}, we obtain the following total maximization objective:

\begin{align}
\log p(\rxboth)
\geq & - \beta_1\KL(\blue{q(\rzbefore\mid\rxbefore) \Mid p(\rzbefore)})\nn
     &+\E_{q(\rzbefore\mid\rxbefore)}
 \left[
 \begin{array}{ll}
  - \beta_2\KL(\red{q(\raction\mid\rxboth) \Mid p(\raction\mid\rzbefore)})\nn
   + \E_{q(\raction\mid\rxboth)}
  \left[
   \begin{array}{ll}
    \log \violet{p(\rxbefore \mid \rzbefore)} \nn
     + \frac{1}{2}\E_{q(\rzafter\mid\rxafter)} \log \orange{p(\rxafter\mid\rzafter)}\nn
     - \frac{1}{2}\beta_3\KL(\cyan{q(\rzafter\mid\rxafter) \Mid p(\rzafter\mid\rzbefore,\raction)})\nn
     + \frac{1}{2}\E_{p(\rzafter\mid\rzbefore, \raction)} \log \teal{p(\rxafter\mid\rzafter)}
   \end{array}
  \right]\nn
 \end{array}
 \right]\nn
=
 & - \beta_1\KL(\blue{q(\rzbefore\mid\rxbefore) \Mid p(\rzbefore)})\nn
 & +\E_{q(\rzbefore\mid\rxbefore)} \left[ - \beta_2\KL(\red{q(\raction\mid\rxboth) \Mid p(\raction\mid\rzbefore)})\right]\nn
 & +\E_{q(\rzbefore\mid\rxbefore)} \log \violet{p(\rxbefore \mid \rzbefore)}\nn
 & +\frac{1}{2}\E_{q(\rzafter\mid\rxafter)} \log \orange{p(\rxafter\mid\rzafter)}\nn
 &+\frac{1}{2}\E_{q(\rzbefore\mid\rxbefore)q(\raction\mid\rxboth)}
 \left[
 - \beta_3\KL(\cyan{q(\rzafter\mid\rxafter) \Mid p(\rzafter\mid\rzbefore,\raction)})
 \right]\nn
 & +\frac{1}{2}\E_{q(\rzbefore\mid\rxbefore)q(\raction\mid\rxboth)p(\rzafter\mid\rzbefore, \raction)} \log \teal{p(\rxafter\mid\rzafter)}. \nn
 \label{eq:vanilla-elbo7}
\end{align}

\renewcommand{\defaultindex}{i}
\renewcommand{\defaultcomma}{,}

Compared to \refeq{eq:vanilla-elbo8-main} we showed in the main.text, this formula appears much more complex.
In particular, it contains a large number of complex expectations ($\E\ldots$).
We now explain what these expectations imply.

\renewcommand{\defaultindex}{}
\renewcommand{\defaultcomma}{}

Practical VAE implementations typically compute the expectations of quantities that depend on stochastic variables
(e.g., reconstruction loss $\E_{q(\rzafter\mid\rxafter)}\log p(\rxafter \mid \rzafter)$ in \refeq{eq:vanilla-elbo5},
which depends on stochastic $\rzafter$ sampled from $q(\rzafter\mid\rxafter)$)
by merely taking a \emph{single} sample of the variables rather than taking multiple samples and computing the mean of the quantities.
Recall that $\zafter=\BC(\lafter)$ is a noisy function,
thus computing the value of $\zafter$ from $\lafter$ is equivalent to sampling the value once.

The reason VAE implementations, including ours, follow this approach
is that it reduces the runtime, simplifies the code, and empirically works well.
If we want to compute an expectation $\E_{q(\rzafter\mid\rxafter)}\log p(\rxafter \mid \rzafter)$ from multiple samples instead,
we could do as follows:
First,
we compute $\lafter=\encode(\xafter)$, which is deterministic.
Next, we compute a noisy function $\BC(\lafter)$ for $K$ times, i.e.,
take $K$ samples $\zafter[1,1],\ldots,\zafter[1,K]$ from a single $\lafter$.
We then run the decoder $K$ times to obtain $K$ samples of $\xafterrec[1,k]=\decode(\zafter[1,k])$,
compute $K$ reconstruction losses $\log p(\xafter[1,k] \mid \zafter[1,k])$ (e.g., scaled squared error in \refeq{eq:rec-loss}),
then obtain the mean $\frac{1}{k}\sum_{k=1}^{K} \log p(\xafter[1,k] \mid \zafter[1,k])$.
Limiting $K=1$ avoids these complications.

As a result, practically, \refeq{eq:vanilla-elbo7} for each data sample $\xboth$ is implemented as follows.
Reordering the formula for readability,
\begin{align}
 & \log \violet{p(\xbefore \mid \zbefore)} + \frac{1}{2}\log \orange{p(\xafter\mid\zafter)} + \frac{1}{2}\log \teal{p(\xafter\mid\zafteralt)} &\text{(Reconstruction losses)}      \nn
 & - \beta_1\KL(\blue{q(\zbefore\mid\xbefore) \Mid p(\zbefore)})                                                                              &\text{(Prior for $\zbefore$)}       \nn
 & - \beta_2\KL(\red{q(\action\mid\xboth) \Mid p(\action\mid\zbefore)})                                                                       &\text{(Prior for $\action$)}        \nn
 & - \frac{1}{2}\beta_3\KL(\cyan{q(\rzafter                                                                                                   =\zafter\mid\xafter) \Mid p(\rzafter =\zafteralt\mid\zbefore,\action)}). &
 \label{eq:vanilla-elbo8}
\end{align}

Reconstruction losses are square errors due to Gaussian assumptions.
For KL terms, for example,
$\KL(\cyan{q(\zafter\mid\xafter) \Mid p(\zafteralt\mid\zbefore,\action)})$ is obtained by
converting the logits $\lafter,\lafteralt \in \R$ to
probabilities $\qafter=\sigmoid(\lafter), \qafteralt=\sigmoid(\lafteralt)$,
then computing the KL divergence as follows.
\[
 \KL(\qafter\Mid\qafteralt)= \qafter\frac{\log \qafter}{\log \qafteralt}+(1-\qafter)\frac{\log (1-\qafter)}{\log (1-\qafteralt)}.
\]

\dnote{
\textbf{Comparison to AMA$_3$:}
AMA$_3$ contains a number of ad-hoc differences from AMA$_3^+$.
For example, AMA$_3$ uses an \emph{absolute error loss} between two latent vectors $\zafter$ and $\zafteralt$,
instead of using the KL divergence $\KL(\qafter\Mid\qafteralt)$.
This is not theoretically justified, and is inefficient because losses using the sampled values $\zafter$ introduce noise in the loss function.

}

\renewcommand{\defaultindex}{i}
\renewcommand{\defaultcomma}{,}

\clearpage
\section{Bayesian Interpretation of Back-to-Logit}
\label{sec:btl-bayesian-interpretation}

\begin{figure}[htb]
 \centering
 \includegraphics{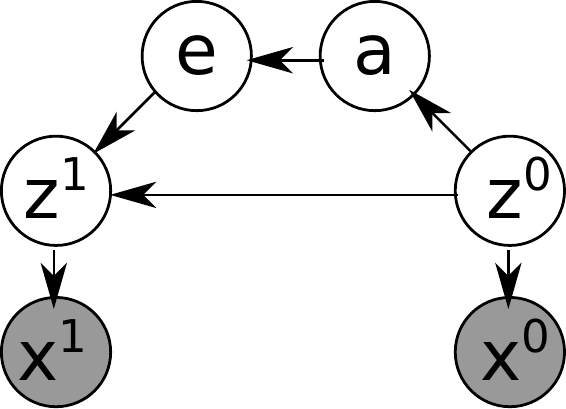}
 \caption{DAG notation (graphical model) of CSAE.}
 \label{fig:graphical-model-csae}
\end{figure}

\renewcommand{\defaultindex}{}
\renewcommand{\defaultcomma}{}
We can view BTL as a mechanism that injects a new assumption into $p(\rzafter|\rzbefore,\raction)$
by introducing a random variable $\rve$ and removing the dependency from $\rve$ to $\rzbefore$ as follows:
\begin{align}
 p(\rzafter|\rzbefore,\raction)
&=\sum_{\rve} p(\rzafter|\rzbefore,\raction,\rve) p(\rve|\rzbefore,\raction)\nn
&=\sum_{\rve} p(\rzafter|\rzbefore,\rve) p(\rve|\raction)
\end{align}
where $p(\rve|\raction)$ is modeled by $\effect$, and $p(\rzafter|\rzbefore,\rve)$ is modeled by $m$ and an addition.
This approach is supported by the Bayesian theoretical analysis of batch normalization \cite{teye2018bayesian}.
Obtaining a value of (i.e., sampling the value of) $\rve$ from $\raction$ with $p(\rve|\raction)$
is a stochastic process because it is affected by other data points in the same batch,
which are randomly sampled in each training iteration.
\renewcommand{\defaultindex}{i}
\renewcommand{\defaultcomma}{,}
This model can be visualized as a graphical model in \refig{fig:graphical-model-csae}.

\clearpage
\section{Proof of Monotonicity}
\label{sec:proof-monotonicity}

Proof of Theorem \refthm{thm:monotonicity}:

\begin{proof}
For readability, we omit $j$ and assume a 1-dimensional case.
Let $e=\effect(\action[])\in \R$,
which is a constant for a fixed action $\action[]$.
In the test time, Binary Concrete is replaced by a step function (\refsec{sec:argmax}). The BTL is then simplified to
\begin{align}
 \zafter=\function{step}(m(\zbefore)+e).
\end{align}

The possible values of a pair $(\zbefore,\zafter)$ is $(0,0),(0,1),(1,0),(1,1)$.
Since both \function{step} and $m=\BN$ are deterministic at the testing time \citep{ioffe2015batch},
we consider only the deterministic mapping from $\zbefore$ to $\zafter$.
There are only 4 deterministic mappings from $\braces{0,1}$ to $\braces{0,1}$:
 $\braces{(0,1), (1,1)}$,
 $\braces{(1,0), (0,0)}$,
 $\braces{(0,0), (1,1)}$,
 $\braces{(0,1), (1,0)}$.
Thus our goal is to show that the last mapping $\braces{(0,1), (1,0)}$ is impossible
in the latent space produced by an ideally trained BTL.

To prove this, first, assume that there is $(\zbefore,\zafter)=(0,1)$ for some index $i$. Then
\begin{align}
 1=\function{step}(m(0)+e) \Rightarrow m(0)+e>0 \Rightarrow m(1)+e>0 \Rightarrow \forall i; m(\zbefore)+e > 0.
\end{align}
The second step is due to the monotonicity $m(0)<m(1)$.
This shows $\zafter$ is constantly $1$ regardless of $\zbefore$,
therefore it proves that $(\zbefore,\zafter)=(1,0)$ cannot happen in any $i$.

Likewise, if $(\zbefore,\zafter)=(1,0)$ for some index $i$,
\begin{align}
 0=\function{step}(m(1)+e) \Rightarrow m(1)+e<0 \Rightarrow m(0)+e<0 \Rightarrow \forall i; m(\zbefore)+e < 0.
\end{align}
Therefore, $\zafter=0$ regardless of $\zbefore$,
and thus $(\zbefore,\zafter)=(0,1)$ cannot happen in any $i$.

Finally, if the data points do not contain $(0,1)$ or $(1,0)$, then by assumption they do not coexist.
Therefore, the embedding learned by BTL cannot contain $(0,1)$ and $(1,0)$ at the same time.
\end{proof}

\clearpage

\section{Training Curve}
\label{sec:curve}

\begin{figure}[htbp]
 \begin{tabular}{cc}
  Training loss & Validation loss \\
  \multicolumn{2}{c}{ELBO} \\
  \includegraphics[width=0.45\linewidth]{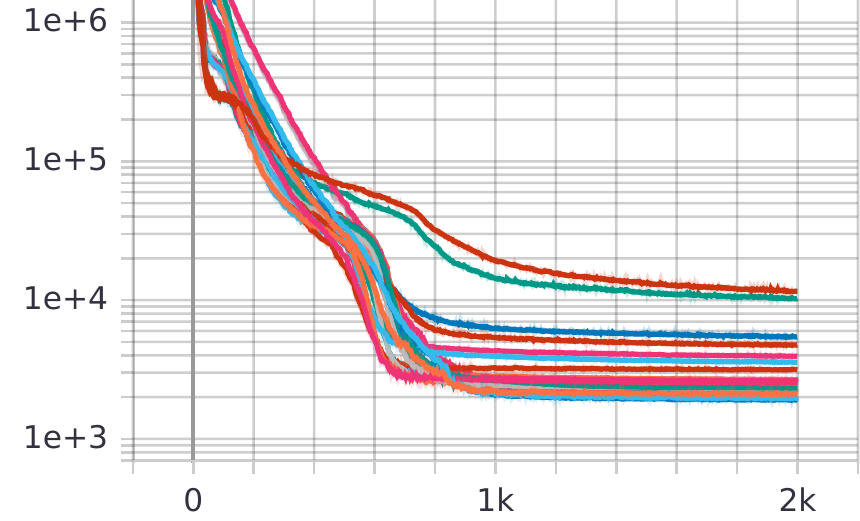}&    \includegraphics[width=0.45\linewidth]{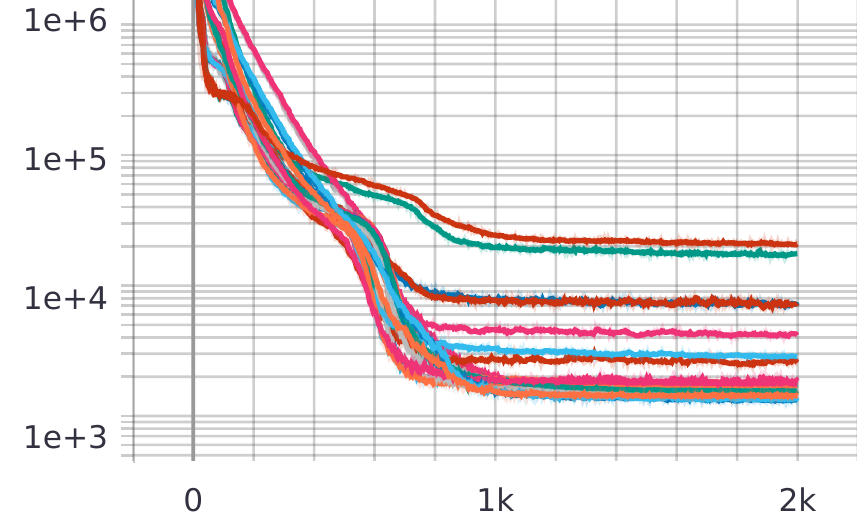}\\
  \multicolumn{2}{c}{$\E_i\brackets{\sum_f \KL(q(\zbefore_f|\xbefore)||p(\rzbefore)=\bern(\epsilon=0.1))}$}\\
  \includegraphics[width=0.45\linewidth]{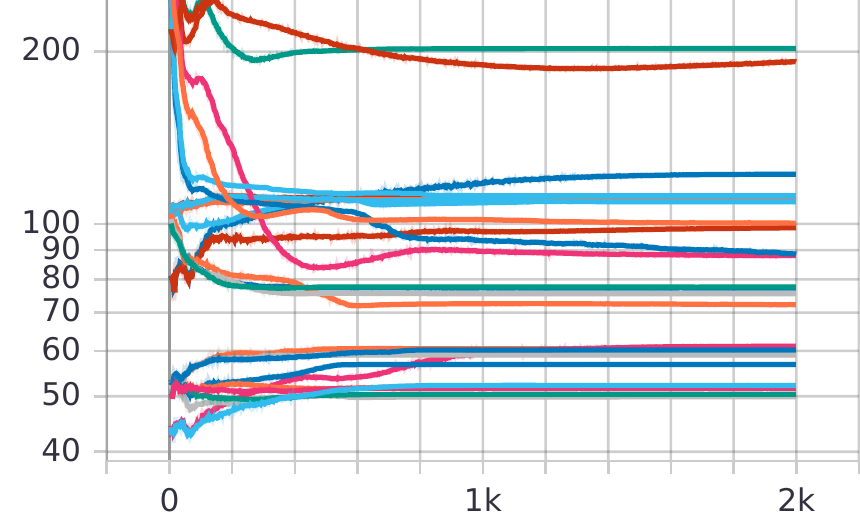}&   \includegraphics[width=0.45\linewidth]{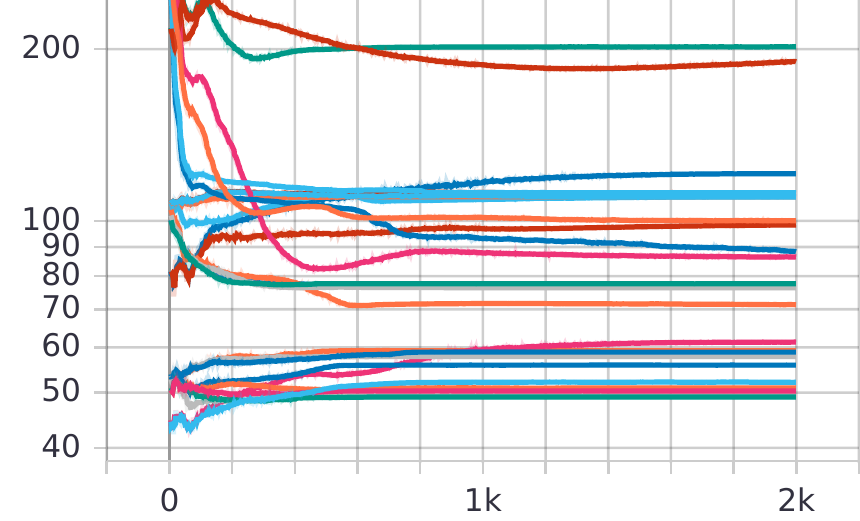}\\
  \multicolumn{2}{c}{$\E_i\brackets{\sum_f \KL(q(\zafter_f|\xafter)||p(\zafteralt_f|\zbefore,\action))}$} \\
  \includegraphics[width=0.45\linewidth]{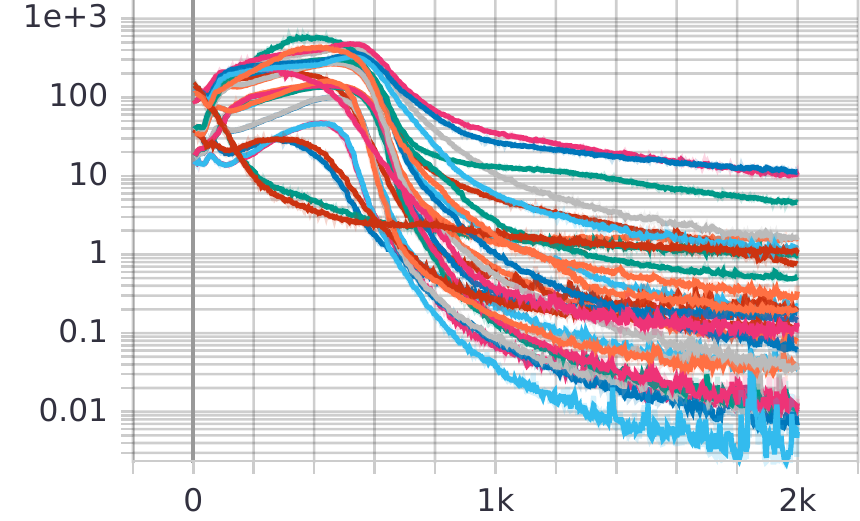}& \includegraphics[width=0.45\linewidth]{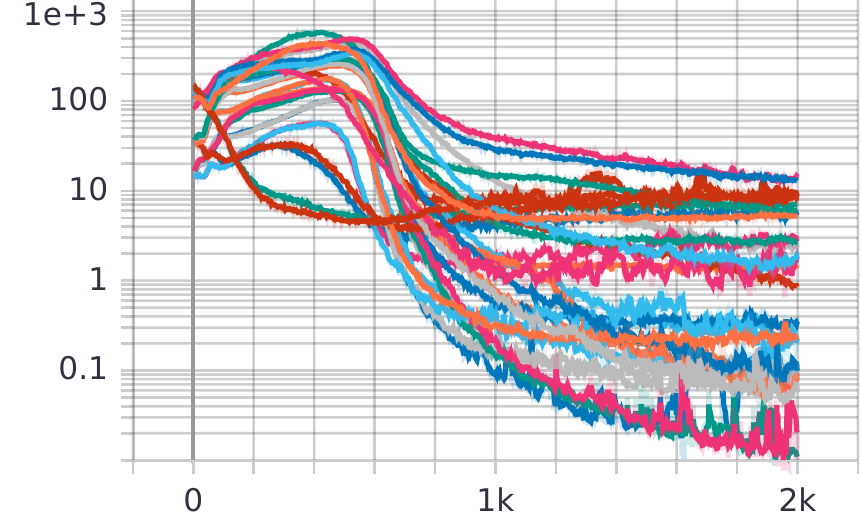}\\
  \multicolumn{2}{c}{$\tau$}\\
  \multicolumn{2}{c}{\includegraphics[width=0.4\linewidth]{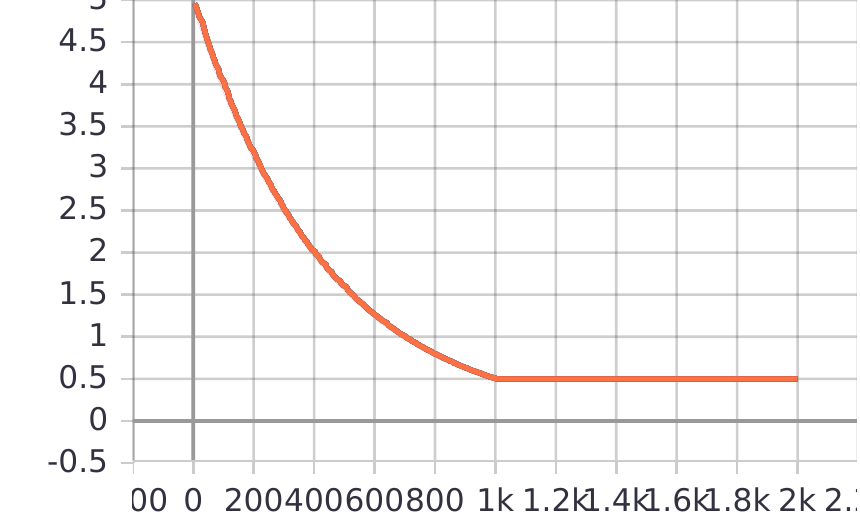}}
 \end{tabular}
 \caption{Training and validation curves.
The $x$-axes show the training epochs.
The $y$-axes are in a logarithmic scale except for $\tau$.
}
 \label{fig:curve}
\end{figure}

To facilitate reproducibility,
we show the training and validation loss curves and what to expect.
\refig{fig:curve} shows the results of training \ama4 networks on 8-Puzzle.
Similar behaviors were observed in other domains and \ama3 networks.
We see multiple curves in each subfigure due to multiple hyperparameters.
The figure includes three metrics as well as the annealing parameter $\tau$.
We could make several observations from these plots.

\textbf{ELBO}: In the successful training curves with lower final ELBO, we do not observe overfitting behavior.
ELBO curves show that annealing $\tau$ and training the network with higher $\tau$ is essential,
as the networks stop improving the ELBO after the annealing phase is finished at epoch 1000.
Indeed, we performed the same experiments with a lower initial value $\tau_{\text{min}}=1.0$,
and they exhibited significantly less accuracy.
This is because lower $\tau$ in $\BC$ makes the latent $\sigmoid$ function closer to a step function
(steeper around 0, flatter away from 0) and produce smaller gradients, as discussed by \citet{jang2017categorical}.

\textbf{KL divergence against the prior distribution}: The group with the lower initial,
\[\KL(q(\zbefore|\rxbefore)||p(\zbefore)),\]
consists of those with hyperparameter $F=50$.
This is because
the KL divergence is the sum across latent dimensions and the KL divergence in each dimension tends to have a similar value initially.

\textbf{KL divergence for successor prediction}: While the overall ELBO loss monotonically decreases,
the $\KL(\allowbreak q(\zafter|\xafter)\allowbreak ||\allowbreak p(\zafteralt|\zbefore,\action))$ loss representing successor prediction accuracy
initially goes up, then goes down.
This indicates that the network initially focuses on learning the reconstructions,
then moves on to adjust the state representation and the action model in the later stage of the training.
The KL divergence continues to improve after the annealing is finished at epoch 1000.
The curves seem to indicate that we could train the network even longer in order to achieve better successor prediction loss.

\clearpage
\section{Domain-wise colorization of heuristics figure}
\label{sec:heuristics-domainwise}

Cf. Figure \refig{fig:heuristics}.

\begin{center}
\begin{minipage}{0.8\linewidth}
 \centering
 \includegraphics[width=0.47\linewidth]{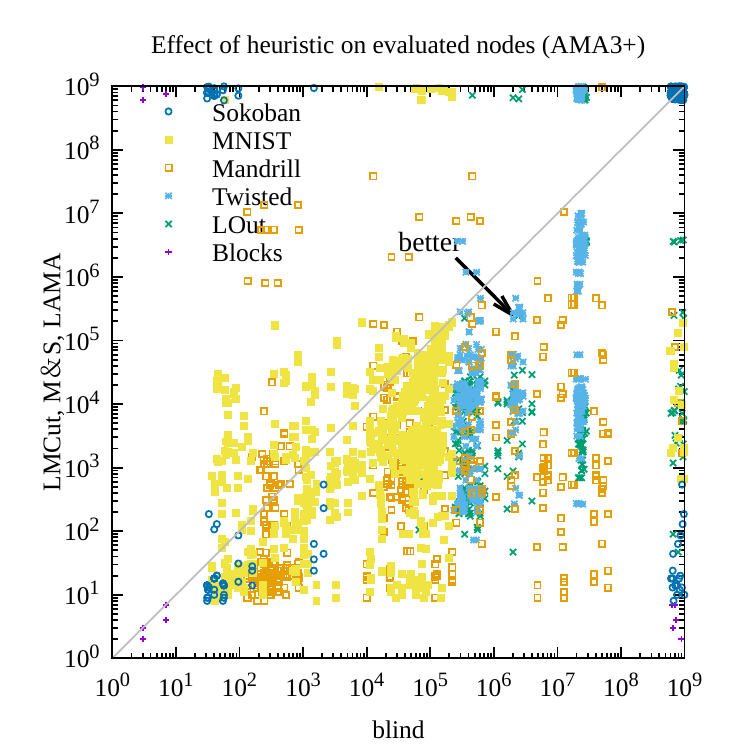}
 \includegraphics[width=0.47\linewidth]{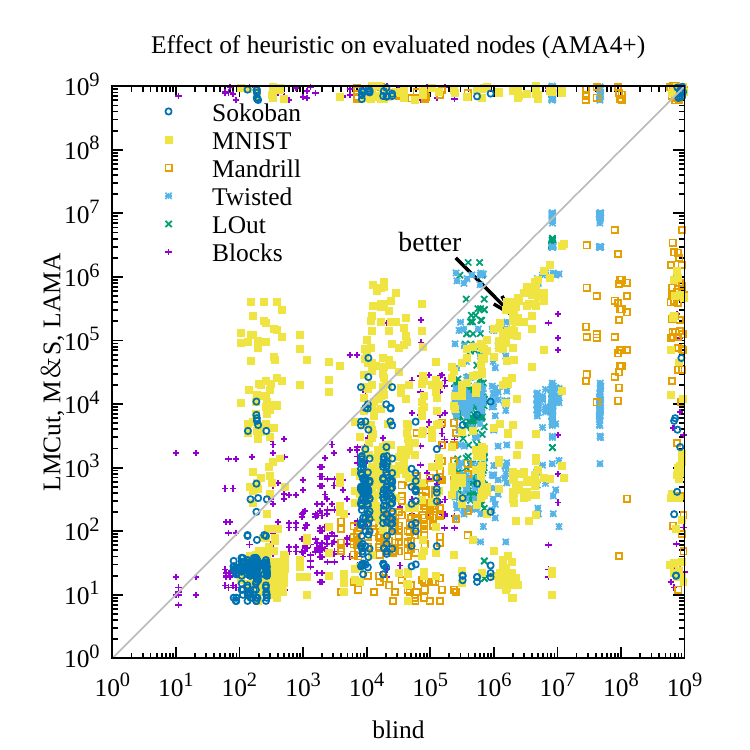}\\
 \includegraphics[width=0.47\linewidth]{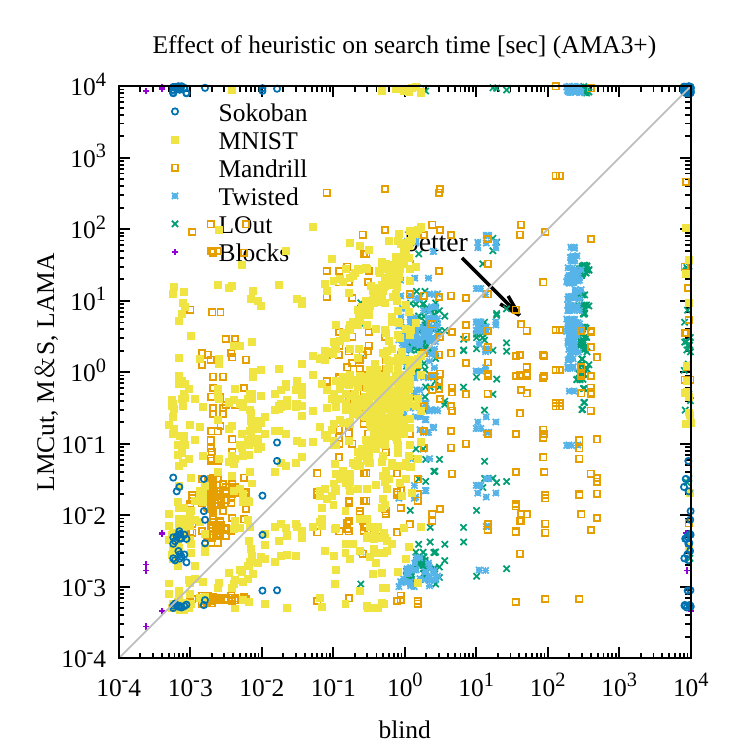}
 \includegraphics[width=0.47\linewidth]{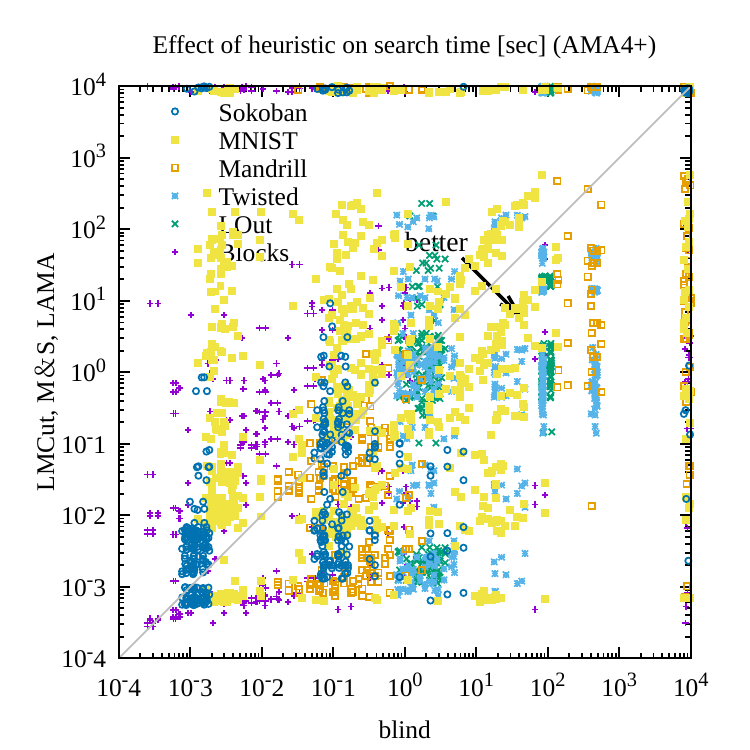}
\end{minipage}
\end{center}

\section{Domain-wise colorization of solvability experiments}
\label{sec:exhaust-domainwise}

\begin{figure}[p]
 \centering
 \includegraphics[width=0.47\linewidth]{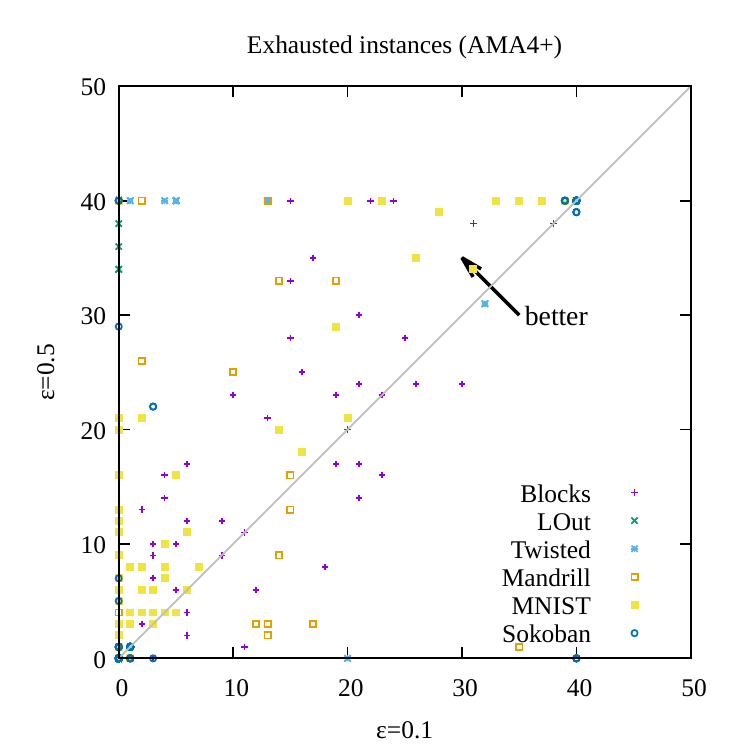}
 \includegraphics[width=0.47\linewidth]{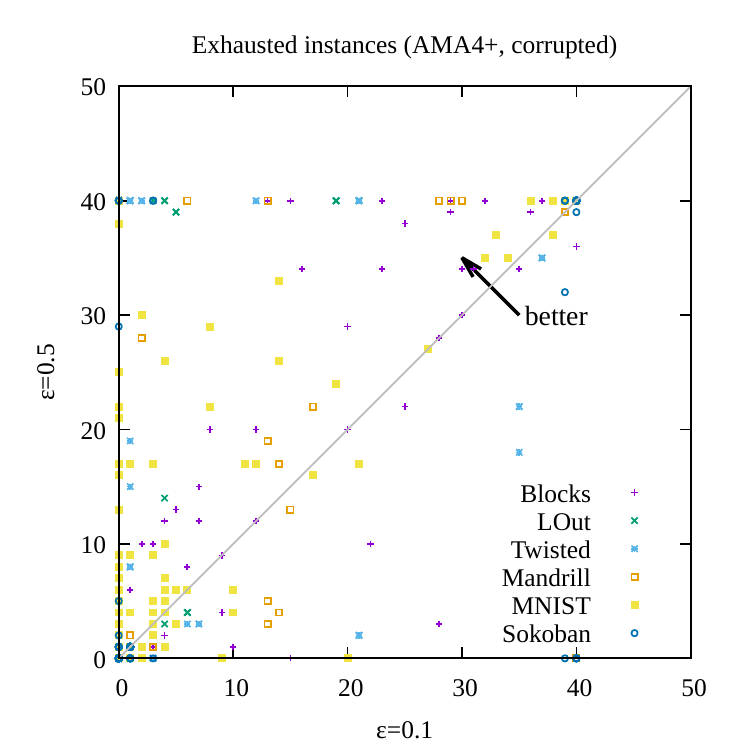}\\
 \caption{
 Domain-wise colorization of \refig{fig:exhaust-vs-nozsae}.
 }
 \label{fig:exhaust-vs-nozsae-domain}
\end{figure}

\clearpage
\section{Plan Validators}
\label{sec:validators}

We developed a plan validator for each visual planning domain
to evaluated the accuracy of the learned representation and
the performance of the planner.
This section describe the details of these validators.

All validators consist of two stages:
State-wise validation (\texttt{validate\_states})
and transition validation (\texttt{validate\_transitions}).
State-wise validation typically checks for inaccurate reconstructions
and violation of domain-specific constraints in the image.
Transition validation checks for the correctness of the transitions.

The source code of these validators are available in the source code of Latplan repository
\footnote{\url{https://github.com/guicho271828/latplan/}}.

\subsection{Sliding Tile Puzzle Validators (MNIST, Mandrill)}

Since we know the number of rectangular grids in each domain,
we first segment each image in the plan into tile patches.
For each patch, we compare the mean absolute errors against a fixed set of ground-truth tile patterns
which are used to generate the dataset.

After computing the distance, we do not simply pick the nearest tile ---
it would result in selecting the same ground-truth tile pattern for several patches.
Instead, the algorithm searches for the best threshold value $\theta$ for the absolute error by a binary search
under $0.0\leq\theta\leq 0.5$.
A patch is considered a ``match'' when the absolute error is below $\theta$.
The reason we search for the threshold is because MNIST and Mandrill images have different pixel value ranges.
Manually searching for the threshold is not only cumbersome but would also result in an arbitrary decision
(for example, we could set a very forgiving / strict value and claim that it works well / does not work well).
Another reason is that Sliding Tile Puzzle assumes that there are no duplicate tiles,
which can be used to detect invalid states.

The threshold $\theta$ is increased/decreased
so that it balances the number $n_1$ of ambiguous patches that matches more than one ground-truth tiles
and the number $n_2$ of patches that fail to match any ground truth.
$\theta$ is increased when $n_1<n_2$ and decreased when $n_1>n_2$.
The search is stopped when $|n_1-n_2|\leq 1$.
We consider an image is invalid when $n_1\not=0$ and $n_2\not=0$.

To validate a transition, we map an image into a compact state configuration representing each tile as an integer id
using the threshold value discovered in the previous step.
We then check if only two tiles are changed, if the tiles are adjacent, and if one of them is the empty tile (tile id 0).

\subsection{LightsOut and Twisted LightsOut Validator}

We reuse some of the code from Sliding Tile Puzzle validators
with two ground-truth tiles being the ``On'' tile (indicated by a + sign)
and the ``Off'' tile (a blank black tile).

Unlike MNIST and Mandrill, LightsOut images consist of complete black-and-white pixels,
and also multiple patches can match the same groud-truth tile (either ``on'' or ``off'').
We therefore set the threshold to 0.01,
a value small enough that any noticeable + marks will be detected as an ``On'' state of each button.
We map each state into a 0/1 configuration representing a puzzle state
and validate the transitions.

To handle Twisted LightsOut images,
we undo this swirl effect and apply the same validator.
The threshold is increased to 0.04 to account for the numerical errors (e.g., anti-aliasing effects)
caused by the swirl effect.

\subsection{Blocksworld Validator}

We use the ground-truth knowledge of the image generator that
the image consists of 4 blocks with four different colors (red, green, blue, black).

The first step of the validator parses the image into a compact representation.
To do so,
it quantifies the image into 3-bit colors,
then counts the number of pixels for each color to detect the objects.
The most frequent color (gray) is ignored as the background.
Also, colors which occur less than 0.1\% of the entire image are ignored as a noise.

For each color, we compute the centroid and the width/height to obtain the estimate of the object location.
While doing so, we ignore the outliers by
first computing the 25\% and 75\% quantiles, and ignoring the coordinates outside
1.5 times the width of the quantiles.
Assume $\vx$ is a vector of $x$-axis or $y$-axis coordinates for each pixel in the image with the target color of interest.
Formally, we perform the following operation for each coordinate axis and each color:

\begin{align*}
 Q_1 &\from \text{Quantile}(\vx, 1/4)\\
 Q_3 &\from \text{Quantile}(\vx, 3/4)\\
 w   &\from Q_3 - Q_1\\
 l &\from Q_1 - 1.5 \cdot w\\
 u &\from Q_3 + 1.5 \cdot w\\
 \text{Return:} &\quad \braces{x\in \vx \mid l<x<u}.
\end{align*}

After parsing the image into a set of objects, we check the validity of the state as follows.
For each object $o$, it first searches for objects below it using
the half-width $o.w$ (half the width of the bounding box, i.e., dimension from the center), the half-height $o.h$ and the centroid coordinates $o.x, o.y$.
An object $o_1$ is below another object $o_2$ when $o_1.y > o_2.y$
and in the same tower, i.e., $|o_1.x - o_2.x| < \frac{o_1.w + o_2.w}{2}$.
(Note that the pixel coordinates are measured with the top-left being the origin.)

If an object $o$ is a ``bottom'' object (nothing is directly below it),
it compares the bottom edge of the bounding box $o.y+o.h$ with other bottom objects
and check if they have the similar $y$-coordinates, ensuring that none of them are in the mid air.
The $y$-coordinates are considered ``similar''
when the differences are within half the average height of two objects, i.e., $|o_1.y-o_2.y| < \frac{o_1.h + o_2.h}{2}$.

Otherwise, there are other objects below the object $o_1$.
It collects all objects below it, and extracts the object $o_2$ with the smallest $y$-coordinate.
It checks if $o_1$ and $o_2$ are in direct contact
by checking if $y$-coordinate difference is around the sum of both heights,
i.e., $(o_1.h + o_2.h) \cdot 0.5 < |o_1.y - o_2.y| < (o_1.h + o_2.h) \cdot 1.5$.

To check the validity of the transitions,
it first looks for unaffected objects, and check if exactly one object is moved.
If this is satisfied, it further checks if the moved object is a ``top'' object both before and after the transition.

\subsection{Sokoban Validator}

The sokoban validator partially shares code with LightsOut, therefore it does not use the binary search to find the threshold.
Moreover, it does not use the threshold and it assigns each patch to the closest ground-truth panel
(wall, stone, player, goal, clear, stone-at-goal) available in PDDLGym.

To validate the state, we use the fact that the environment was generated from a single PDDL instance (p006-microban-sequential).
We enumerate the entire state space with Dijkstra search and store the ground-truth configurations of the states into an archive.
A state is valid when the same configuration is found in an archive.

To validate the transition,
we can't use the same approach used for states because the archive contains all states but not all transitions ---
Since the enumeration uses a Dijkstra search to enumerate states, there is only one transition for each state as a destination.
We reimplemented the precondition and the effects of each PDDL action (move, push-to-goal, push-to-nongoal)
for our representation.

\clearpage
\section{Additional Experiments for Towers of Hanoi Domain (Negative Results)}
\label{sec:planning-hanoi}

As a failure case of Latplan, we describe \textbf{Towers of Hanoi} (ToH) domain and the experiments we performed.
A $(d,t)$-ToH domain with $t$ towers and $d$ disks has $t^d$ states and $t^{d+1}-2$ transitions in total.
The optimal solution for moving a complete tower takes $2^d-1$ steps if $t=3$,
but we parameterized our image generator so that it can generate images with more than 3 towers.
Each input image has a dimension of $(H,W,C)=(hd, wt, 3)$ (height, width, color channel),
where each disk is presented as a $h$x$w$ pixel rectangle ($h=1,w=4$) with distinct colors.
The validator of ToH images is identical to that of the Sliding Tile Puzzles
except for the fixed patterns that are matched against each image patch.

We generated several datasets with increasing generator parameters, namely
$(d,t)=(4,4),\allowbreak (3,9),\allowbreak (4,9),\allowbreak (5,9)$.
The number of states in each configurations is 256, 729, 6561, 59049.
We generated 20 problem instances for each configuration,
where the number of optimal solution length is 5 for $(3,9)$ because it lacks 7-step optimal plans\footnote{The diameter of the state space graph is less than 7.}, and
is 7 for $(4,4)$, $(4,9)$, and $(5,9)$ because they lack 14-step optimal plans.
While several 3-disk instances were solved successfully,
no valid plans were produced with larger number of disks, as seen in \reftbl{tab:planning-hanoi}.
Visualizations of successful and failure cases are shown in \refigs{fig:planning-hanoi}{fig:planning-hanoi2}.

The apparent failure of Latplan in ToH is surprising
because its state space tends to be smaller and thus is ``simpler'' than the other domains we tested.
It could be because the size of the dataset is too small for deep-learning based approaches, or
that the images are horizontally long and the 5x5 convolutions are not adequate.
It could also be due to the lack of visual cues -- however, alternative image renderers did not change the results
(e.g., black/white tiles with varying disk size,
thicker disks, disks painted with hand-crafted patterns.)
Lastly, there may be imbalance in the visual features
because the smallest disk rarely appears in the top region of the image.
For $d$-disk, $t$-tower ToH domain,
the smallest disk can appear only $t$ times in the top row among $t^d$ states,
which is exponentially rare in a uniformly randomly sampled dataset.
As a result, pixel data in the top rows are biased toward being gray (the background color).
We suspect deep-learning may have an assumption that all of the factors (i.e., fluents that distinguish the states) are seen often enough, which is not satisfied in this dataset.

\begin{table}[htbp]
\centering
\begin{adjustbox}{width=0.95\linewidth,keepaspectratio}
\begin{tabular}{|r|*{4}{ccc|ccc|}}
\hline
 & \multicolumn{6}{c|}{Blind}   & \multicolumn{6}{c|}{LAMA}   & \multicolumn{6}{c|}{LMCut}   & \multicolumn{6}{c|}{M\&S}   \\ \hline
 & \multicolumn{3}{c|}{kltune} & \multicolumn{3}{c|}{default} 
 & \multicolumn{3}{c|}{kltune} & \multicolumn{3}{c|}{default} 
 & \multicolumn{3}{c|}{kltune} & \multicolumn{3}{c|}{default} 
 & \multicolumn{3}{c|}{kltune} & \multicolumn{3}{c|}{default}  \\ \hline
domain
& {f} & {v} & {o} & {f} & {v} & {o} 
& {f} & {v} & {o} & {f} & {v} & {o} 
& {f} & {v} & {o} & {f} & {v} & {o} 
& {f} & {v} & {o} & {f} & {v} & {o} \\ \hline
\multicolumn{25}{|c|}{\ama3}\\\hline
$(d,t)=(4,4)$ & 0 & 0 & 0 & 0 & 0 & 0 & 0 & 0 & 0 & 0 & 0 & 0 & 0 & 0 & 0 & 0 & 0 & 0 & 0 & 0 & 0 & 0 & 0 & 0 \\
$(d,t)=(3,9)$ & 3 & 2 & 0 & 0 & 0 & 0 & 0 & 0 & 0 & 0 & 0 & 0 & 3 & 2 & 0 & 0 & 0 & 0 & 0 & 0 & 0 & 0 & 0 & 0 \\
$(d,t)=(4,9)$ & 0 & 0 & 0 & 0 & 0 & 0 & 0 & 0 & 0 & 0 & 0 & 0 & 0 & 0 & 0 & 0 & 0 & 0 & 0 & 0 & 0 & 0 & 0 & 0 \\
$(d,t)=(5,9)$ & 0 & 0 & 0 & 0 & 0 & 0 & 0 & 0 & 0 & 0 & 0 & 0 & 0 & 0 & 0 & 0 & 0 & 0 & 0 & 0 & 0 & 0 & 0 & 0 \\ \hline
\multicolumn{25}{|c|}{\ama4}\\\hline
$(d,t)=(4,4)$ & 0 & 0 & 0 & 0 & 0 & 0 & 0 & 0 & 0 & 0 & 0 & 0 & 0 & 0 & 0 & 0 & 0 & 0 & 0 & 0 & 0 & 0 & 0 & 0 \\
$(d,t)=(3,9)$ & 20 & 12 & 8 & 13 & 5 & 0 & 17 & 1 & 0 & 0 & 0 & 0 & 20 & 12 & 7 & 13 & 5 & 0 & 20 & 12 & 8 & 13 & 5 & 0 \\
$(d,t)=(4,9)$ & 0 & 0 & 0 & 0 & 0 & 0 & 0 & 0 & 0 & 0 & 0 & 0 & 0 & 0 & 0 & 0 & 0 & 0 & 0 & 0 & 0 & 0 & 0 & 0 \\ 
$(d,t)=(5,9)$ & 0 & 0 & 0 & 0 & 0 & 0 & 0 & 0 & 0 & 0 & 0 & 0 & 0 & 0 & 0 & 0 & 0 & 0 & 0 & 0 & 0 & 0 & 0 & 0 \\ \hline
\end{tabular}
\end{adjustbox}
\caption{The number of solutions found (f), valid solutions (v), and optimal solutions (o) for $(d,t)$-Towers of Hanoi domains using \ama3 and \ama4.}
\label{tab:planning-hanoi}
\end{table}

\clearpage
\section{Examples of Planning Results}
\label{sec:results-gallery}

\begin{figure}[htb]
 \centering
 \includegraphics[width=0.7\linewidth]{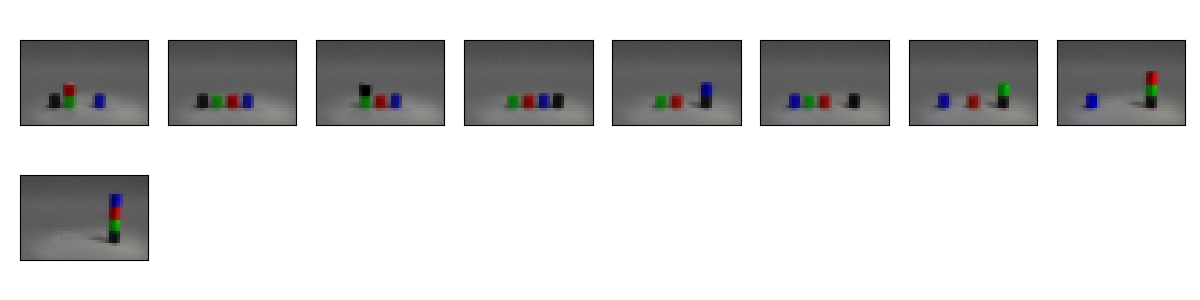}\\[1em]
 \includegraphics[width=0.7\linewidth]{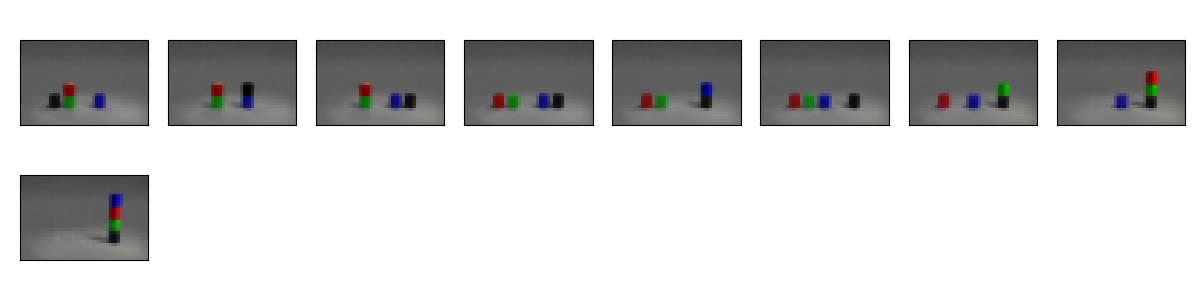}\\[1em]
 \includegraphics[width=0.7\linewidth]{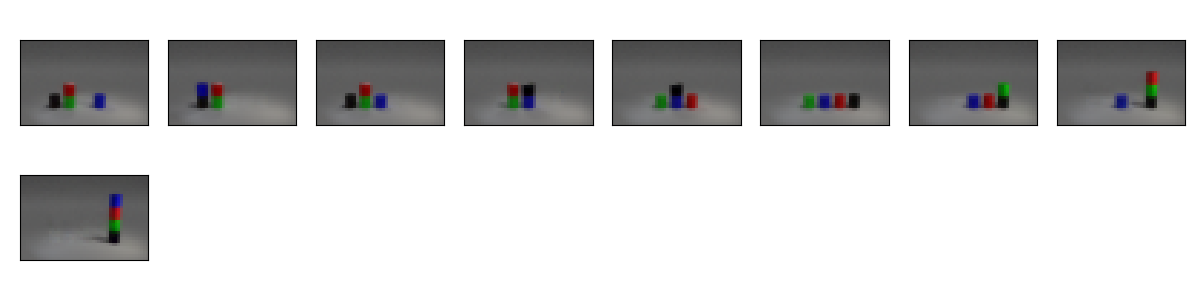}
 \caption{
 Examples of valid suboptimal plans in Blocksworld obtained from \ama4 with default priors ($\epsilon=0.5$).
 The first and the second plans were generated by the same network, but with \lmcut and \mands heuristics.
 The third plan was generated by \mands with another training result with different hyperparameters.
 They all resulted in different plans because different heuristics and state encoding result in a different node expansion order.
 See also: \refsec{sec:vs-nozsae} discusses why \astar + admissible heuristics can generate suboptimal visual plans.
 }
 \label{fig:planning-examples1}
\end{figure}

Finally, \refigs{fig:planning-examples1}{fig:planning-hanoi2} show
several examples of successful plans (optimal and suboptimal) as well as invalid plans
in order to illustrate how Latplan could behave or fail.

We did not see any qualitative differences between the visualized results
obtained by various configurations, conditioned by the same validity status of the plan (invalid/suboptimal/optimal).
For example,
we did not find any particular differences between the sets of invalid results
regardless of the choice of \ama3, \ama4, $\epsilon=0.1$ (kltune), $\epsilon=0.5$ (default), or other hyperparameter differences.
All invalid results look similarly invalid (albeit individual variations),
except the statistical differences observed in the overall number of invalid plans as discussed in \refsec{sec:planning-evaluation}.
Similarly,
suboptimal results look similarly suboptimal and no configuration particularly / visually / qualitatively stands out.
Optimal results also looks similarly optimal.
One exception is the case of using LAMA, in which case plans tend to be much longer than in the other configurations.

Due to this and the obvious space reasons,
we do not attempt to exhaustively enumerate the results of all configurations.
We instead hand-picked several examples without much consideration on the configuration.
\textbf{Thus, although we describe each configuration in the description,
it is not our intention to compare these results and make any claims based on them.}

\begin{figure}[htb]
 \centering
 \includegraphics[width=0.7\linewidth]{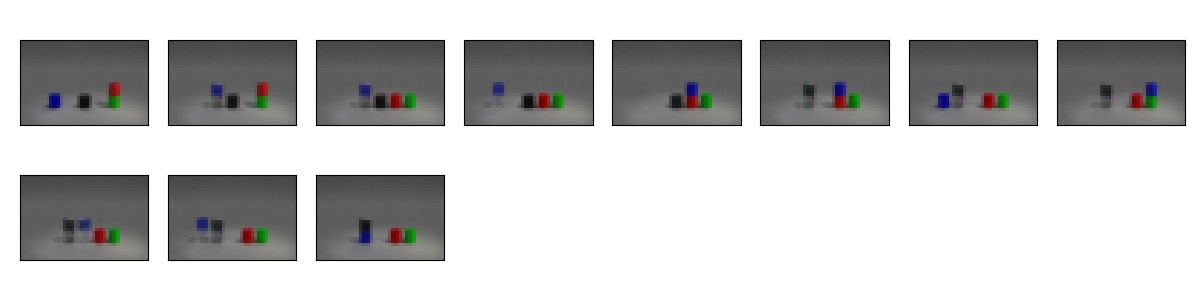}\\[1em]
 \includegraphics[width=0.7\linewidth]{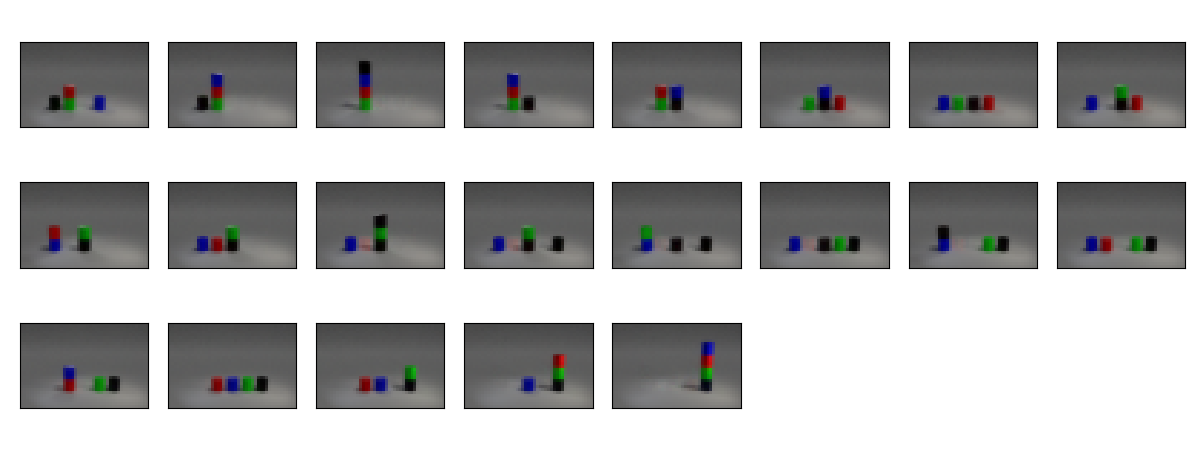}
 \caption{
 Examples of invalid plans in Blocksworld for a 7-steps problem and a 14-steps problem.
 In the first instance, floating objects can be observed.
 In the second instance, the colors of some blocks change and some objects are duplicated.
 (The first instance was generated by \ama4, default prior, LAMA.
 The second instance was produced by \ama4, $\epsilon=0.1$ prior, \mands.)
 }
\end{figure}

\begin{figure}[htb]
 \centering
\includegraphics[width=0.7\linewidth]{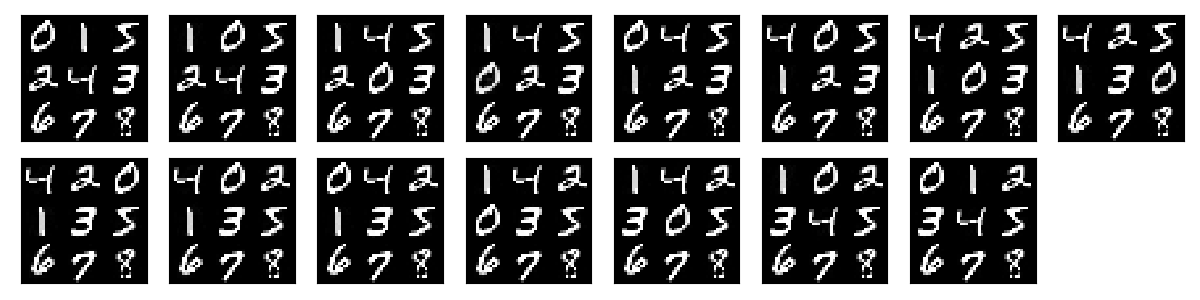}\\[1em]
\includegraphics[width=0.7\linewidth]{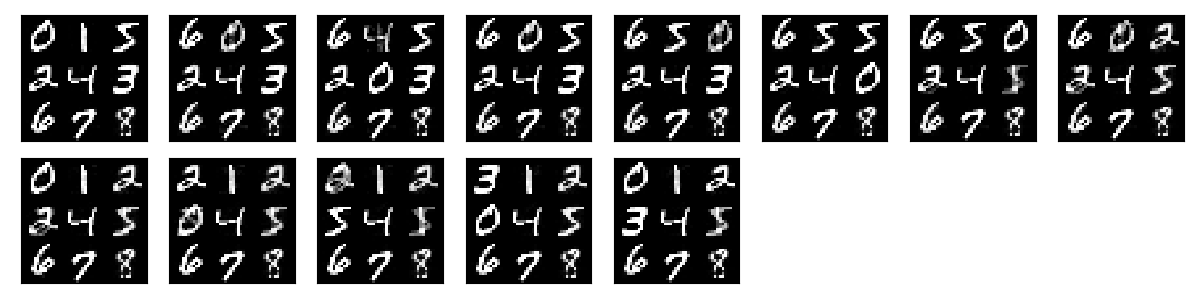}
 \caption{
 Examples from MNIST 8-puzzle.
 The first plan is optimal (14 steps).
 The second plan is invalid due to the duplicated ``6'' tiles in the second step.
 (The first instance was generated by \ama3, default prior, \blind.
 The second instance was produced by \ama4, $\epsilon=0.1$ prior, \blind.)
 }
\end{figure}

\begin{figure}[htb]
 \centering
 \includegraphics[width=0.7\linewidth]{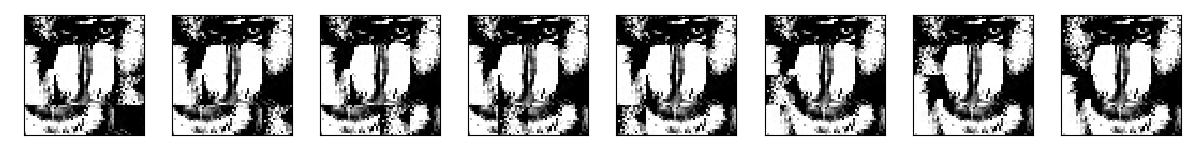}\\[1em]
 \includegraphics[width=0.7\linewidth]{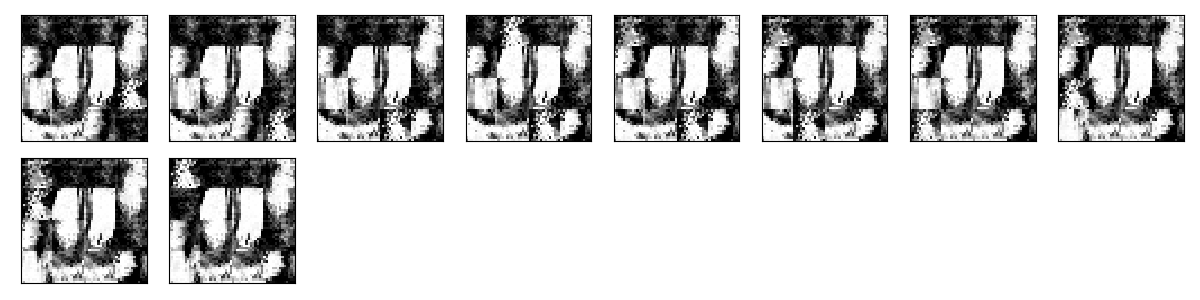}
 \caption{
 Examples from Mandrill 15-puzzle.
 The first plan is optimal (7 steps).
 The second plan is invalid because it is hard to distinguish different tiles.
 (Both instances were generated by \ama3, $\epsilon=0.1$ prior, \blind but with different hyperparameters $(F,\beta_1,\beta_3)$.)
 }
\end{figure}

\begin{figure}[htb]
 \centering
\includegraphics[width=0.7\linewidth]{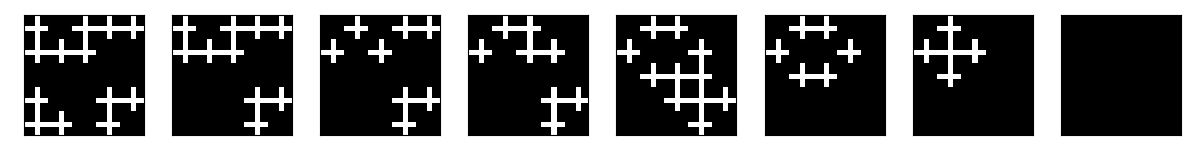}\\[1em]
\includegraphics[width=0.7\linewidth]{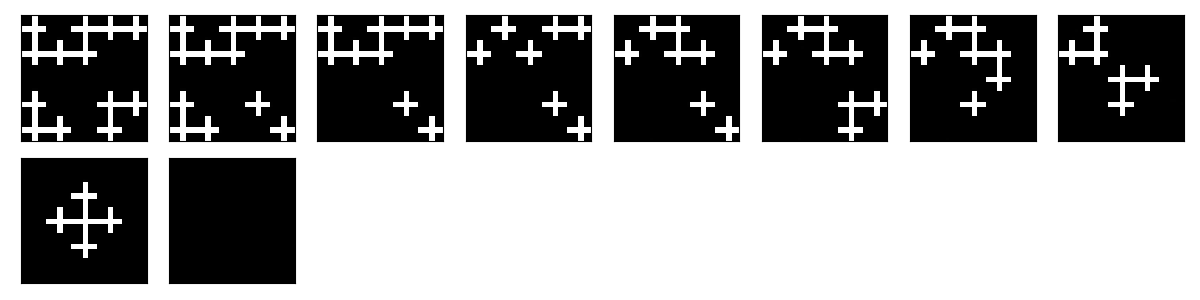}\\[1em]
\includegraphics[width=0.7\linewidth]{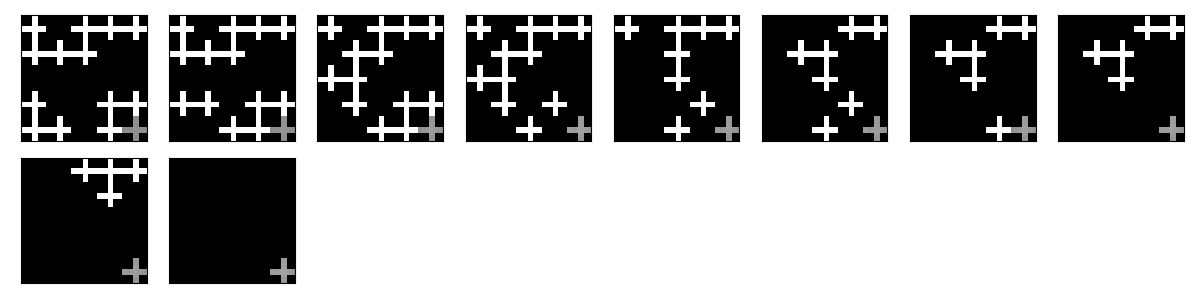}
 \caption{
 Examples from LightsOut.
 The first plan is optimal (7 steps).
 The second plan is suboptimal (9 steps) because it hits the bottom-right button twice (step 1 and 5), which is unnecessary (it reverses the effect).
 The third plan is invalid because of the bottom-right tile being above the threshold and the init/goal states do not match.
 (%
 The first instance was generated by \ama3, $\epsilon=0.1$ prior, \blind.
 The second instance was generated by the same network combined with LAMA.
 The third instance was produced by \ama3, $\epsilon=0.1$ prior, \blind with a different hyperparameter.)
 }
\end{figure}

\begin{figure}[htb]
 \centering
\includegraphics[width=0.7\linewidth]{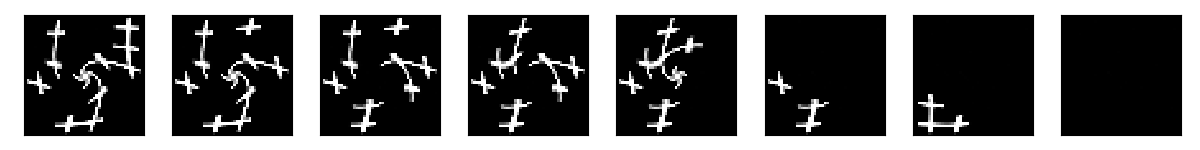}\\[1em]
 \caption{
 An optimal plan generated for Twisted LightsOut (7 steps).
 (Generated by \ama3, $\epsilon=0.1$ prior, \blind.)
 }
\end{figure}

\begin{figure}[htb]
 \centering
 \includegraphics[width=0.49\linewidth]{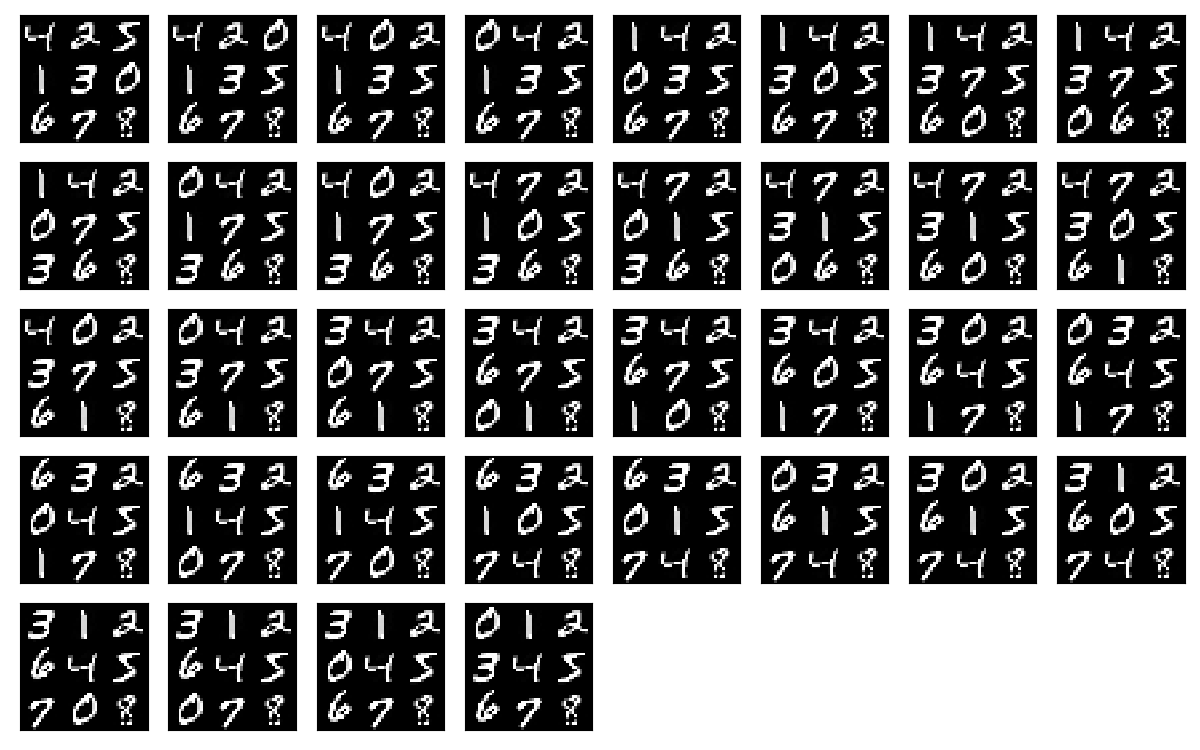}
 \includegraphics[width=0.49\linewidth]{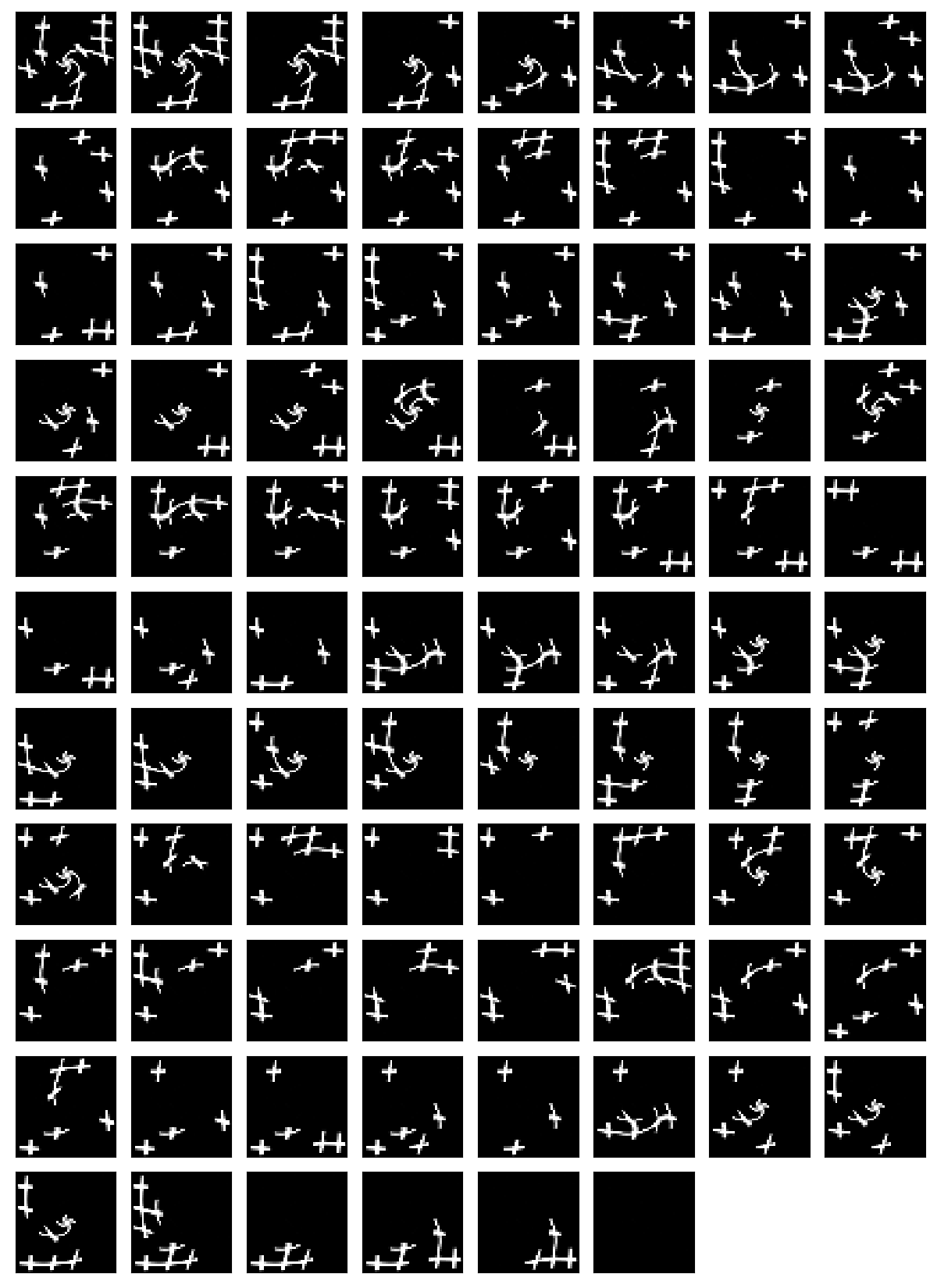}
 \caption{
 Suboptimal plans generated for 7-steps instances of MNIST 8-Puzzle and Twisted LightsOut by LAMA.
 Plans generated by LAMA tends to be hugely suboptimal compared to the suboptimal visual plans
 generated by \astar and admissible heuristics (e.g., \blind, \lmcut, \mands).
 See also: \refsec{sec:vs-nozsae} discusses why \astar + admissible heuristics can generate suboptimal visual plans.
 (%
 The first instance was generated by \ama4, default prior, LAMA.
 The second instance was generated by \ama3, $\epsilon=0.1$ prior, LAMA.)}
\end{figure}

\begin{figure}[htb]
 \centering
\includegraphics[width=0.7\linewidth]{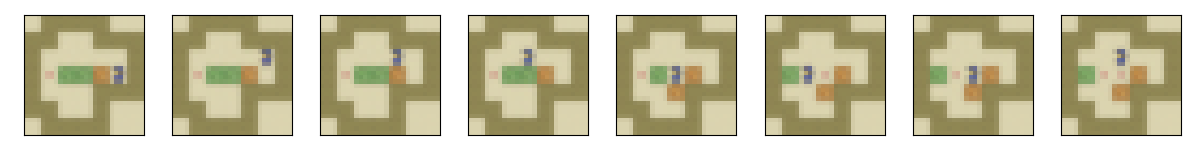}\\[1em]
\includegraphics[width=0.7\linewidth]{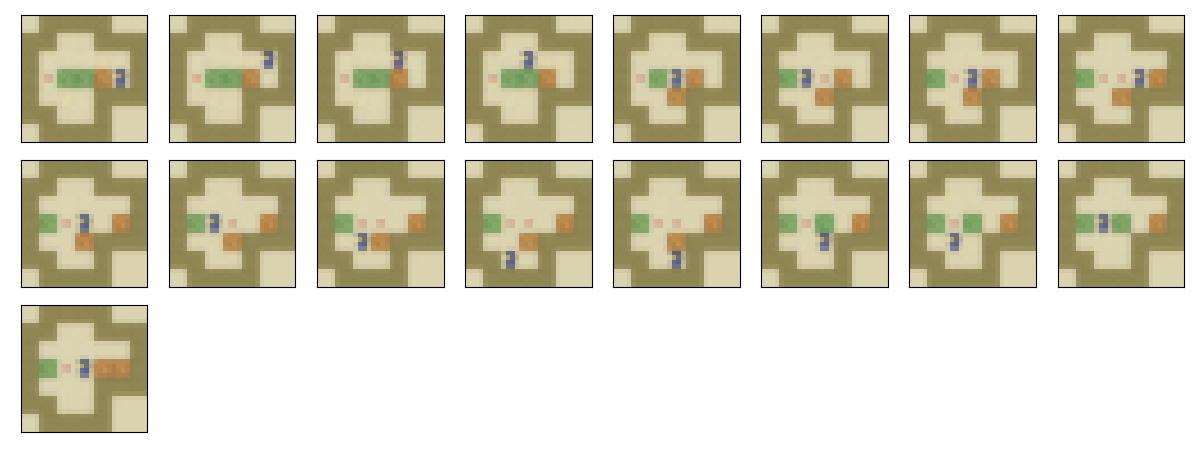}\\[1em]
\includegraphics[width=0.7\linewidth]{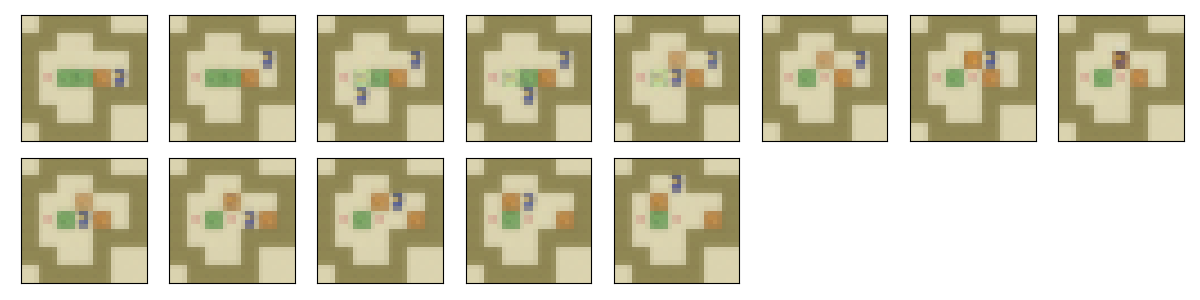}
 \caption{
 Examples from Sokoban.
 The first plan is optimal (7 steps for a 7-steps instance).
 The second plan is suboptimal (16 steps for a 14-steps instance).
 The third plan is invalid because it duplicates the player.
 (%
 The first instance was generated by \ama3, $\epsilon=0.1$ prior, \blind.
 The second instance was generated by \ama4, $\epsilon=0.1$ prior, \blind.
 The third instance was produced by \ama4, default prior, \mands.)
 }
 \label{fig:planning-examples2}
\end{figure}

\begin{figure}[htb]
 \centering
 \includegraphics[width=\linewidth]{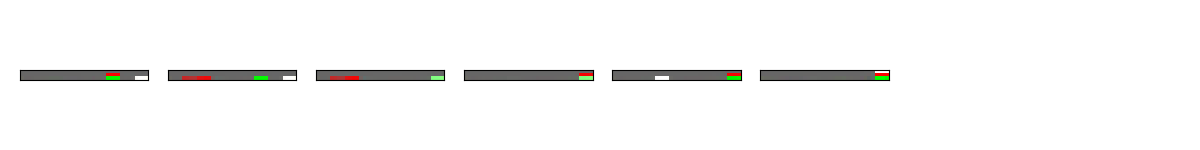}
 \includegraphics[width=\linewidth]{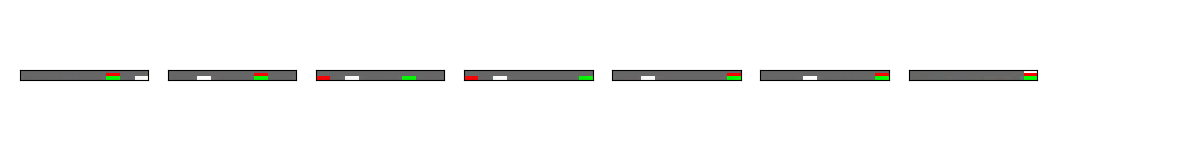}
 \includegraphics[width=\linewidth]{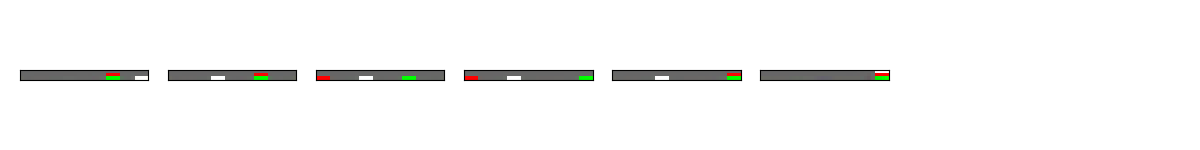}
 \caption{
 An invalid plan, a valid suboptimal plan and an optimal plan generated in $(3,9)$-ToH.
 (Each example was generated by \ama4, $\epsilon=0.1$ prior, \blind with different hyperparameters $(F,\beta_1,\beta_3)$.)
 }
 \label{fig:planning-hanoi}
\end{figure}

\begin{figure}[htb]
 \centering
 \includegraphics[width=0.7\linewidth]{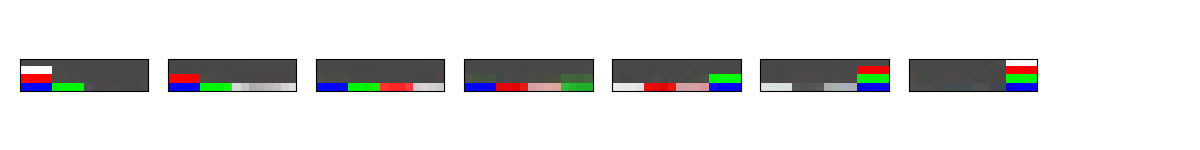}
 \includegraphics[width=0.7\linewidth]{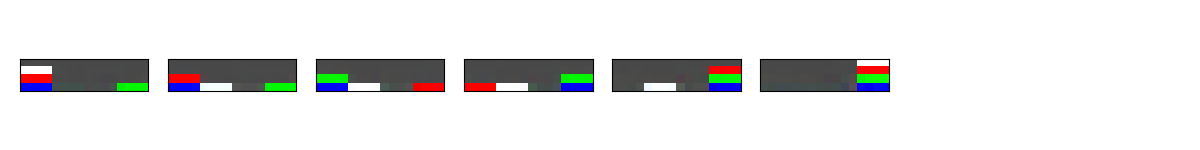}
 \includegraphics[width=0.7\linewidth]{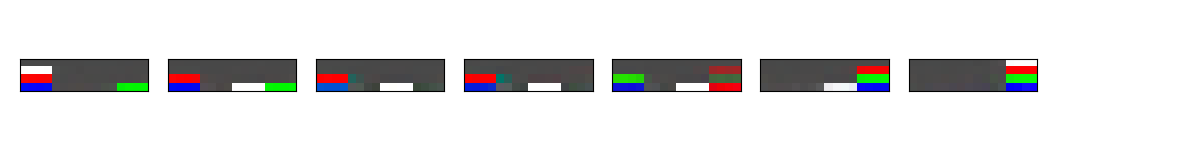}
 \caption{
 Invalid plans for $(4,4)$-ToH.
 (Each example was generated by \ama4, $\epsilon=0.1$ prior, \blind with different hyperparameters $(F,\beta_1,\beta_3)$.)
 }
 \label{fig:planning-hanoi2}
\end{figure}

\end{document}